\documentclass[8pt,a4paper]{elsarticle} 

\usepackage{caption,setspace}
\usepackage{booktabs} 
\usepackage{appendix}
\usepackage{rotating}
\usepackage{algorithm,algpseudocode}
\algblock{ParFor}{EndParFor}

\algnewcommand\algorithmicparfor{\textbf{parfor}}
\algnewcommand\algorithmicpardo{\textbf{do}}
\algnewcommand\algorithmicendparfor{\textbf{end\ parfor}}
\algrenewtext{ParFor}[1]{\algorithmicparfor\ #1\ \algorithmicpardo}
\algrenewtext{EndParFor}{\algorithmicendparfor}

\makeatletter
\newcommand{\algorithmfootnote}[2][\footnotesize]{%
  \let\old@algocf@finish\@algocf@finish
  \def\@algocf@finish{\old@algocf@finish
    \leavevmode\rlap{\begin{minipage}{\linewidth}
    #1#2
    \end{minipage}}%
  }%
}
\def\ps@pprintTitle{
 \let\@oddhead\@empty
 \let\@evenhead\@empty
 \def\@oddfoot{}%
 \let\@evenfoot\@oddfoot}
\makeatother

\newcommand{\mgin}{\vspace{2pt}}
\newcommand{\tabincell}[2]{\begin{tabular}{@{}#1@{}}#2\end{tabular}}
\newcommand{\BibTeX}{\textsc{B\kern-0.1emi\kern-0.017emb}\kern-0.15em\TeX}
\usepackage{threeparttable}
\usepackage{makecell}
\usepackage{textcomp}
\usepackage{mathtools,amsthm}

\usepackage{geometry}
\usepackage{amssymb}
\usepackage{multirow}
\usepackage{color}
\usepackage{threeparttable}
\usepackage{lineno,hyperref}
\usepackage{amsmath}    

\begin{document}
\begin{frontmatter}
\title{Parallel Sampling for Efficient High-dimensional Bayesian Network Structure Learning}


\author[mymainaddress]{Zhigao Guo\corref{mycorrespondingauthor}}
\cortext[mycorrespondingauthor]{Corresponding author}
\ead{zhigao.guo@qmul.ac.uk}

\author[mymainaddress]{Anthony C. Constantinou}

\address[mymainaddress]{Bayesian Artificial Intelligence Research Lab, School of Electronic Engineering and Computer Science, Queen Mary University of London, London, UK, E1 4NS.}

\begin{abstract}
Score-based algorithms that learn the structure of Bayesian networks can be used for both exact and approximate solutions. While approximate learning scales better with the number of variables, it can be computationally expensive in the presence of high dimensional data. This paper describes an approximate algorithm that performs parallel sampling on Candidate Parent Sets (CPSs), and can be viewed as an extension of MINOBS which is a state-of-the-art algorithm for structure learning from high dimensional data. The modified algorithm, which we call Parallel Sampling MINOBS (PS-MINOBS), constructs the graph by sampling CPSs for each variable. Sampling is performed in parallel under the assumption the distribution of CPSs is half-normal when ordered by Bayesian score for each variable. Sampling from a half-normal distribution ensures that the CPSs sampled are likely to be those which produce the higher scores. Empirical results show that, in most cases, the proposed algorithm discovers higher score structures than MINOBS when both algorithms are restricted to the same runtime limit.
\end{abstract}


\end{frontmatter}

\section{Introduction}

Bayesian Networks (BNs) are based on Pearl’s causal framework \citep{pearl-1988} and represent the most popular probabilistic graphical model for decision making under uncertainty in a wide range of applications Formally, a BN is the compact representation of the joint distribution over a set of random variables $X=(X_1,X_2,\cdots,X_n)$, as specified by structure $G$ and parameters $\theta$. The structure of a BN captures the conditional relationships between nodes, whereas the conditional parameters capture the magnitude and shape of those relationships. In practice, it is common to combine observational data with knowledge to construct the BN structure \citep{Lucas-2017, Anthony-2021-information-fusion} or to learn the parameters \citep{guo-2017-1,Yang-2019}.

In this paper, we focus on the problem of Bayesian Network Structure Learning (BNSL) from observational high-dimensional data. Given data $D$, a BN \{$G$,$\theta$\} can be learnt by maximising the likelihood:

\begin{equation}
P(G,\theta|D) =  P(G|D)P(\theta|G,D).
\end{equation}
and the parameters of the network can be learnt by maximising

\begin{equation}
P(\theta|G,D) = \prod_{i=1}^{n} P(\theta_{i}|\Pi_{i},D)
\end{equation}
where $\Pi_{i}$ denotes the parents of node $X_i$ as specified by structure $G$. Given $G$, the parameters of the network can be decomposed into maximising the fitting of local networks over the empirical distributions, $P(\theta_{i}|\Pi_{i},D)$. Therefore, the complexity of parameter learning is polynomial in the number of the variables. In contrast, the complexity of structure learning is NP-hard, where the number of potential acyclic structures grows super-exponential with the number of nodes $n$ \citep{Robinson-1973}:

\begin{equation}
f(n) = \sum_{i=1}^{n} (-1)^{i+1} \frac{n!}{(n-i)!n!} 2^{i(n-i)} f(n-1).
\end{equation}
As a result, the problem of structure learning is known to be considerably more challenging compared to the problem of parameter learning.

Structure learning algorithms are typically classified into three classes\citep{BN-Survey-2021}: a) the score-based class of learning that searches for the graph that maximises a given scoring function, b) the constraint-based class of learning that determines and orientates edges using conditional independence tests, and c) a hybrid learning process that combines the above two classes of learning. This paper focuses on the class of score-based learning.

A score-based algorithm is generally composed of two parts. The first part involves a scoring function that performs model selection on candidate structures given the data. The score is determined by the selected scoring function. Commonly used such functions include the BDeu \citep{Heckerman-1995, Buntine-1991}, BDs \citep{Marco-2016}, AIC \citep{Akaike-1973}, BIC/MDL \citep{Suzuki-1993}, MIT \citep{Luis-2006}, fNML \citep{Silander-2008}, and qNML \citep{Silander-2018}. The second part of a score-based algorithm involves a search heuristic or an optimisation strategy that transverses the candidate structure space. The search space can be explored with traditional heuristics such as greedy hill-climbing and Tabu search techniques \citep{Marco-2019-2} that transverse the graphical space via arc reversals, additions and removals. Alternatively, the search space can be explored via combinatorial optimisation of the Candidate Parent Sets (CPSs) which represent local networks of a node and its parents.

A distinction between search heuristics and optimisation strategies is that the former will usually score an acyclic graph only when the graph is visited, whereas the latter typically relies on pre-computed CPSs scores bounded by maximum in-degree (i.e., the maximum number of parents a node can have). Because combinatorial optimisation can be computationally expensive, these methods tend to employ pruning rules that prune off part of the CPSs. An important difference between the two is that, search heuristics represent approximate solutions that tend to get stuck in a local optimum graph, whereas combinatorial optimisation can be designed to offer exact solutions that return the highest scoring graph; for example, by ensuring that the pruning strategy is sound and preserves the CPSs that are present in the global maximum graph. Exact algorithms that operate on CPSs include Dynamic Programming (DP) \citep{Koivisto-2004,Silander-2006}, the A* algorithm \citep{Malone-2013}, Branch-and-Bound (B\&B) \citep{Jin-2000,Cassio-2011,vanBeek-2015}, and Integer Linear Programming (ILP) \citep{Jaakkola-2010,Cussens-2015}; whereas approximate algorithms that operate on optimisation of CPSs include the OBS algorithm \citep{Koller-2005}, the ASOBS algorithm \citep{mauro-2015,mauro-2018} and the MINOBS algorithm \citep{colin-2017}.

Evidently, the efficiency and scalability of the algorithms that operate on combinatorial optimisation of CPSs depend on the number of the data variables. For example, amongst the available exact algorithms, the well-established ILP algorithm is restricted to optimising up to one million CPSs, and this means that it is restricted to problems that contain (often considerably) less than 100 variables \citep{colin-2017,beek-2017,Anthony-2021-noisy-paper}. In contrast, approximate algorithms scale to tens of millions of CPSs \citep{beek-2017}, but at the cost of optimisation accuracy. In addition to the number of variables, the computational complexity of CPSs depends on the assumed level of maximum in-degree as well as the sample size of data set (see Section 2). As a result, many algorithms are often restricted to learning sparse structures of large networks, based on relatively limited data sample size; otherwise, computational and time complexity become limiting factors.  

This paper describes a score-based algorithm that combines the features of combinatorial optimisation with sampling and parallel processing, intended for learning from high dimensional data. The remainder of this paper is organised as follows: Section 2 provides the problem statement, Section 3 describes the algorithm, Section 4 presents the empirical results, and we provide our concluding remarks and directions for future research in Section 5.

\section{Problem Statement}

As discussed in Section 1, BNSL can be formulated as a combinatorial optimisation problem where each candidate graph is regarded as a combination of CPSs (one for each node) that satisfy acyclicity. Each CPS is related to a score that represents the posterior probability of the local structure, composed of the given node and its parents. The BDeu score, which we employ in this paper, represents one of the most commonly used functions in generating these scores, and is defined as:

\begin{equation}
Score_{BDeu}(G,D) = log P(G) + \sum_{i=1}^{n}\sum_{j=1}^{q_{i}} [log \frac{\Gamma (\frac{N^{'}}{q_{i}})}{\Gamma (N_{ij}+\frac{N^{'}}{{q_{i}}})}+\sum_{k=1}^{r_{i}}log \frac{\Gamma (N_{ijk}+\frac{N^{'}}{r_{i}q_{i}})}{\Gamma (\frac{N^{'}}{r_{i}q_{i}})}].
\end{equation}
where $j$ is the index over $q_{i}$ combinations of parents of node $i$, $k$ is the index over $r_{i}$ possible values of node $i$, $N_{ijk}$ is the number of instances in data $D$ where node $i$ has the $k$th value and its parents have the $j$th combination of values, $N_{ij}$ is the total number of instances in $D$ where the parents of nodes $i$ have the $j$th combination of values, and $N^{'}$ is the equivalent sample size which we assume to be equal to 1.

As an example, Table 1 presents a subset of the CPSs for node ``Disconnect" from the classic Alarm network. The corresponding BDeu scores are ordered by highest score, under the assumption the maximum in-degree is 3 and based on a synthetic data set with sample size 100,000 \citep{Bayesys-2020}.  

\begin{table}[H]
\scriptsize
\centering
\begin{tabular}{cccc}
\toprule
Child node &  Local BDeu score & CPS size & CPS \\
\hline
1  & -4,100.64     & 3     &  \{13, 23, 25\}  \\
1  & -4,590.91     & 3     &  \{8, 13, 25\}  \\
1  & -4,714.20     & 3     &  \{8, 13, 27\}  \\
1  & -5,385.43     & 3     &  \{8, 25, 27\}  \\
1  & -5,388.08     & 3     &  \{8, 13, 23\}  \\
...  & ...     & ...     &  ...  \\
1  & -6,9398.40     & 2     &  \{15, 29\}  \\
1  & -6,9417.56     & 1     &  \{34\}  \\
1  & -6,9471.84     & 1     &  \{29\}  \\
1  & -6,9798.83     & 1     &  \{37\}  \\
1  & -6,9865.10     & 0     &  \{\}  \\
\bottomrule
\end{tabular}
\caption{A subset of the CPSs of node ``Disconnect" from the Alarm network ordered by highest BDeu score, and assuming the sample size is 100,000 and the maximum in-degree is 3. The Child node with value 1 is node ``Disconnect", and nodes with values 2 to 37 represent the remaining 36 nodes of the network.}
\end{table}

The number of CPSs is determined by the number of nodes and the maximum node in-degree. For a graph $G$ with $n$ nodes and maximum in-degree $d$, the maximum number of CPSs over all nodes is 

\begin{equation}
f(n,d) = n\sum_{k=0}^{d} C_{n-1}^k.
\end{equation}
Table 2 presents the number of CPSs for different combinations of node size and maximum in-degree. It shows that the number of CPSs increases exponentially in the number of nodes and maximum in-degree, and this makes BNSL computationally prohibitive for high dimensional problems. As a result, the aim is to reduce the number of CPSs and improve the scalability and the efficiency of algorithms that operate on optimisation of CPSs. As discussed in the introduction, one way of achieving this is via pruning rules that reduce the number of CPSs considered for optimisation, often ensuring that the CPSs that make the global maximum graph are not pruned off, under the assumption that “\emph{a parent set cannot be optimal if its subsets have higher scores}” \citep{Koller-2005}.

Table 3 presents an example of the slightly modified pruning strategy by Campos and Ji \citep{Cassio-2010-pruning-rules} that is implemented in the GOBNILP software \footnote{https://www.cs.york.ac.uk/aig/sw/gobnilp/}, applied to a hypothetical data set with sample size 50 and the four variables of $\{$Asia, Tub, Smoke, Lung$\}$ that are part of the Asia network. Note that the Asia network contains eight variables, and that we only consider four of them in this example for simplicity. The CPSs highlighted in bold in Table 3, represent the CPS that will be considered in constructing the graph, while the rest represent the CPSs pruned off. In this example, 21 out of 32 possible CPSs are pruned off. For example, the CPS $\{$Lung$\}$ is pruned off for both Asia and Smoke because it generates a lower BDeu score than the empty CPS $\{\}$, which represents the score each node has when it has no parents. Another example is the CPS $\{$Tub, Smoke, Lung$\}$ which is pruned off for node Asia since the score of CPS $\{$Tub, Smoke, Lung$\}$ is lower than its subsets $\{$Tub, Lung$\}$ and $\{$Tub, Smoke$\}$. This pruning procedure is the most commonly used amongst the different pruning rules that have been proposed in the literature \citep{Cassio-2011,Cassio-2010,Cussens-2012,Joe-2017,Cassio-2020}, and which ensure the CPSs preserved will include the optimal structure.

\begin{table}[H]
\scriptsize
\centering
\begin{tabular}{cccccc}
\toprule
\multirow{2}{*}{Node size}&  \multicolumn{5}{c}{Maximum in-degree}  \\
\cline{2-6}
 & 1 & 2 & 3 & 4 & 5  \\
\hline
10 & $1.00 \times 10^{2}$ & $4.60 \times 10^{2}$ & $1.30 \times 10^{3}$ & $2.56 \times 10^{3}$ & $3.82 \times 10^{3}$ \\
\hline
50 & $2.50 \times 10^{3}$ & $6.13 \times 10^{4}$ & $9.83 \times 10^{5}$ & $1.16 \times 10^{7}$ & $1.07 \times 10^{8}$ \\
\hline
100 & $1.00 \times 10^{4}$ & $4.95 \times 10^{5}$ & $1.62 \times 10^{7}$ & $3.93 \times 10^{8}$ & $7.54 \times 10^{9}$ \\
\hline
500 & $2.50 \times 10^{5}$ & $6.24 \times 10^{7}$ & $1.04 \times 10^{10}$ & $1.29 \times 10^{12}$ & $1.28 \times 10^{14}$ \\
\hline
1,000 & $1.00 \times 10^{6}$ & $4.99 \times 10^{8}$ & $1.66 \times 10^{11}$ & $4.14 \times 10^{13}$ & $8.25 \times 10^{15}$ \\
\hline
5,000 & $2.50 \times 10^{7}$ & $6.25 \times 10^{10}$ & $1.04 \times 10^{14}$ & $1.30 \times 10^{17}$ & $1.29 \times 10^{20}$ \\
\hline
10,000 & $1.00 \times 10^{8}$ & $4.99 \times 10^{11}$ & $1.67 \times 10^{15}$ & $4.16 \times 10^{18}$ & $8.33 \times 10^{21}$ \\
\bottomrule
\end{tabular}
\caption{The number of CPSs for different combinations of node size and maximum in-degree.}
\end{table}
 
 \begin{table}[H]
\scriptsize
\centering
\begin{tabular}{ccccccccc}
\toprule
\multirow{2}{*}{Node}&  \multicolumn{8}{c}{Maximum in-degree}  \\
\cline{2-9}
 & 0 & 1 & 1 & 1 & 2 & 2 & 2 & 3  \\
\hline
\multirow{2}{*}{Asia} &\pmb{$\{\}$} & \pmb{$\{$Tub$\}$}     & $\{$Lung$\}$     &  $\{$Smoke$\}$   &   $\{$Tub, Lung$\}$    & $\{$Tub, Smoke$\}$     &   $\{$Smoke, Lung$\}$ &   $\{$Tub, Smoke, Lung$\}$ \\
&\pmb{-2.531} & \pmb{-2.381}     & -2.727     &  -3.032   &  -2.758     &   -2.948   &   -3.790 &  -3.302  \\
\hline
\multirow{2}{*}{Tub} &\pmb{$\{\}$} & \pmb{$\{$Lung$\}$}     & \pmb{$\{$Asia$\}$}     &  $\{$Smoke$\}$   &   \pmb{$\{$Smoke, Lung$\}$}    & $\{$Asia, Lung$\}$     &   $\{$Asia, Smoke$\}$ &   \pmb{$\{$Asia, Smoke, Lung$\}$} \\
&-\pmb{7.126} & \pmb{-5.292}     & \pmb{-6.976}     &  -7.426   &   \pmb{-3.790}    &   -5.323   &   -7.342 &  \pmb{-3.302}  \\
\hline
\multirow{2}{*}{Smoke} &\pmb{$\{\}$} & $\{$Tub$\}$     & $\{$Asia$\}$     &  $\{$Lung$\}$   &   $\{$Tub, Lung$\}$    & $\{$Asia, Tub$\}$     &   $\{$Asia, Lung$\}$ &   $\{$Asia, Tub, Lung$\}$ \\
& \pmb{-36.201} & -36.502     & -36.702     &  -37.760   &   -36.258    & -37.069     &   -38.823 &   -36.803 \\
\hline
\multirow{2}{*}{Lung} & \pmb{$\{\}$} & \pmb{$\{$Tub$\}$}     & $\{$Asia$\}$     &  $\{$Smoke$\}$   &   \pmb{$\{$Tub, Smoke$\}$}    & $\{$Asia, Tub$\}$     &   $\{$Asia, Smoke$\}$ &   $\{$Asia, Tub, Smoke$\}$ \\
& \pmb{-16.133} & \pmb{-14.299}     & -16.329     &  -17.692   &   \pmb{-14.056}    & -14.676     &   -18.450 &  -14.410 \\
\bottomrule
\end{tabular}
\caption{An example of the pruning strategy described in \citep{Cassio-2010-pruning-rules} applied to four of the eight variables available in the Asia network, assuming the maximum in-degree is 3, the score is BDeu, and a sample size of 50. CPSs in bold represent those retained after pruning.}
\end{table}

The efficacy of pruning rules can be significant in reducing the number of CPSs considered for optimisation. The level of pruning not only depends on the maximum in-degree assumed, but also on the number of observations available in the data. Table 4 presents the number and percentages of the CPSs preserved when optimising for BDeu using the pruning rules illustrated in Table 3, as implemented in the GOBNILP software. The results are presented over different sample size and maximum in-degree combinations, with application to the Pathfinder data set \citep{Bayesys-2020} that contains 109 variables. Note that while Table 4 suggests that the level of pruning decreases with sample size and increases with maximum in-degree, this outcome is not consistent across all data sets and it is possible to observe a reverse effect where the rate of pruning increases with sample size (see, for example, https://www.cs.york.ac.uk/aig/sw/gobnilp/ for further examples). While the number of local structures pruned off increases with maximum in-degree, the rate of pruning decreases since the number of possible structures increases much faster than the number of structures pruned off. In general, this means that the number of CPSs preserved is still a large enough number that makes BNSL intractable, even for data sets of moderate complexity.

\begin{table}[H]
\scriptsize
\centering
\begin{tabular}{cccccc}
\toprule
\multirow{2}{*}{\tabincell{c}{Maximum\\ in-degree}}&\multirow{2}{*}{\tabincell{c}{Number of\\ total CPSs}}&  \multicolumn{4}{c}{Number and $\%$ of CPSs preserved for each sample size}  \\
\cline{3-6}
 & &  100      & 1000      & 10,000  & 100,000  \\
\hline
\multirow{2}{*}{1} & \multirow{2}{*}{11,881}     &  1,539  & 4,323  & 7,781  & 10,281\\
&& 12.9$\%$  & 36.4$\%$  & 65.5$\%$ & 86.5$\%$\\
\hline
\multirow{2}{*}{2} & \multirow{2}{*}{653,455}    &  7,327 & 42,117 & 194,986 & 413,059\\
&& 1.1$\%$  & 6.4$\%$  & 29.8$\%$ & 63.2$\%$\\
\hline
\multirow{2}{*}{3} & \multirow{2}{*}{23,536,261} &  19,248  & 173,889  & 2,182,242 & 8,701,893\\
&& 0.1$\%$  & 0.7$\%$  & 9.3$\%$ & 36.9$\%$\\
\bottomrule
\end{tabular}
\caption{The number and percentage of CPSs preserved when applying the pruning rules of GOBNILP to the Pathfinder data set, over different sample size and maximum in-degree combinations.}
\end{table}

The number of possible DAGs that can be formulated from a given set of preserved CPSs has an upper bound. Specifically, there are $\{N_{1},N_{2},...,N_{n}\}$ CPSs for corresponding nodes $X=(X_1,X_2,\cdots,X_n)$, and from this we deduce that there are $N_{sum} = \prod_{i=1}^{n} N_{i}$ combinations of CPSs. These CPS combinations can be used to produce both Directed Cyclic Graphs (DCGs) and Directed Acyclic Graphs (DAGs). Therefore, the number of DAGs is upper-bounded by a number that is often magnitudes lower than $N_{sum}$.

\section{The Parallel Sampling MINOBS algorithm}

This section first introduces MINOBS, which is an approximate BNSL algorithm that operates on CPSs, and on which the algorithm proposed in this paper is based. We then describe the first modification which involves performing sampling on CPSs, and then the second modification that involves parallel processing.

\subsection{The MINOBS Algorithm}

The Memetic Insert Neighbourhood Order-Based Search (MINOBS) algorithm \citep{colin-2017}\footnote{The implementation of the MINOBS algorithm is freely available at https://github.com/kkourin/mobs.} is an extended version of the Order-Based Search (OBS) algorithm \citep{Koller-2005}. The OBS algorithm searches over the space of node orderings, which is significantly smaller than the unrestricted space of structures. Given a node ordering $O$, the possible parent sets of node $X_{i}$ can be defined as:

\begin{equation}
u_{i,O}=\{U: U \prec X_{i}\}
\end{equation}
where $U \prec X_{i}$ means all nodes in $U$ precede $X_{i}$ in $O$. Then, the optimal parent set of $X_{i}$ is

\begin{equation}
PS_{O}^{*}(X_{i})=\mathop{\arg\max}_{U \in u_{i,O}} Score(X_{i},U)
\end{equation}
where $Score(X_{i},U)$ represent the scores of local networks that can be pre-computed beforehand. The optimal parent sets for different nodes form the optimal structure consistent with the given ordering $O$. Therefore, the score of the optimal structure $G^{*}$ is

\begin{equation}
Score(G^{*}) = \sum_{i=1}^{n} PS_{O}^{*}(X_{i}).
\end{equation}

The OBS algorithm involves a swap-adjacent neighbourhood operator that is used to generate alternative node orderings. For example, given the ordering $O = \{X_{1}\prec X_{2}\prec X_{3}\prec X_{4}\}$, a new ordering $O^{'}=\{X_{1}\prec X_{3}\prec X_{2}\prec X_{4}\}$ can be generated by swapping adjacent variables $X_{2}$ and $X_{3}$. Over the space of orderings, the hill-climbing algorithm with random restarts and tabu list is applied to find the optimal ordering, where the optimal ordering also contains the optimal structure.

MINOBS improves the efficiency of OBS in determining the optimal node ordering by performing two, instead of one, modifications to the node ordering at a time. Firstly, instead of the swap-adjacent neighbourhood operator, MINOBS uses the insert neighborhood operator that makes it more likely to escape from a local optimum \citep{colin-2017}. For example, MINOBS can obtain ordering $O^{'}=\{X_{2}\prec X_{3}\prec X_{1}\prec X_{4}\}$ from ordering $O = \{X_{1}\prec X_{2}\prec X_{3}\prec X_{4}\}$ by directly inserting $X_{1}$ to index $3$, whereas OBS would require at least two swap-adjacent neighbourhood operations to reach that state of ordering. The abbreviation Insert Neighborhood OBS (INOBS) comes from this first modification. Secondly, the optimisation performance of INOBS is further improved through a memetic implementation where INOBS is applied to a population of orderings that are optimised by the INOBS method. Then, crossover and mutation operators are applied onto the locally optimal orderings to expand the population of orderings. Both orderings produced by the crossover and mutation operators are optimised by the INOBS algorithm. Finally, the extended population is pruned to maintain the initial size of the ordering population. The abbreviation Memetic INOBS (MINOBS) comes from this second modification.

\subsection{The CPS Sampling Process}

Because the evaluation of the different combinations of CPSs can be performed separately, we propose to partition the combinatorial optimisation problem into subsets of CPSs, and optimise each subset in parallel. Moreover, because high dimensional problems involve notoriously large numbers of CPSs, we choose to sample a percentage $p$ of CPSs from the available CPSs of every node, and form a new CPS subset of size, $\{p \cdot N_{1},p \cdot N_{2},...,p \cdot N_{n}\}$, where the number of possible combinations of CPSs is $\prod_{i=1}^{n} (p \cdot N_{i})$. Each subset of CPSs can then be optimised using a single processor, or thread, and the process can be repeated $m$ times; one for each subset. For $m$ subsets of CPSs that correspond to $m$ independent optimisation problems, the total number of CPS combinations is:

\begin{equation}
N_{sum}^{'} = m \prod_{i=1}^{n} (p \cdot N_{i}).
\end{equation}
To approximate $N_{sum}^{'}=N_{sum}$, we set $m$ subsets of CPS sampled to

\begin{equation}
m=\frac{1}{p^{n}}.
\end{equation}

From Equation (10), we can conclude that the number of CPSs sampled for each $m$ increases as the number of nodes $n$ increases, while the sampling percentage $p$ decreases. Figure 1 illustrates the range of sampling percentage $p \in [1$\%$, 99$\%$]$ required for $N_{sum}^{'}=N_{sum}$ over different ranges of $m$ and over different number of nodes $n \in [100, 1000]$. The results show that lower sampling percentages allow for thousands of independent multi-threaded processes to be performed before the total number of parallel combinations of CPSs $N_{sum}^{'}$ reaches the total combinations of CPSs $N_{sum}$ when performing non-parallel optimisation across all preserved CPSs. For example, the empirical experiments show that when $m$ ranges from 1 to 10 (the light blue area in Figure 1), the total number of parallel combinations of CPSs $N_{sum}^{'}$ will always be lower than the total combinations of CPSs $N_{sum}$, as long as we sample less than approximately 98$\%$ of CPSs $N_{sum}$ for each parallel run $m$. This implies that there is a considerable reduction in computational complexity to be explored via an approach that combines sampling with parallel processing.      

\begin{figure}[H]
\centering
\includegraphics[height=9cm,width=18cm]{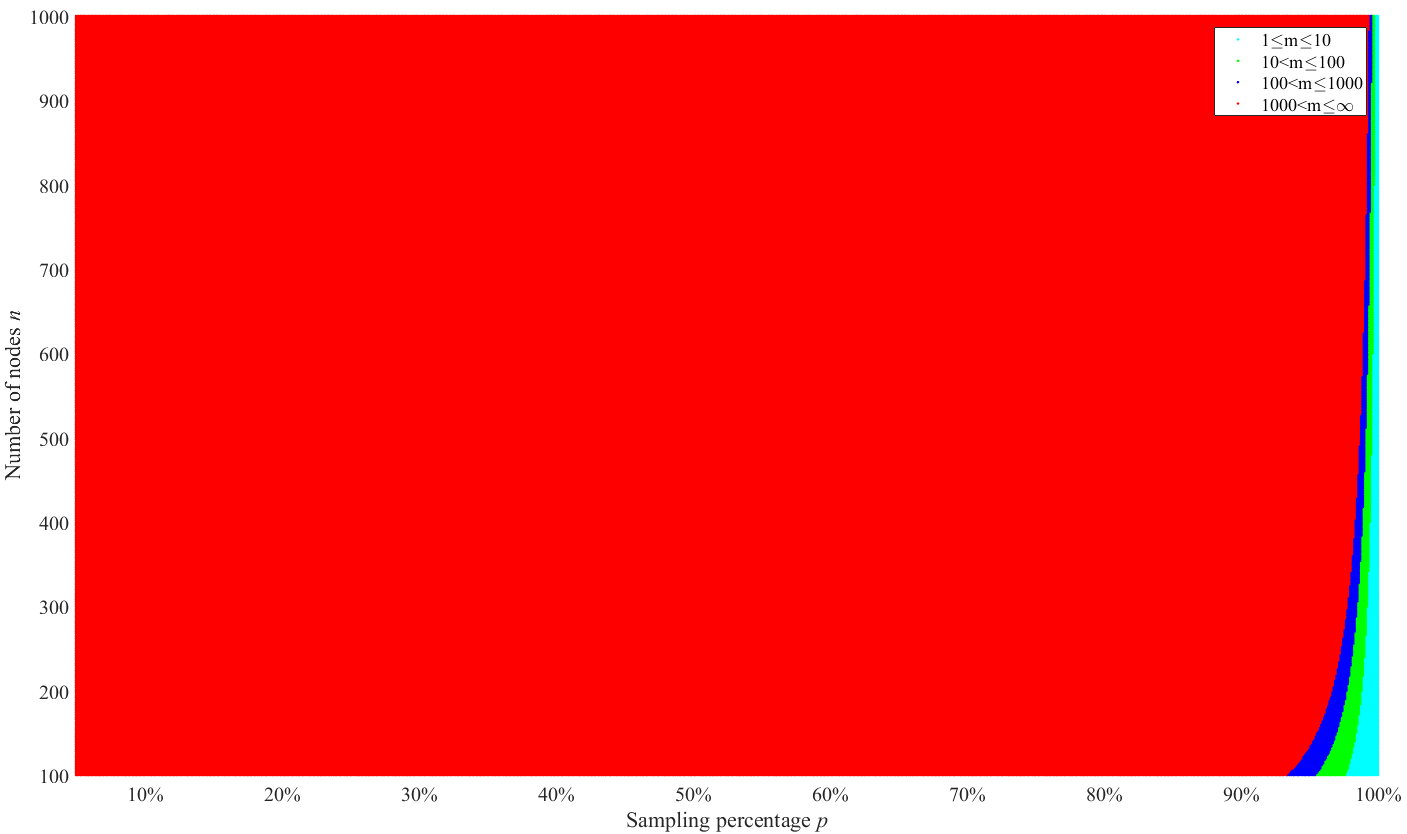}
\captionof{figure}{{The sampling rate $p \in [1$\%$, 99$\%$]$ required for $N_{sum}^{'}=N_{sum}$ for different node-size $n \in [100, 1000]$ and different ranges of $m$.}}
\label{fig:1}
\end{figure} 

Despite the potential for significant reduction in computational complexity, the combinatorial optimisation problem remains intractable for very large networks that include thousands of nodes. In order to meaningfully improve computational efficiency, we found that we had to sample lower rates of CPSs that violate Equation (10). However, decreasing the percentage of CPSs considered in each parallel run can have a considerable impact on the accuracy of the learnt graph. To minimise the negative repercussions on accuracy and ensure that the learnt graph is reasonably accurate, we sample CPSs that are most likely to be present\footnote{Note that only one CPS per node can form part of the learnt structure.} in the optimal structures. We achieve this by ordering the CPSs for each node by their local score and assume that the ordering of scores assigned to each CPS follows a half-normal distribution:

\begin{equation}
f(CPS_{i,k};\sigma) = \frac{\sqrt {2}}{{\sigma \sqrt {\pi} }} \exp(-\frac{CPS_{i,k}^2}{2\sigma^2}) = \frac{\sqrt {2}}{{\sigma \sqrt {\pi} }} \exp(-\frac{k^2}{2\sigma^2}),k \geq 1
\end{equation}
where $\sigma$ is the standard deviation, and $CPS_{i,k}$ is an ordered set of CPSs in which the $k$th CPS represents the $k$th highest scoring CPS amongst all CPSs of node $i$. Because the list of CPSs of node $i$ is sorted by highest BDeu score, such as $\{CPS_{i,1},CPS_{i,2},...,CPS_{i,N_{i}}\}$, we assume that the probabilities for each ordered CPS being present in the optimal structure have the following relationship:

\begin{equation}
f(CPS_{i,1};\sigma) \geq f(CPS_{i,2};\sigma) ... \geq f(CPS_{i,N_{i}};\sigma).
\end{equation}
Since the goal of score-based learning is to find the structure with the highest score, the CPSs that have the highest scores for each node are more likely to form part of that structure. Because we are interested in acyclic graphs, there is no guarantee that the highest scoring CPSs will be present in the optimal DAG. This is because the number of possible DCGs increases much faster than the number of possible DAGs, in the number of nodes and maximum-in-degree. As a result, the highest scoring CPSs tend to produce DCGs with multiple cycles. This means that the optimal DAG consists of multiple CPSs that do not represent the highest scoring CPS of a given node. In general, acyclicity is enforced by iteratively replacing the CPSs that contain high number of parents with their highest scoring subset, since CPSs that contain many nodes are more likely to lead to cycles.

For an example, consider the Alarm network discussed in Section 2, which contains 37 variables. The total number of CPSs that are preserved after pruning (refer to Table 3) is 12,292. Table 5 presents the number of CPS preserved per node along with the DAG-optimal\footnote{This example is based on one of the optimal Markov equivalent DAGs.} CPS index of ordered CPSs; i.e., index 1 represents the highest scoring CPS for a given child node. Indices marked with (*) represent empty CPSs that are placed at the end of each ordering, and which are required to guarantee acyclicity. The results depicted in Table 5 are based on the ILP algorithm available in the GOBNILP software, which performs exact learning. Excluding the 11 nodes whose empty CPS is selected to be part of the optimal DAG, the results in Table 5 show that the optimal CPSs are found at index 1 in 13 cases, between indices 2-10 in three cases, between indices 11-50 in four cases, between indices 51-100 in three cases, and at an index higher than 100 in three cases. This example shows that most of the optimal CPSs are found very early in the ordering of highest scoring CPSs.

\begin{table}[H]
\scriptsize
\centering
\begin{tabular}{ccc}
\toprule
Node index &  Number of CPSs & Optimal CPS index\\
\hline
1 &  767 & 237\\
\hline
2 &  261 & 1\\
\hline
3 &  284 & 284($*$)\\
\hline
4 &  310 & 310($*$)\\
\hline
5 &  273 & 1\\
\hline
6 &  7 & 7($*$)\\
\hline
7 &  265 & 44\\
\hline
8 &  765 & 1\\
\hline
9 &  307 & 1\\
\hline
10 &  57 & 57($*$)\\
\hline
11 &  350 & 1\\
\hline
12 &  28 & 28($*$)\\
\hline
13 &  705 & 1\\
\hline
14 &  174 & 174($*$)\\
\hline
15 &  179 & 97\\
\hline
16 &  109 & 52\\
\hline
17 &  100 & 51\\
\hline
18 &  104 & 104($*$)\\
\hline
19 &  313 & 1\\
\hline
20 &  72 & 72($*$) \\
\hline
21 &  13 & 1 \\
\hline
22 &  5 & 5($*$) \\
\hline
23 &  445 & 445($*$) \\
\hline
24 &  483 & 8 \\
\hline
25 &  999 & 11 \\
\hline
26 &  953 & 407 \\
\hline
27 &  1022 & 23 \\
\hline
28 &  394 & 257 \\
\hline
29 &  756 & 2 \\
\hline
30 &  692 & 1 \\
\hline
31 &  378 & 378($*$) \\
\hline
32 &  73 & 1 \\
\hline
33 &  1 & 1 \\
\hline
34 &  370 & 2 \\
\hline
35 &  30 & 17 \\
\hline
36 &  62 & 1 \\
\hline
37 &  186 & 1 \\
\bottomrule
\end{tabular}
\caption{The number of CPS preserved after pruning off the lower scoring CPS supersets (refer to Table 3), along with the index of the DAG-optimal CPSs. This example is based on the Alarm network.}
\end{table}

We now repeat the above example on 40 real-world data sets\footnote{https://github.com/arranger1044/DEBD \label{2}} with variables ranging from 16 to 500. The number of CPSs preserved after pruning varies between 114 to 888,746 for each individual node, and the total number of CPSs for each individual data set varies from 4,701 to 38,858,778. Exact learning algorithms, such as the commonly used ILP, are generally restricted to globally optimal structures obtained from up to 1,000,000 CPSs. This means that exact optimisation is impractical on the larger data sets, which is why approximate learning algorithms are preferred when working with high dimensional data. Because of this limitation, we have applied the ILP algorithm on data sets where the number of CPSs preserved is lower than 1,000,000, and the MINOBS algorithm on the remaining, more complex data sets on which ILP cannot be applied. For practical reasons, we have also applied a runtime limit of 10 hours per experiment, to both algorithms. This means that across all the 40 data sets tested, some results will reflect the highest scoring structure and some a high scoring structure.

Figure 2 presents two graphs depicting the number of times an ordered CPS index was present in the optimal DAG, across all ordered CPSs and over all 40 experiments. The graph at the bottom represents a truncated version of the graph above to improve visualisation, where the y-axis is limited to 100 occurrences. The results show that while the learnt DAGs contain most of the highest scoring CPSs (i.e., those placed at index 1), large parts of those DAGs contain CPSs that are far from the highest local score. The shape of the empirical distribution in Figure 2 appears to approximate a half-normal distribution, and it is on this basis we decided to simplify the sampling process of CPSs and assume that the ordered probability of a CPS being present in the optimal DAG follows a half-normal distribution.  

\begin{figure}[H]
\centering
\includegraphics[height=9cm,width=18cm]{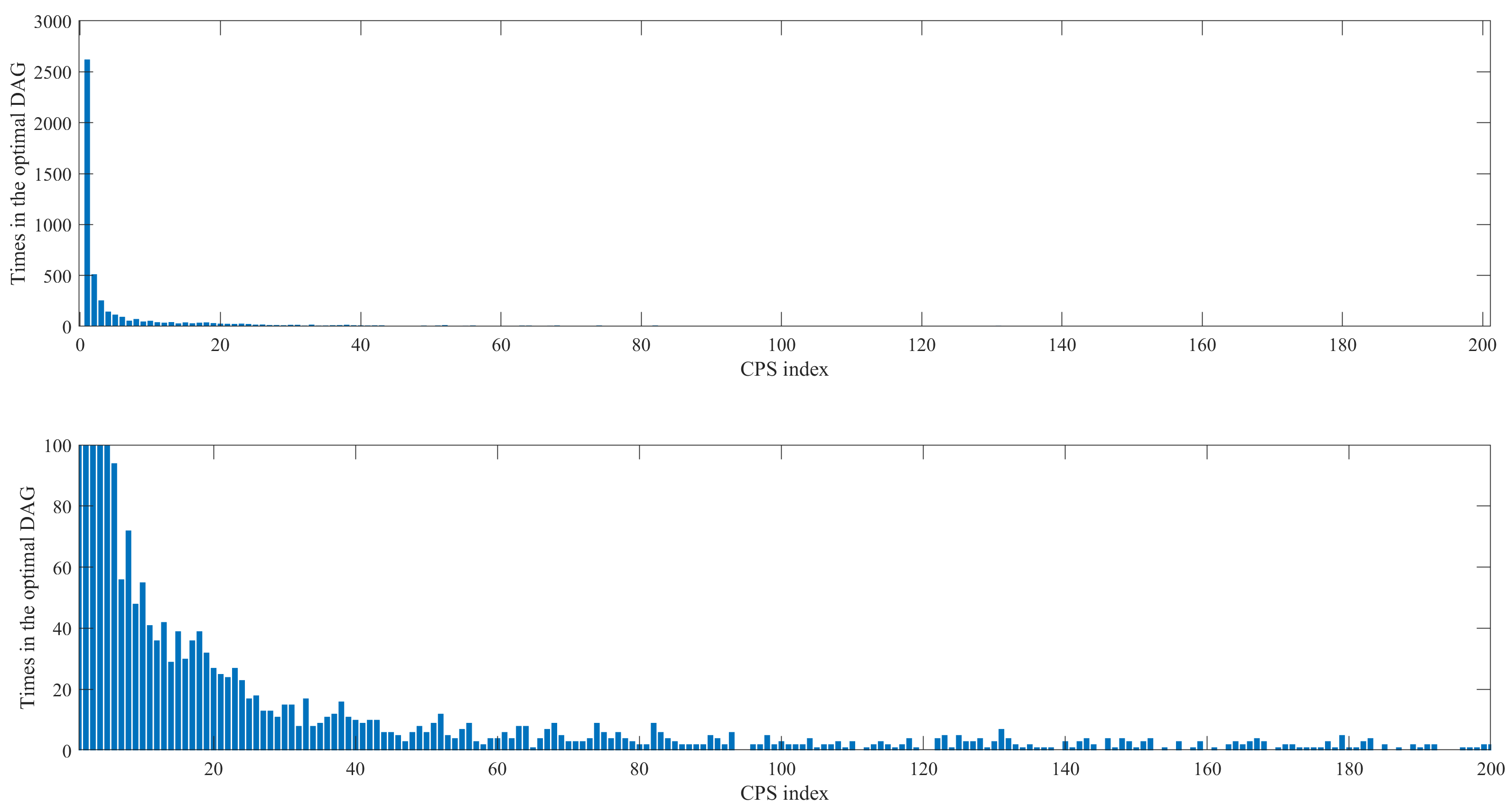}
\caption{The number of times an ordered CPS index appeared in the optimal DAG. The results are based on 40 data sets. The graph at the bottom is a truncated version of the graph at the top. Only the first 200 of the 888,746 CPS indices are shown.}
\end{figure}

We test two different sampling strategies to obtain CPSs from the half-normal distributions. The first strategy involves sampling only from the first $N_{i}\cdot p$ half-normally distributed CPSs based on \citep{Zhigao-2020}. The second sampling strategy involves sampling from the entire half-normal distribution; i.e., $N_{i}\cdot p$ CPSs over all $N_{i}$ CPSs, as described in Algorithm 1 and based on \citep{Dahua-2021}. Note that setting $p=100\%$ in Algorithm 1 makes the first sampling strategy equal to the second sampling strategy.

\begin{algorithm}[!h]
\hspace*{0.1cm} \textbf{Input:} $\sigma \rightarrow$ standard deviation, $p \rightarrow$ sampling percentage, $s \rightarrow$ index of CPS subset,  $CPS \rightarrow$ CPSs after pruning.\\ \mgin
\hspace*{0.1cm} \textbf{Output:} $CPS_{s}^{'} \rightarrow$ the $s$-th subset of sampled CPSs.
\begin{algorithmic}[1]
\caption{: Sampling CPSs from a discrete half-normal distribution}\label{euclid}
\Function{Sampling}{$s$, $\sigma$, $p$, $CPS$} 
 \mgin
  \If{$s$=1} \vspace{2 mm}  \Comment{\textit{If $s=1$, sample the first $N_{i}\cdot p$ CPSs}}.
    \For{$i\gets 1, n$} \mgin
      \State $CPS_{i}^{'}=\{CPS_{i,1},CPS_{i,2},...,CPS_{i,p\cdot N_{i}},CPS_{i,N_{i}}\}$, \Comment{\textit{$N_{i}$ is the number of CPSs for node $i$}}.\mgin
    \EndFor \mgin
  \Else{} \vspace{1 mm}  \Comment{\textit{If $s\neq1$, sample from the entire half-normal distribution}}.
  \For{$i\gets 1, n$} \mgin  \algorithmfootnote{$y_0$ denotes the initial value.}
    \State $pdf_{i}$=$HalfNormal(\sigma, N_{i})$; \Comment{\textit{Generate the probability density function}}.\mgin
    \State $CPS_{i}^{'}$=$DiscreteSample(p\cdot N_{i},pdf_{i})$; \Comment{\textit{Sample CPSs from the half-normal distribution}}.\mgin
  \EndFor \mgin
  \EndIf \vspace{1 mm}
\EndFunction
\end{algorithmic}
\end{algorithm}

We use Algorithm 1 to sample a pre-determined rate $p$ of CPSs for each node $i$, where the number of sampled CPSs is $N_{i}\cdot p$. The lower the sampling rate, the faster the number of CPSs is reduced. The aim is to improve the efficiency of learning from high dimensional data, while at the same time minimising the impact on structure learning accuracy. Figures 3a and 3b illustrate an example where the two sampling strategies are applied to a hypothetical node with 10,000 possible CPSs, assuming a sampling rate $p=20\%$ and standard deviation $\sigma=1000$. The first sampling strategy samples CPSs from the first 2,000 (the first $20\%$) highest scoring CPSs, whereas the second sampling strategy samples CPSs from the entire distribution.

\begin{figure}[H]
\centering
\includegraphics[height=9cm,width=18cm]{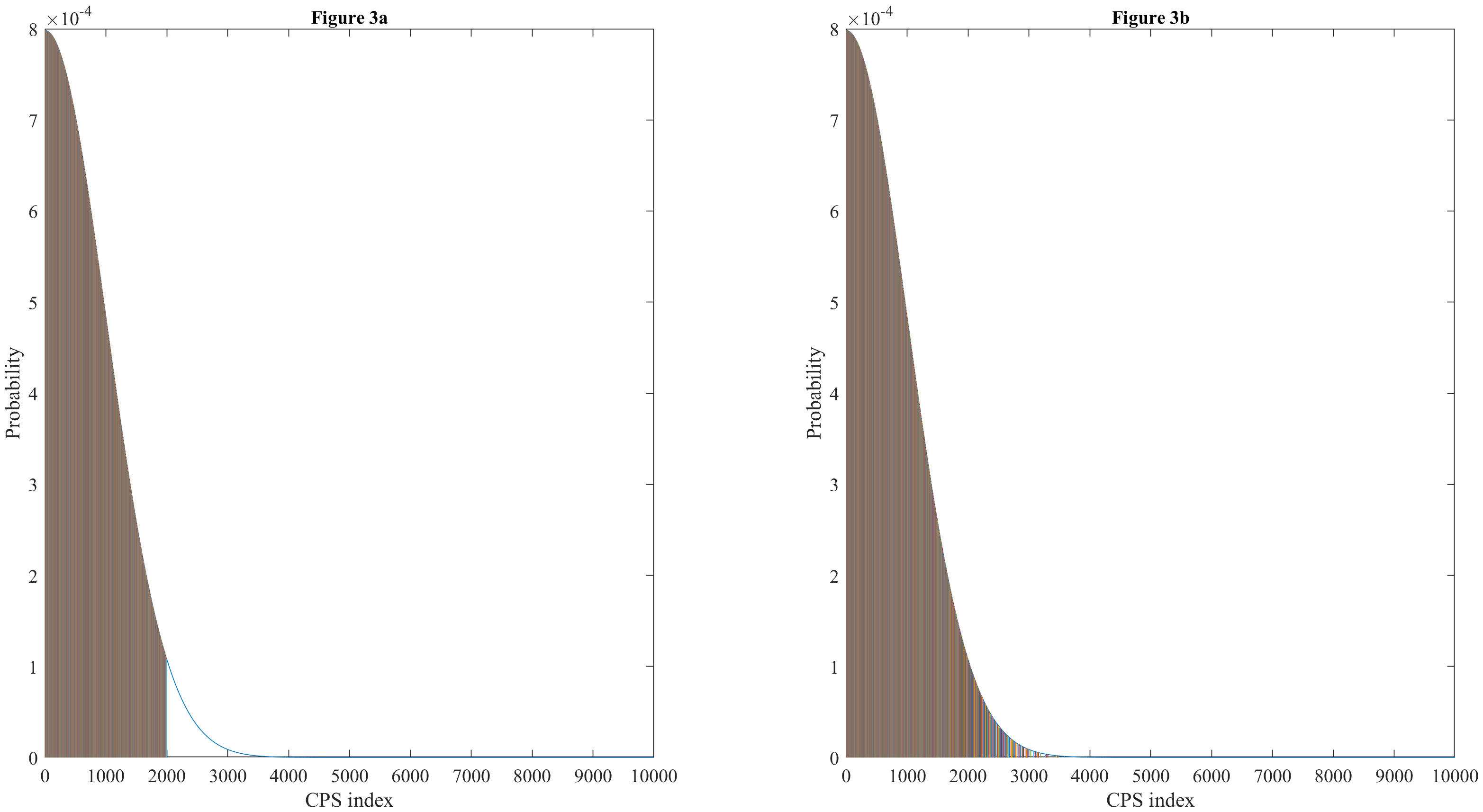}
\caption{An illustration of the two sampling strategies applied to a hypothetical node with 10,000 CPSs, assuming a sampling rate of $p=20\%$ and standard deviation $\sigma=1000$. The first sampling strategy is restricted to sampling CPSs from the first 2,000 ($p=20\%$) CPSs of the distribution (Fig 3a), whereas the second strategy (Fig 3b) samples CPSs from the entire distribution.}
\end{figure}

\subsection{Parallel Optimisation}

The combinatorial optimisation of CPSs can be divided into local combinatorial optimisation problems; one per node. This means that the sampling process of CPSs for each node in the network can be performed in parallel. In this subsection, we present a parallel optimisation framework that is based on the sampling strategy described in Algorithm 1.

Figure 4 illustrates how we combine MINOBS (refer to subsection 3.1) with this framework to produce the Parallel Sampling MINOBS (PS-MINOBS) algorithm. Specifically, once the full set of CPSs is obtained for each node, we divide learning into subprocesses that we execute in parallel. Each subprocess involves a sampling strategy and a set of sampled CPSs that is given as an input to MINOBS for structure learning. Each parallel process returns an optimal DAG, and the highest scoring DAG across all parallel processes is then selected as the preferred DAG. The procedure of this modified algorithm is described in Algorithm 2, where $T$ represents the runtime limit for MINOBS to spend on combinatorial optimisation. The implementation of the PS-MINOBS algorithm is freely available online\footnote{https://github.com/ZHIGAO-GUO/Parallel-MINOBS}. 

\begin{figure}[H]
\centering
\includegraphics[height=7cm,width=11cm]{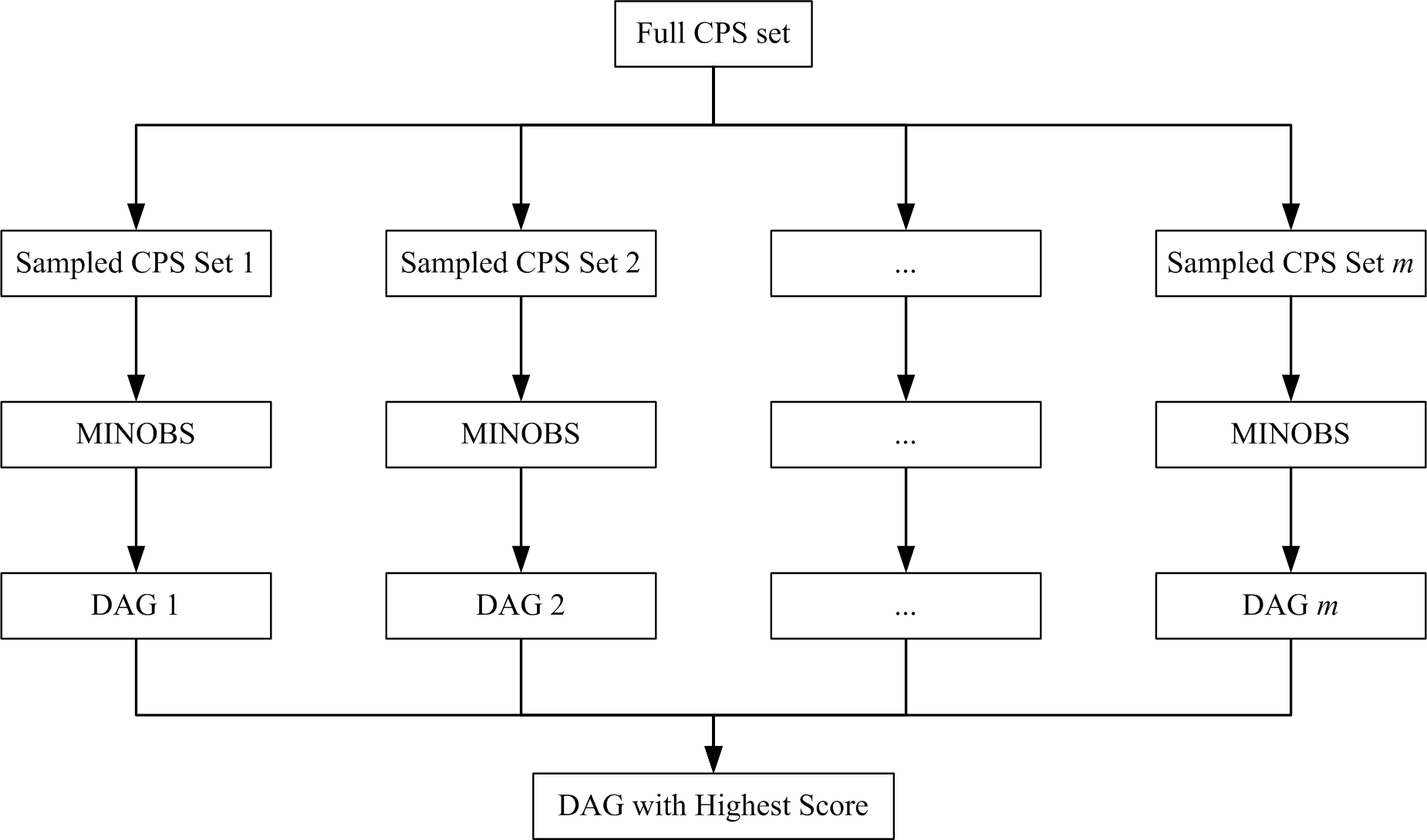}
\caption{: The Parallel Sampling MINOBS algorithm}\label{euclid}
\label{fig:1}
\end{figure}

\begin{algorithm}[!h]
\hspace*{0.1cm} \textbf{Input:} $\sigma \rightarrow$ standard deviation, $p \rightarrow$ sampling percentage, $m \rightarrow$ number of parallel threads,  $CPS \rightarrow$ CPSs after pruning, $T \rightarrow$ runtime limit.\\ \mgin
\hspace*{0.1cm} \textbf{Output:} $G^{*} \rightarrow$ highest scoring DAG.
\begin{algorithmic}[1]
\caption{: The Parallel Sampling MINOBS algorithm}\label{euclid}
\ParFor{$s\gets 1, m$} \mgin
  \State $CPS_{s}^{'} \gets$ SAMPLING($s,\sigma,p,CPS$)  \Comment{\textit{Sampling CPSs in parallel}}.\mgin
\EndParFor \mgin
\ParFor{$s\gets 1, m$} \mgin
    \State $\{Score_{s},G_{s}\} \gets$ MINOBS($T,CPS_{s}^{'}$) \Comment{\textit{Seek for the optimal structure of each CPS subset}}.\mgin
\EndParFor \mgin
\State $s^{*}= arg \max_{s=1}^{m} Score_{s}$ \mgin
\State $G^{*}= G_{s^{*}}$ \mgin
\end{algorithmic}
\end{algorithm}

\section{Experiments and results}

Experiments are carried out on the 6-core 12-thread Intel Core i7-8700 CPU at 3.2 GHz with 32 GB of RAM. However, only up to 10 threads are utilised, and only 25GB of RAM were made available for learning and optimisation. We have tested PS-MINOBS on the nine high-dimensional real-world data sets \footnote{https://github.com/arranger1044/DEBD \label{2}} presented in Table 6, where the number of variables ranges from 100 to 910. The BDeu scores for the CPS are pre-computed using the GOBNILP software, which employs the pruning strategy described in Section 2 (refer to Table 3). Due to practical limitations, the CPSs scored are restricted to maximum in-degree of 3 for data sets with less than 500 variables (i.e., Audio-train, Jester-train, Pumsb-star-test, and Kosarek-train), and to maximum in-degree of 2 for data sets that contain 500 or more variables. Table 6 lists the total number of CPSs preserved after pruning. These values show that the size of CPSs across all case studies is far higher than the maximum size of CPSs an exact algorithm, such as ILP, can handle.

We evaluate the performance of PS-MINOBS with reference to the corresponding results of the MINOBS algorithm. We investigate different hyperparameter inputs for PS-MINOBS by exploring its performance over a) the sampling percentages of $p=10\%$, $p=20\%$, $p=25\%$, $p=33\%$, $p=50\%$, and $p=80\%$, and b) the number of parallel sampling processes of $m=10$, $m=5$, $m=4$, $m=3$, and $m=2$. The standard deviation $\sigma$ is set to

\begin{equation}
\sigma_{i} = 0.5\cdot p \cdot N_{i}
\end{equation}
where $N_{i}$ represents the number of CPSs preserved after pruning for each node $i$.

The results are investigated in terms of BDeu score relative to runtime spent for optimisation. We compare the discrepancy $\Delta$ between the BDeu scores produced by the two algorithms over 30-minute intervals, and up to a total of four hours spent on optimising the CPSs for each experiment. The total number of hours needed to complete the experiments was 216. We define the discrepancy in BDeu scores as

\begin{equation}
\Delta = (S^{*}-S_{i})/S^{*}
\end{equation}
where $S^{*}$ denotes the BDeu score of the structure learnt by MINOBS and $S_{i}$ the BDeu score of the corresponding structure learnt by PS-MINOBS at the $i$th parallel process. A positive $\Delta$ indicates a higher score for PS-MINOBS, and vice-versa. When $i=1$, PS-MINOBS uses the first sampling strategy which involves sampling CPSs from the truncated half-normal distribution of CPSs ordered by BDeu score (refer to Figure 3a). When $i>1$, PS-MINOBS uses the second sampling strategy, which involves sampling CPSs from the entire distribution (refer to Figure 3b).

Tables 7 to 15 present the $\Delta$ scores as defined above, where each table corresponds to a data set from the nine data sets described in Table 6. PS-MINOBS is evaluated over different hyperparameter combinations of $p$ and $m$. Each hyperparameter combination associates with a ``Highest DAG $\Delta$" row (in Tables 7 to 15) that highlights the highest performance across parallel processes; i.e., the one that PS-MINOBS will return. For example, in Table 7, PS-MINOBS discovers a DAG with a BDeu score that is 0.031{\textperthousand} more accurate than that discovered by MINOBS after 30 minutes of structure learning, and when $m=10$; i.e., when all 10 threads are utilised. The raw BDeu scores generated by the algorithms can be found in the Appendix A.

\begin{table}[H]
\scriptsize
\centering
\begin{tabular}{cccccccccc}
\toprule
\tabincell{c}{ }  &  \tabincell{c}{Audio\\train} & \tabincell{c}{Jester\\train} & \tabincell{c}{Pumsb-star\\test} & \tabincell{c}{Kosarek\\train} &
\tabincell{c}{EachMovie\\train} &
\tabincell{c}{Reuters\\test} &
\tabincell{c}{Reuters\\train} &
\tabincell{c}{NewsGroup\\test} & \tabincell{c}{NewsGroup\\valid} \\\\
\hline
Nodes & 100     & 100     & 163 & 190 & 500 & 889     & 889     & 910 & 910   \\
Sample size & 15,000  & 9,000  & 2,452 & 33,375 & 4,525 & 1,540     & 6,532     & 3,764 & 3,764   \\
Preserved CPSs  & 7,343,077     & 10,307,532     &19,346,718 & 38,858,778 & 21,985,307 & 10,487,832     & 37,479,789     & 34,544,307 &  41,020,532   \\
\bottomrule
\end{tabular}
\caption{The nine real-world data sets considered for evaluation.}
\end{table}


\begin{table}[H]
\scriptsize
\centering
\begin{tabular}{cccccccccc}
\toprule
\tabincell{c}{PS-MINOBS\\Hyperparameters} & Threads & 0.5h & 1h & 1.5h & 2h & 2.5h & 3h & 3.5h & 4h\\
\hline
\multirow{11}{*}{\tabincell{c}{$p=10\%$\\$m=10$}} & S$_{1}$ &  -0.587\textperthousand &  -0.589\textperthousand &  -0.593\textperthousand &   0.000\textperthousand &   0.000\textperthousand &   0.020\textperthousand &   0.023\textperthousand &   0.023\textperthousand \\
&S$_{2}$ &   0.031\textperthousand &   0.032\textperthousand &   0.026\textperthousand &   0.026\textperthousand &   0.026\textperthousand &   0.026\textperthousand &   0.026\textperthousand &   0.026\textperthousand \\
&S$_{3}$ &  -0.024\textperthousand &  -0.024\textperthousand &  -0.014\textperthousand &  -0.014\textperthousand &  -0.014\textperthousand &  -0.014\textperthousand &  -0.014\textperthousand &  -0.014\textperthousand \\
&S$_{4}$ &   0.027\textperthousand &   0.025\textperthousand &   0.019\textperthousand &   0.019\textperthousand &   0.019\textperthousand &   0.019\textperthousand &   0.019\textperthousand &   0.019\textperthousand \\
&S$_{5}$ &  -0.025\textperthousand &  -0.027\textperthousand &  -0.033\textperthousand &  -0.033\textperthousand &  -0.033\textperthousand &  -0.033\textperthousand &  -0.033\textperthousand &  -0.033\textperthousand \\
&S$_{6}$ &   0.004\textperthousand &   0.002\textperthousand &   0.003\textperthousand &   0.023\textperthousand &   0.023\textperthousand &   0.023\textperthousand &   0.023\textperthousand &   0.023\textperthousand \\
&S$_{7}$ &   0.022\textperthousand &   0.020\textperthousand &   0.020\textperthousand &   0.023\textperthousand &   0.023\textperthousand &   0.023\textperthousand &   0.023\textperthousand &   0.023\textperthousand \\
&S$_{8}$ &   0.024\textperthousand &   0.024\textperthousand &   0.018\textperthousand &   0.018\textperthousand &   0.018\textperthousand &   0.018\textperthousand &   0.018\textperthousand &   0.018\textperthousand \\
&S$_{9}$ &  -0.003\textperthousand &  -0.005\textperthousand &  -0.011\textperthousand &   0.006\textperthousand &   0.006\textperthousand &   0.006\textperthousand &   0.006\textperthousand &   0.006\textperthousand \\
&S$_{10}$ &  -0.007\textperthousand &  -0.009\textperthousand &  -0.015\textperthousand &  -0.015\textperthousand &  -0.015\textperthousand &  -0.015\textperthousand &  -0.015\textperthousand &  -0.015\textperthousand \\
&Highest DAG $\Delta$ & \textbf{0.031\textperthousand} & \textbf{0.032\textperthousand} & \textbf{0.026\textperthousand} & \textbf{0.026\textperthousand} & \textbf{0.026\textperthousand} & \textbf{0.026\textperthousand} & \textbf{0.026\textperthousand} & \textbf{0.026\textperthousand} \\
\hline
\multirow{6}{*}{\tabincell{c}{$p=20\%$\\$m=5$}} & S$_{1}$ &   0.010\textperthousand &   0.008\textperthousand &   0.002\textperthousand &   0.002\textperthousand &   0.002\textperthousand &   0.002\textperthousand &   0.002\textperthousand &   0.002\textperthousand \\
& S$_{2}$ &   0.019\textperthousand &   0.028\textperthousand &   0.022\textperthousand &   0.022\textperthousand &   0.022\textperthousand &   0.022\textperthousand &   0.022\textperthousand &   0.022\textperthousand \\
& S$_{3}$ &   0.010\textperthousand &   0.008\textperthousand &   0.002\textperthousand &   0.002\textperthousand &   0.002\textperthousand &   0.002\textperthousand &   0.002\textperthousand &   0.002\textperthousand \\
& S$_{4}$ &   0.008\textperthousand &   0.008\textperthousand &   0.002\textperthousand &   0.002\textperthousand &   0.022\textperthousand &   0.022\textperthousand &   0.022\textperthousand &   0.022\textperthousand \\
& S$_{5}$ &   0.030\textperthousand &   0.028\textperthousand &   0.022\textperthousand &   0.022\textperthousand &   0.022\textperthousand &   0.022\textperthousand &   0.022\textperthousand &   0.022\textperthousand \\
& Highest DAG $\Delta$ &   \textbf{0.030\textperthousand} &  \textbf{0.028\textperthousand} &   \textbf{0.022\textperthousand} &   \textbf{0.022\textperthousand} &   \textbf{0.022\textperthousand} &   \textbf{0.022\textperthousand} &   \textbf{0.022\textperthousand} &   \textbf{0.022\textperthousand} \\
\hline
\multirow{5}{*}{\tabincell{c}{$p=25\%$\\$m=4$}} & S1 &   0.030\textperthousand &   0.028\textperthousand &   0.022\textperthousand &   0.022\textperthousand &   0.022\textperthousand &   0.022\textperthousand &   0.022\textperthousand &   0.022\textperthousand \\
& S$_{2}$ &   0.010\textperthousand &   0.028\textperthousand &   0.022\textperthousand &   0.022\textperthousand &   0.022\textperthousand &   0.022\textperthousand &   0.022\textperthousand &   0.022\textperthousand \\
& S$_{3}$ &   0.030\textperthousand &   0.028\textperthousand &   0.022\textperthousand &   0.022\textperthousand &   0.022\textperthousand &   0.022\textperthousand &   0.022\textperthousand &   0.022\textperthousand \\
& S$_{4}$ &  -0.027\textperthousand &  -0.027\textperthousand &  -0.033\textperthousand &  -0.033\textperthousand &  -0.033\textperthousand &  -0.033\textperthousand &  -0.033\textperthousand &  -0.033\textperthousand \\
& Highest DAG $\Delta$ &   \textbf{0.030\textperthousand} &   \textbf{0.028\textperthousand} &   \textbf{0.022\textperthousand} &   \textbf{0.022\textperthousand} &   \textbf{0.022\textperthousand} &   \textbf{0.022\textperthousand} &   \textbf{0.022\textperthousand} &   \textbf{0.022\textperthousand} \\
\hline
\multirow{4}{*}{\tabincell{c}{$p=33\%$\\$m=3$}} & S$_{1}$&   0.010\textperthousand &   0.008\textperthousand &   0.002\textperthousand &   0.002\textperthousand &   0.022\textperthousand &   0.022\textperthousand &   0.022\textperthousand &   0.022\textperthousand \\
& S$_{2}$ &   0.030\textperthousand &   0.028\textperthousand &   0.022\textperthousand &   0.022\textperthousand &   0.022\textperthousand &   0.022\textperthousand &   0.022\textperthousand &   0.022\textperthousand \\
& S$_{3}$ &  -0.041\textperthousand &  -0.043\textperthousand &  -0.049\textperthousand &  -0.049\textperthousand &  -0.049\textperthousand &  -0.049\textperthousand &  -0.049\textperthousand &  -0.049\textperthousand \\
& Highest DAG $\Delta$ &   \textbf{0.030\textperthousand} &   \textbf{0.028\textperthousand} &   \textbf{0.022\textperthousand} &   \textbf{0.022\textperthousand} &   \textbf{0.022\textperthousand} &   \textbf{0.022\textperthousand} &   \textbf{0.022\textperthousand} &   \textbf{0.022\textperthousand} \\
\hline
\multirow{3}{*}{\tabincell{c}{$p=50\%$\\$m=2$}} & S$_{1}$&   0.028\textperthousand &   0.026\textperthousand &   0.020\textperthousand &   0.022\textperthousand &   0.022\textperthousand &   0.022\textperthousand &   0.022\textperthousand &   0.022\textperthousand \\
& S$_{2}$ &  -0.003\textperthousand &  -0.003\textperthousand &  -0.009\textperthousand &  -0.009\textperthousand &  -0.009\textperthousand &  -0.009\textperthousand &  -0.009\textperthousand &  -0.009\textperthousand \\
& Highest DAG $\Delta$ &   \textbf{0.028\textperthousand} &   \textbf{0.026\textperthousand} &   \textbf{0.020\textperthousand} &   \textbf{0.022\textperthousand} &   \textbf{0.022\textperthousand} &   \textbf{0.022\textperthousand} &   \textbf{0.022\textperthousand} &   \textbf{0.022\textperthousand} \\
\hline
\multirow{3}{*}{\tabincell{c}{$p=80\%$\\$m=2$}} & S$_{1}$&   0.008\textperthousand &   0.008\textperthousand &   0.002\textperthousand &   0.002\textperthousand &   0.002\textperthousand &   0.002\textperthousand &   0.002\textperthousand &   0.002\textperthousand \\
& S$_{2}$ &  -0.044\textperthousand &  -0.040\textperthousand &  -0.044\textperthousand &  -0.044\textperthousand &  -0.044\textperthousand &  -0.044\textperthousand &  -0.044\textperthousand &  -0.044\textperthousand \\
& Highest DAG $\Delta$ &   \textbf{0.008\textperthousand} &   \textbf{0.008\textperthousand} &   \textbf{0.002\textperthousand} &   \textbf{0.002\textperthousand} &   \textbf{0.002\textperthousand} &   \textbf{0.002\textperthousand} &   \textbf{0.002\textperthousand} &   \textbf{0.002\textperthousand} \\
\bottomrule
\end{tabular}
\caption{The discrepancy $\Delta$ in the BDeu scores of PS-MINOBS and MINOBS, where a positive discrepancy indicates a better performance for PS-MINOBS. The results are based on case study Audio-train, and are shown across different hyperparameter settings $p$ and $m$ for PS-MINOBS. The learning runtime is restricted to four hours, and the results are depicted at each 30-minute interval.}
\end{table}


\begin{table}[H]
\scriptsize
\centering
\begin{tabular}{cccccccccc}
\toprule
 \tabincell{c}{PS-MINOBS\\Hyperparameters} & Threads & 0.5h & 1h & 1.5h & 2h & 2.5h & 3h & 3.5h & 4h\\
\hline
\multirow{11}{*}{\tabincell{c}{$p=10\%$\\$m=10$}} & S$_{1}$&  -0.557\textperthousand &  -0.108\textperthousand &  -0.079\textperthousand &  -0.009\textperthousand &  -0.002\textperthousand &   0.008\textperthousand &   0.008\textperthousand &   0.008\textperthousand \\
&S$_{2}$ &   0.066\textperthousand &   0.041\textperthousand &   0.039\textperthousand &   0.036\textperthousand &   0.034\textperthousand &   0.034\textperthousand &   0.034\textperthousand &   0.034\textperthousand \\
&S$_{3}$ &   0.092\textperthousand &   0.067\textperthousand &   0.065\textperthousand &   0.062\textperthousand &   0.060\textperthousand &   0.060\textperthousand &   0.060\textperthousand &   0.060\textperthousand \\
&S$_{4}$ &   0.059\textperthousand &   0.035\textperthousand &   0.033\textperthousand &   0.029\textperthousand &   0.027\textperthousand &   0.027\textperthousand &   0.027\textperthousand &   0.027\textperthousand \\
&S$_{5}$ &   0.108\textperthousand &   0.084\textperthousand &   0.096\textperthousand &   0.093\textperthousand &   0.091\textperthousand &   0.091\textperthousand &   0.091\textperthousand &   0.091\textperthousand \\
&S$_{6}$ &  -0.554\textperthousand &  -0.578\textperthousand &  -0.576\textperthousand &  -0.569\textperthousand &  -0.571\textperthousand &  -0.571\textperthousand &  -0.571\textperthousand &  -0.571\textperthousand \\
&S$_{7}$ &   0.092\textperthousand &   0.067\textperthousand &   0.065\textperthousand &   0.062\textperthousand &   0.060\textperthousand &   0.060\textperthousand &   0.060\textperthousand &   0.060\textperthousand \\
&S$_{8}$ &   0.025\textperthousand &   0.001\textperthousand &  -0.002\textperthousand &  -0.005\textperthousand &  -0.007\textperthousand &  -0.007\textperthousand &  -0.007\textperthousand &  -0.007\textperthousand \\
&S$_{9}$ &   0.030\textperthousand &   0.027\textperthousand &   0.059\textperthousand &   0.087\textperthousand &   0.085\textperthousand &   0.085\textperthousand &   0.085\textperthousand &   0.085\textperthousand \\
&S$_{10}$ &   0.054\textperthousand &   0.041\textperthousand &   0.065\textperthousand &   0.062\textperthousand &   0.060\textperthousand &   0.060\textperthousand &   0.060\textperthousand &   0.060\textperthousand \\
&\tabincell{c}{Highest DAG $\Delta$} & \textbf{  0.108\textperthousand} & \textbf{  0.084\textperthousand} & \textbf{  0.096\textperthousand} & \textbf{  0.093\textperthousand} & \textbf{  0.091\textperthousand} & \textbf{  0.091\textperthousand} & \textbf{  0.091\textperthousand} & \textbf{  0.091\textperthousand} \\
\hline
\multirow{6}{*}{\tabincell{c}{$p=20\%$\\$m=5$}} & S$_{1}$&   0.048\textperthousand &   0.024\textperthousand &   0.022\textperthousand &   0.019\textperthousand &   0.016\textperthousand &   0.016\textperthousand &   0.037\textperthousand &   0.037\textperthousand \\
& S$_{2}$ &  -0.071\textperthousand &  -0.095\textperthousand &  -0.097\textperthousand &  -0.100\textperthousand &  -0.102\textperthousand &  -0.102\textperthousand &  -0.102\textperthousand &  -0.102\textperthousand \\
& S$_{3}$ &  -0.037\textperthousand &  -0.061\textperthousand &  -0.063\textperthousand &  -0.067\textperthousand &  -0.069\textperthousand &  -0.069\textperthousand &  -0.069\textperthousand &  -0.069\textperthousand \\
& S$_{4}$ &   0.106\textperthousand &   0.081\textperthousand &   0.100\textperthousand &   0.097\textperthousand &   0.095\textperthousand &   0.095\textperthousand &   0.095\textperthousand &   0.095\textperthousand \\
& S$_{5}$ &   0.118\textperthousand &   0.097\textperthousand &   0.095\textperthousand &   0.092\textperthousand &   0.089\textperthousand &   0.089\textperthousand &   0.089\textperthousand &   0.089\textperthousand \\
& Highest DAG $\Delta$ & \textbf{  0.118\textperthousand} & \textbf{  0.097\textperthousand} & \textbf{  0.100\textperthousand} & \textbf{  0.097\textperthousand} & \textbf{  0.095\textperthousand} & \textbf{  0.095\textperthousand} & \textbf{  0.095\textperthousand} & \textbf{  0.095\textperthousand} \\
\hline
\multirow{5}{*}{\tabincell{c}{$p=25\%$\\$m=4$}} & S$_{1}$&   0.063\textperthousand &   0.038\textperthousand &   0.036\textperthousand &   0.033\textperthousand &   0.031\textperthousand &   0.031\textperthousand &   0.031\textperthousand &   0.031\textperthousand \\
& S$_{2}$ &   0.127\textperthousand &   0.103\textperthousand &   0.100\textperthousand &   0.097\textperthousand &   0.095\textperthousand &   0.095\textperthousand &   0.095\textperthousand &   0.095\textperthousand \\
& S$_{3}$ &   0.083\textperthousand &   0.100\textperthousand &   0.100\textperthousand &   0.097\textperthousand &   0.095\textperthousand &   0.095\textperthousand &   0.095\textperthousand &   0.095\textperthousand \\
& S$_{4}$ &   0.025\textperthousand &   0.062\textperthousand &   0.060\textperthousand &   0.057\textperthousand &   0.054\textperthousand &   0.054\textperthousand &   0.054\textperthousand &   0.054\textperthousand \\
& Highest DAG $\Delta$ & \textbf{  0.127\textperthousand} & \textbf{  0.103\textperthousand} & \textbf{  0.100\textperthousand} & \textbf{  0.097\textperthousand} & \textbf{  0.095\textperthousand} & \textbf{  0.095\textperthousand} & \textbf{  0.095\textperthousand} & \textbf{  0.095\textperthousand} \\
\hline
\multirow{4}{*}{\tabincell{c}{$p=33\%$\\$m=3$}} & S$_{1}$ &   0.127\textperthousand &   0.103\textperthousand &   0.100\textperthousand &   0.097\textperthousand &   0.095\textperthousand &   0.095\textperthousand &   0.095\textperthousand &   0.095\textperthousand \\
& S$_{2}$ &   0.043\textperthousand &   0.072\textperthousand &   0.070\textperthousand &   0.067\textperthousand &   0.064\textperthousand &   0.064\textperthousand &   0.064\textperthousand &   0.064\textperthousand \\
& S$_{3}$ &   0.023\textperthousand &   0.034\textperthousand &   0.033\textperthousand &   0.030\textperthousand &   0.028\textperthousand &   0.028\textperthousand &   0.028\textperthousand &   0.028\textperthousand \\
& Highest DAG $\Delta$ & \textbf{  0.127\textperthousand} & \textbf{  0.103\textperthousand} & \textbf{  0.100\textperthousand} & \textbf{  0.097\textperthousand} & \textbf{  0.095\textperthousand} & \textbf{  0.095\textperthousand} & \textbf{  0.095\textperthousand} & \textbf{  0.095\textperthousand} \\
\hline
\multirow{3}{*}{\tabincell{c}{$p=50\%$\\$m=2$}} & S$_{1}$ &   0.067\textperthousand &   0.045\textperthousand &   0.042\textperthousand &   0.039\textperthousand &   0.037\textperthousand &   0.037\textperthousand &   0.037\textperthousand &   0.037\textperthousand \\
& S$_{2}$ &   0.024\textperthousand &   0.020\textperthousand &   0.018\textperthousand &   0.015\textperthousand &   0.012\textperthousand &   0.012\textperthousand &   0.012\textperthousand &   0.012\textperthousand \\
& Highest DAG $\Delta$ & \textbf{  0.067\textperthousand} & \textbf{  0.045\textperthousand} & \textbf{  0.042\textperthousand} & \textbf{  0.039\textperthousand} & \textbf{  0.037\textperthousand} & \textbf{  0.037\textperthousand} & \textbf{  0.037\textperthousand} & \textbf{  0.037\textperthousand} \\
\hline
\multirow{3}{*}{\tabincell{c}{$p=80\%$\\$m=2$}} & S$_{1}$ &   0.012\textperthousand &  -0.004\textperthousand &   0.036\textperthousand &   0.033\textperthousand &   0.037\textperthousand &   0.037\textperthousand &   0.037\textperthousand &   0.037\textperthousand \\
& S$_{2}$ &   0.052\textperthousand &   0.045\textperthousand &   0.042\textperthousand &   0.039\textperthousand &   0.037\textperthousand &   0.037\textperthousand &   0.037\textperthousand &   0.037\textperthousand \\
& Highest DAG $\Delta$ & \textbf{  0.052\textperthousand} & \textbf{  0.045\textperthousand} & \textbf{  0.042\textperthousand} & \textbf{  0.039\textperthousand} & \textbf{  0.037\textperthousand} & \textbf{  0.037\textperthousand} & \textbf{  0.037\textperthousand} & \textbf{  0.037\textperthousand} \\
\bottomrule
\end{tabular}
\caption{The discrepancy $\Delta$ in the BDeu scores of PS-MINOBS and MINOBS, where a positive discrepancy indicates a better performance for PS-MINOBS. The results are based on case study Jester-train, and are shown across different hyperparameter settings $p$ and $m$ for PS-MINOBS. The learning runtime is restricted to four hours, and the results are depicted at each 30-minute interval.}
\end{table}


\begin{table}[H]
\scriptsize
\centering
\begin{tabular}{cccccccccc}
\toprule
 \tabincell{c}{PS-MINOBS\\Hyperparameters} & Threads & 0.5h & 1h & 1.5h & 2h & 2.5h & 3h & 3.5h & 4h\\
\hline
\multirow{11}{*}{\tabincell{c}{$p=10\%$\\$m=10$}} & S$_{1}$ &  -8.138\textperthousand &  -8.264\textperthousand &  -8.333\textperthousand &  -8.342\textperthousand &  -7.779\textperthousand &  -8.148\textperthousand &  -8.116\textperthousand &  -8.169\textperthousand \\
&S$_{2}$ &  -5.797\textperthousand &  -5.677\textperthousand &  -5.722\textperthousand &  -5.744\textperthousand &  -5.820\textperthousand &  -6.328\textperthousand &  -6.328\textperthousand &  -6.383\textperthousand \\
&S$_{3}$ &  -5.975\textperthousand &  -5.778\textperthousand &  -5.792\textperthousand &  -5.807\textperthousand &  -5.883\textperthousand &  -6.391\textperthousand &  -6.391\textperthousand &  -6.428\textperthousand \\
&S$_{4}$ &  -6.506\textperthousand &  -6.254\textperthousand &  -6.088\textperthousand &  -6.103\textperthousand &  -6.127\textperthousand &  -6.630\textperthousand &  -6.625\textperthousand &  -6.656\textperthousand \\
&S$_{5}$ &  -6.114\textperthousand &  -5.572\textperthousand &  -5.537\textperthousand &  -5.552\textperthousand &  -5.595\textperthousand &  -6.100\textperthousand &  -6.096\textperthousand &  -6.122\textperthousand \\
&S$_{6}$ &  -6.902\textperthousand &  -6.996\textperthousand &  -6.973\textperthousand &  -6.931\textperthousand &  -7.007\textperthousand &  -7.512\textperthousand &  -7.512\textperthousand &  -7.567\textperthousand \\
&S$_{7}$ &  -6.015\textperthousand &  -5.892\textperthousand &  -5.882\textperthousand &  -5.898\textperthousand &  -5.936\textperthousand &  -6.444\textperthousand &  -6.444\textperthousand &  -6.448\textperthousand \\
&S$_{8}$ &  -6.014\textperthousand &  -5.688\textperthousand &  -5.695\textperthousand &  -5.693\textperthousand &  -5.743\textperthousand &  -6.115\textperthousand &  -6.008\textperthousand &  -5.927\textperthousand \\
&S$_{9}$ &  -5.740\textperthousand &  -5.715\textperthousand &  -5.790\textperthousand &  -5.808\textperthousand &  -5.884\textperthousand &  -6.249\textperthousand &  -6.242\textperthousand &  -6.297\textperthousand \\
&S$_{10}$ &  -6.300\textperthousand &  -5.870\textperthousand &  -5.605\textperthousand &  -5.628\textperthousand &  -5.686\textperthousand &  -6.149\textperthousand &  -6.149\textperthousand &  -6.200\textperthousand \\
&\tabincell{c}{Highest DAG $\Delta$} & \textbf{  -5.740\textperthousand} & \textbf{ -5.572\textperthousand} & \textbf{ -5.537\textperthousand} & \textbf{ -5.552\textperthousand} & \textbf{ -5.595\textperthousand} & \textbf{ -6.100\textperthousand} & \textbf{ -6.008\textperthousand} & \textbf{ -6.122\textperthousand} \\
\hline
\multirow{6}{*}{\tabincell{c}{$p=20\%$\\$m=5$}} & S$_{1}$ & -4.076\textperthousand & -4.115\textperthousand &  -4.087\textperthousand &  -4.090\textperthousand &  -4.165\textperthousand &  -4.673\textperthousand & -4.671\textperthousand &  -4.719\textperthousand \\
& S$_{2}$ & -1.884\textperthousand & -1.865\textperthousand &  -1.880\textperthousand & -1.901\textperthousand & -1.977\textperthousand &  -2.483\textperthousand & -2.483\textperthousand & -2.537\textperthousand \\
& S$_{3}$ & -2.342\textperthousand & -2.132\textperthousand &  -1.986\textperthousand & -1.989\textperthousand & -2.025\textperthousand &  -2.530\textperthousand & -2.530\textperthousand & -2.585\textperthousand \\
& S$_{4}$ & -1.684\textperthousand & -1.749\textperthousand &  -1.798\textperthousand & -1.820\textperthousand & -1.865\textperthousand &  -2.364\textperthousand & -2.364\textperthousand & -2.415\textperthousand \\
& S$_{5}$ & -2.661\textperthousand & -2.503\textperthousand &  -2.529\textperthousand & -2.390\textperthousand & -2.463\textperthousand &  -2.931\textperthousand & -2.930\textperthousand & -2.985\textperthousand \\
& Highest DAG $\Delta$ & \textbf{  -1.684\textperthousand} & \textbf{ -1.749\textperthousand} & \textbf{ -1.798\textperthousand} & \textbf{ -1.820\textperthousand} & \textbf{ -1.865\textperthousand} & \textbf{ -2.364\textperthousand} & \textbf{ -2.364\textperthousand} & \textbf{ -2.415\textperthousand} \\
\hline
\multirow{5}{*}{\tabincell{c}{$p=25\%$\\$m=4$}} & S$_{1}$ &  -2.704\textperthousand &  -1.816\textperthousand &  -1.805\textperthousand &  -1.753\textperthousand &  -1.828\textperthousand &  -2.334\textperthousand &  -2.326\textperthousand &  -2.381\textperthousand \\
& S$_{2}$ &  -1.010\textperthousand &  -1.109\textperthousand &  -1.131\textperthousand &  -1.154\textperthousand &  -1.229\textperthousand &  -1.735\textperthousand &  -1.735\textperthousand &  -1.786\textperthousand \\
& S$_{3}$ &  -1.325\textperthousand &  -1.284\textperthousand &  -1.214\textperthousand &  -1.205\textperthousand &  -1.252\textperthousand &  -1.736\textperthousand &  -1.732\textperthousand &  -1.787\textperthousand \\
& S$_{4}$ &  -1.356\textperthousand &  -1.359\textperthousand &  -1.377\textperthousand &  -1.382\textperthousand &  -1.449\textperthousand &  -1.910\textperthousand &  -1.910\textperthousand &  -1.962\textperthousand \\
& Highest DAG $\Delta$ & \textbf{  -1.010\textperthousand} & \textbf{ -1.109\textperthousand} & \textbf{ -1.131\textperthousand} & \textbf{ -1.154\textperthousand} & \textbf{ -1.229\textperthousand} & \textbf{ -1.735\textperthousand} & \textbf{ -1.732\textperthousand} & \textbf{ -1.786\textperthousand} \\
\hline
\multirow{4}{*}{\tabincell{c}{$p=33\%$\\$m=3$}} & S$_{1}$ &  -1.688\textperthousand &  -0.980\textperthousand &  -0.682\textperthousand &  -0.634\textperthousand &  -0.688\textperthousand &  -1.192\textperthousand &  -1.192\textperthousand &  -1.247\textperthousand \\
& S$_{2}$ &  -0.881\textperthousand &  -0.703\textperthousand &  -0.785\textperthousand &  -0.802\textperthousand &  -0.839\textperthousand &  -1.344\textperthousand &  -1.311\textperthousand &  -1.365\textperthousand \\
& S$_{3}$ &  -0.486\textperthousand &  -0.432\textperthousand &  -0.498\textperthousand &  -0.520\textperthousand &  -0.555\textperthousand &  -1.056\textperthousand &  -1.056\textperthousand &  -1.111\textperthousand \\
& Highest DAG $\Delta$ & \textbf{  -0.486\textperthousand} & \textbf{ -0.432\textperthousand} & \textbf{ -0.498\textperthousand} & \textbf{ -0.520\textperthousand} & \textbf{ -0.555\textperthousand} & \textbf{ -1.056\textperthousand} & \textbf{ -1.056\textperthousand} & \textbf{ -1.111\textperthousand} \\
\hline
\multirow{3}{*}{\tabincell{c}{$p=50\%$\\$m=2$}} & S$_{1}$ &  -0.460\textperthousand &  -0.198\textperthousand &  -0.179\textperthousand &  -0.034\textperthousand &  -0.109\textperthousand &  -0.614\textperthousand &  -0.614\textperthousand &  -0.669\textperthousand \\
& S$_{2}$ &  -0.004\textperthousand &   0.576\textperthousand &   0.616\textperthousand &   0.594\textperthousand &   0.519\textperthousand &   0.031\textperthousand &   0.034\textperthousand &  -0.020\textperthousand \\
& Highest DAG $\Delta$ & \textbf{  -0.004\textperthousand} & \textbf{  0.576\textperthousand} & \textbf{  0.616\textperthousand} & \textbf{  0.594\textperthousand} & \textbf{  0.519\textperthousand} & \textbf{  0.031\textperthousand} & \textbf{  0.034\textperthousand} & \textbf{ -0.020\textperthousand} \\
\hline
\multirow{3}{*}{\tabincell{c}{$p=80\%$\\$m=2$}} & S$_{1}$ &   0.200\textperthousand &   0.594\textperthousand &   0.585\textperthousand &   0.578\textperthousand &   0.502\textperthousand &  -0.001\textperthousand &  -0.001\textperthousand &  -0.056\textperthousand \\
& S$_{2}$ &   0.220\textperthousand &   0.576\textperthousand &   0.559\textperthousand &   0.597\textperthousand &   0.550\textperthousand &   0.045\textperthousand &   0.046\textperthousand &  -0.005\textperthousand \\
& Highest DAG $\Delta$ & \textbf{  0.220\textperthousand} & \textbf{  0.594\textperthousand} & \textbf{  0.585\textperthousand} & \textbf{  0.597\textperthousand} & \textbf{  0.550\textperthousand} & \textbf{  0.045\textperthousand} & \textbf{  0.046\textperthousand} & \textbf{ -0.005\textperthousand} \\
\bottomrule
\end{tabular}
\caption{The discrepancy $\Delta$ in the BDeu scores of PS-MINOBS and MINOBS, where a positive discrepancy indicates a better performance for PS-MINOBS. The results are based on case study Pumsb-star-test, and are shown across different hyperparameter settings $p$ and $m$ for PS-MINOBS. The learning runtime is restricted to four hours, and the results are depicted at each 30-minute interval.}
\end{table}


\begin{table}[H]
\scriptsize
\centering
\begin{tabular}{cccccccccc}
\toprule
 \tabincell{c}{PS-MINOBS\\Hyperparameters} & Threads & 0.5h & 1h & 1.5h & 2h & 2.5h & 3h & 3.5h & 4h\\
\hline
\multirow{11}{*}{\tabincell{c}{$p=10\%$\\$m=10$}}&S$_{1}$ &   0.077\textperthousand &   0.017\textperthousand &  -0.008\textperthousand &  -0.004\textperthousand &  -0.004\textperthousand &  -0.004\textperthousand &  -0.004\textperthousand &  -0.004\textperthousand \\
&S$_{2}$ &   0.047\textperthousand &  -0.013\textperthousand &  -0.037\textperthousand &  -0.037\textperthousand &  -0.037\textperthousand &  -0.037\textperthousand &  -0.037\textperthousand &  -0.037\textperthousand \\
&S$_{3}$ &   0.041\textperthousand &  -0.019\textperthousand &  -0.044\textperthousand &  -0.044\textperthousand &  -0.044\textperthousand &  -0.044\textperthousand &  -0.044\textperthousand &  -0.044\textperthousand \\
&S$_{4}$ &   0.090\textperthousand &   0.030\textperthousand &   0.006\textperthousand &   0.006\textperthousand &   0.006\textperthousand &   0.006\textperthousand &   0.006\textperthousand &   0.006\textperthousand \\
&S$_{5}$ &   0.081\textperthousand &   0.021\textperthousand &  -0.004\textperthousand &  -0.004\textperthousand &  -0.004\textperthousand &  -0.004\textperthousand &  -0.004\textperthousand &  -0.004\textperthousand \\
&S$_{6}$ &   0.039\textperthousand &  -0.013\textperthousand &  -0.004\textperthousand &  -0.004\textperthousand &  -0.004\textperthousand &  -0.004\textperthousand &  -0.004\textperthousand &  -0.004\textperthousand \\
&S$_{7}$ &   0.035\textperthousand &  -0.016\textperthousand &  -0.018\textperthousand &  -0.018\textperthousand &  -0.018\textperthousand &  -0.018\textperthousand &  -0.018\textperthousand &  -0.018\textperthousand \\
&S$_{8}$ &   0.057\textperthousand &   0.005\textperthousand &  -0.020\textperthousand &  -0.020\textperthousand &  -0.020\textperthousand &  -0.014\textperthousand &  -0.011\textperthousand &  -0.011\textperthousand \\
&S$_{9}$ &   0.064\textperthousand &   0.004\textperthousand &  -0.012\textperthousand &  -0.012\textperthousand &  -0.012\textperthousand &  -0.012\textperthousand &  -0.012\textperthousand &  -0.004\textperthousand \\
&S$_{10}$ &   0.064\textperthousand &   0.004\textperthousand &  -0.021\textperthousand &  -0.021\textperthousand &  -0.012\textperthousand &  -0.012\textperthousand &  -0.012\textperthousand &  -0.012\textperthousand \\
& Highest DAG $\Delta$ & \textbf{  0.090\textperthousand} & \textbf{  0.030\textperthousand} & \textbf{  0.006\textperthousand} & \textbf{  0.006\textperthousand} & \textbf{  0.006\textperthousand} & \textbf{  0.006\textperthousand} & \textbf{  0.006\textperthousand} & \textbf{  0.006\textperthousand} \\
\hline
\multirow{6}{*}{\tabincell{c}{$p=20\%$\\$m=5$}} & S$_{1}$ &   0.082\textperthousand &   0.030\textperthousand &   0.006\textperthousand &   0.006\textperthousand &   0.006\textperthousand &   0.006\textperthousand &   0.006\textperthousand &   0.006\textperthousand \\
& S$_{2}$ &   0.082\textperthousand &   0.022\textperthousand &  -0.003\textperthousand &  -0.003\textperthousand &   0.006\textperthousand &   0.006\textperthousand &   0.006\textperthousand &   0.006\textperthousand \\
& S$_{3}$ &   0.078\textperthousand &   0.025\textperthousand &   0.004\textperthousand &   0.004\textperthousand &   0.004\textperthousand &   0.004\textperthousand &   0.004\textperthousand &   0.004\textperthousand \\
& S$_{4}$ &   0.030\textperthousand &   0.013\textperthousand &  -0.004\textperthousand &  -0.004\textperthousand &  -0.004\textperthousand &  -0.004\textperthousand &  -0.004\textperthousand &  -0.004\textperthousand \\
& S$_{5}$ &   0.082\textperthousand &   0.022\textperthousand &  -0.003\textperthousand &  -0.003\textperthousand &  -0.003\textperthousand &   0.006\textperthousand &   0.006\textperthousand &   0.006\textperthousand \\
& Highest DAG $\Delta$ & \textbf{  0.082\textperthousand} & \textbf{  0.030\textperthousand} & \textbf{  0.006\textperthousand} & \textbf{  0.006\textperthousand} & \textbf{  0.006\textperthousand} & \textbf{  0.006\textperthousand} & \textbf{  0.006\textperthousand} & \textbf{  0.006\textperthousand} \\
\hline
\multirow{5}{*}{\tabincell{c}{$p=25\%$\\$m=4$}} & S$_{1}$ &   0.072\textperthousand &   0.022\textperthousand &  -0.003\textperthousand &  -0.003\textperthousand &  -0.003\textperthousand &  -0.003\textperthousand &  -0.003\textperthousand &  -0.003\textperthousand \\
& S$_{2}$ &   0.073\textperthousand &   0.014\textperthousand &  -0.003\textperthousand &  -0.003\textperthousand &  -0.003\textperthousand &  -0.003\textperthousand &  -0.003\textperthousand &  -0.003\textperthousand \\
& S$_{3}$ &   0.067\textperthousand &   0.016\textperthousand &  -0.001\textperthousand &  -0.001\textperthousand &  -0.001\textperthousand &  -0.001\textperthousand &  -0.001\textperthousand &  -0.001\textperthousand \\
& S$_{4}$ &   0.045\textperthousand &  -0.015\textperthousand &  -0.040\textperthousand &  -0.040\textperthousand &  -0.040\textperthousand &  -0.040\textperthousand &  -0.040\textperthousand &  -0.040\textperthousand \\
& Highest DAG $\Delta$ & \textbf{  0.073\textperthousand} & \textbf{  0.022\textperthousand} & \textbf{ -0.001\textperthousand} & \textbf{ -0.001\textperthousand} & \textbf{ -0.001\textperthousand} & \textbf{ -0.001\textperthousand} & \textbf{ -0.001\textperthousand} & \textbf{ -0.001\textperthousand} \\
\hline
\multirow{4}{*}{\tabincell{c}{$p=33\%$\\$m=3$}} & S$_{1}$ &   0.084\textperthousand &   0.030\textperthousand &   0.006\textperthousand &   0.006\textperthousand &   0.006\textperthousand &   0.006\textperthousand &   0.006\textperthousand &   0.006\textperthousand \\
& S$_{2}$ &   0.039\textperthousand &  -0.021\textperthousand &  -0.003\textperthousand &  -0.003\textperthousand &  -0.003\textperthousand &  -0.003\textperthousand &  -0.003\textperthousand &  -0.003\textperthousand \\
& S$_{3}$ &   0.086\textperthousand &   0.030\textperthousand &   0.006\textperthousand &   0.006\textperthousand &   0.006\textperthousand &   0.006\textperthousand &   0.006\textperthousand &   0.006\textperthousand \\
& Highest DAG $\Delta$ & \textbf{  0.086\textperthousand} & \textbf{  0.030\textperthousand} & \textbf{  0.006\textperthousand} & \textbf{  0.006\textperthousand} & \textbf{  0.006\textperthousand} & \textbf{  0.006\textperthousand} & \textbf{  0.006\textperthousand} & \textbf{  0.006\textperthousand} \\
\hline
\multirow{3}{*}{\tabincell{c}{$p=50\%$\\$m=2$}} & S$_{1}$ &   0.028\textperthousand &   0.015\textperthousand &  -0.003\textperthousand &  -0.003\textperthousand &  -0.003\textperthousand &  -0.003\textperthousand &  -0.003\textperthousand &  -0.003\textperthousand \\
& S$_{2}$ &   0.030\textperthousand &   0.010\textperthousand &  -0.014\textperthousand &  -0.010\textperthousand &  -0.010\textperthousand &  -0.010\textperthousand &  -0.010\textperthousand &  -0.010\textperthousand \\
& Highest DAG $\Delta$ & \textbf{  0.030\textperthousand} & \textbf{  0.015\textperthousand} & \textbf{ -0.003\textperthousand} & \textbf{ -0.003\textperthousand} & \textbf{ -0.003\textperthousand} & \textbf{ -0.003\textperthousand} & \textbf{ -0.003\textperthousand} & \textbf{ -0.003\textperthousand} \\
\hline
\multirow{3}{*}{\tabincell{c}{$p=80\%$\\$m=2$}} & S$_{1}$ &   0.009\textperthousand &   0.001\textperthousand &  -0.003\textperthousand &  -0.003\textperthousand &  -0.003\textperthousand &  -0.003\textperthousand &  -0.003\textperthousand &  -0.003\textperthousand \\
& S$_{2}$ &   0.005\textperthousand &  -0.021\textperthousand &  -0.003\textperthousand &  -0.003\textperthousand &  -0.003\textperthousand &  -0.003\textperthousand &  -0.003\textperthousand &  -0.003\textperthousand \\
& Highest DAG $\Delta$ & \textbf{  0.009\textperthousand} & \textbf{  0.001\textperthousand} & \textbf{ -0.003\textperthousand} & \textbf{ -0.003\textperthousand} & \textbf{ -0.003\textperthousand} & \textbf{ -0.003\textperthousand} & \textbf{ -0.003\textperthousand} & \textbf{ -0.003\textperthousand} \\
\bottomrule
\end{tabular}
\caption{The discrepancy $\Delta$ in the BDeu scores of PS-MINOBS and MINOBS, where a positive discrepancy indicates a better performance for PS-MINOBS. The results are based on case study Kosarek-train, and are shown across different hyperparameter settings $p$ and $m$ for PS-MINOBS. The learning runtime is restricted to four hours, and the results are depicted at each 30-minute interval.}
\end{table}


\begin{table}[H]
\scriptsize
\centering
\begin{tabular}{cccccccccc}
\toprule
\tabincell{c}{PS-MINOBS\\Hyperparameters} & Threads & 0.5h & 1h & 1.5h & 2h & 2.5h & 3h & 3.5h & 4h\\
\hline
\multirow{11}{*}{\tabincell{c}{$p=10\%$\\$m=10$}}& S$_{1}$ &   0.251\textperthousand &   0.176\textperthousand &   0.125\textperthousand &   0.091\textperthousand &   0.079\textperthousand &   0.068\textperthousand &   0.068\textperthousand &   0.069\textperthousand \\
&S$_{2}$ &   0.250\textperthousand &   0.168\textperthousand &   0.107\textperthousand &   0.074\textperthousand &   0.061\textperthousand &   0.050\textperthousand &   0.051\textperthousand &   0.062\textperthousand \\
&S$_{3}$ &   0.215\textperthousand &   0.130\textperthousand &   0.067\textperthousand &   0.034\textperthousand &   0.028\textperthousand &   0.040\textperthousand &   0.041\textperthousand &   0.041\textperthousand \\
&S$_{4}$ &   0.235\textperthousand &   0.152\textperthousand &   0.094\textperthousand &   0.061\textperthousand &   0.049\textperthousand &   0.038\textperthousand &   0.040\textperthousand &   0.040\textperthousand \\
&S$_{5}$ &   0.088\textperthousand &  -0.002\textperthousand &  -0.066\textperthousand &  -0.099\textperthousand &  -0.110\textperthousand &  -0.121\textperthousand &  -0.121\textperthousand &  -0.121\textperthousand \\
&S$_{6}$ &   0.270\textperthousand &   0.181\textperthousand &   0.118\textperthousand &   0.085\textperthousand &   0.072\textperthousand &   0.068\textperthousand &   0.069\textperthousand &   0.069\textperthousand \\
&S$_{7}$ &   0.124\textperthousand &   0.098\textperthousand &   0.054\textperthousand &   0.022\textperthousand &   0.018\textperthousand &   0.007\textperthousand &   0.007\textperthousand &   0.007\textperthousand \\
&S$_{8}$ &   0.240\textperthousand &   0.174\textperthousand &   0.118\textperthousand &   0.085\textperthousand &   0.073\textperthousand &   0.062\textperthousand &   0.062\textperthousand &   0.062\textperthousand \\
&S$_{9}$ &   0.262\textperthousand &   0.176\textperthousand &   0.111\textperthousand &   0.084\textperthousand &   0.073\textperthousand &   0.062\textperthousand &   0.062\textperthousand &   0.062\textperthousand \\
&S$_{10}$ &   0.269\textperthousand &   0.180\textperthousand &   0.117\textperthousand &   0.084\textperthousand &   0.074\textperthousand &   0.063\textperthousand &   0.063\textperthousand &   0.063\textperthousand \\
& Highest DAG $\Delta$ & \textbf{  0.270\textperthousand} & \textbf{  0.181\textperthousand} & \textbf{  0.125\textperthousand} & \textbf{  0.091\textperthousand} & \textbf{  0.079\textperthousand} & \textbf{  0.068\textperthousand} & \textbf{  0.069\textperthousand} & \textbf{  0.069\textperthousand} \\
\hline
\multirow{6}{*}{\tabincell{c}{$p=20\%$\\$m=5$}} & S$_{1}$ &   0.199\textperthousand &   0.168\textperthousand &   0.115\textperthousand &   0.085\textperthousand &   0.074\textperthousand &   0.063\textperthousand &   0.069\textperthousand &   0.069\textperthousand \\
& S$_{2}$ &   0.085\textperthousand &   0.000\textperthousand &   0.000\textperthousand &  -0.033\textperthousand &  -0.027\textperthousand &  -0.038\textperthousand &  -0.032\textperthousand &  -0.026\textperthousand \\
& S$_{3}$ &   0.208\textperthousand &   0.124\textperthousand &   0.074\textperthousand &   0.040\textperthousand &   0.028\textperthousand &   0.017\textperthousand &   0.021\textperthousand &   0.021\textperthousand \\
& S$_{4}$ &   0.223\textperthousand &   0.161\textperthousand &   0.097\textperthousand &   0.070\textperthousand &   0.059\textperthousand &   0.048\textperthousand &   0.060\textperthousand &   0.060\textperthousand \\
& S$_{5}$ &   0.201\textperthousand &   0.158\textperthousand &   0.094\textperthousand &   0.062\textperthousand &   0.050\textperthousand &   0.055\textperthousand &   0.060\textperthousand &   0.062\textperthousand \\
& Highest DAG $\Delta$ & \textbf{  0.223\textperthousand} & \textbf{  0.168\textperthousand} & \textbf{  0.115\textperthousand} & \textbf{  0.085\textperthousand} & \textbf{  0.074\textperthousand} & \textbf{  0.063\textperthousand} & \textbf{  0.069\textperthousand} & \textbf{  0.069\textperthousand} \\
\hline
\multirow{5}{*}{\tabincell{c}{$p=25\%$\\$m=4$}} & S$_{1}$ &   0.198\textperthousand &   0.138\textperthousand &   0.107\textperthousand &   0.089\textperthousand &   0.077\textperthousand &   0.068\textperthousand &   0.068\textperthousand &   0.068\textperthousand \\
& S$_{2}$ &   0.215\textperthousand &   0.174\textperthousand &   0.112\textperthousand &   0.087\textperthousand &   0.075\textperthousand &   0.065\textperthousand &   0.065\textperthousand &   0.064\textperthousand \\
& S$_{3}$ &   0.180\textperthousand &   0.106\textperthousand &   0.097\textperthousand &   0.078\textperthousand &   0.066\textperthousand &   0.057\textperthousand &   0.057\textperthousand &   0.057\textperthousand \\
& S$_{4}$ &   0.217\textperthousand &   0.154\textperthousand &   0.116\textperthousand &   0.098\textperthousand &   0.086\textperthousand &   0.075\textperthousand &   0.075\textperthousand &   0.075\textperthousand \\
& Highest DAG $\Delta$ & \textbf{  0.217\textperthousand} & \textbf{  0.174\textperthousand} & \textbf{  0.116\textperthousand} & \textbf{  0.098\textperthousand} & \textbf{  0.086\textperthousand} & \textbf{  0.075\textperthousand} & \textbf{  0.075\textperthousand} & \textbf{  0.075\textperthousand} \\
\hline
\multirow{4}{*}{\tabincell{c}{$p=33\%$\\$m=3$}} & S$_{1}$ &   0.203\textperthousand &   0.163\textperthousand &   0.099\textperthousand &   0.068\textperthousand &   0.055\textperthousand &   0.045\textperthousand &   0.059\textperthousand &   0.064\textperthousand \\
& S$_{2}$ &  -0.046\textperthousand &   0.073\textperthousand &   0.010\textperthousand &  -0.024\textperthousand &  -0.034\textperthousand &  -0.045\textperthousand &  -0.045\textperthousand &  -0.046\textperthousand \\
& S$_{3}$ &   0.244\textperthousand &   0.158\textperthousand &   0.094\textperthousand &   0.061\textperthousand &   0.065\textperthousand &   0.054\textperthousand &   0.054\textperthousand &   0.054\textperthousand \\
& Highest DAG $\Delta$ & \textbf{  0.244\textperthousand} & \textbf{  0.163\textperthousand} & \textbf{  0.099\textperthousand} & \textbf{  0.068\textperthousand} & \textbf{  0.065\textperthousand} & \textbf{  0.054\textperthousand} & \textbf{  0.059\textperthousand} & \textbf{  0.064\textperthousand} \\
\hline
\multirow{3}{*}{\tabincell{c}{$p=50\%$\\$m=2$}} & S$_{1}$ &   0.076\textperthousand &   0.144\textperthousand &   0.087\textperthousand &   0.054\textperthousand &   0.042\textperthousand &   0.032\textperthousand &   0.032\textperthousand &   0.031\textperthousand \\
& S$_{2}$ &  -0.014\textperthousand &   0.166\textperthousand &   0.115\textperthousand &   0.082\textperthousand &   0.070\textperthousand &   0.066\textperthousand &   0.068\textperthousand &   0.068\textperthousand \\
& Highest DAG $\Delta$ & \textbf{  0.076\textperthousand} & \textbf{  0.166\textperthousand} & \textbf{  0.115\textperthousand} & \textbf{  0.082\textperthousand} & \textbf{  0.070\textperthousand} & \textbf{  0.066\textperthousand} & \textbf{  0.068\textperthousand} & \textbf{  0.068\textperthousand} \\
\hline
\multirow{3}{*}{\tabincell{c}{$p=80\%$\\$m=2$}} & S$_{1}$ &  -0.270\textperthousand &   0.062\textperthousand &   0.079\textperthousand &   0.052\textperthousand &   0.052\textperthousand &   0.041\textperthousand &   0.041\textperthousand &   0.042\textperthousand \\
& S$_{2}$ &  -0.058\textperthousand &   0.122\textperthousand &   0.087\textperthousand &   0.061\textperthousand &   0.055\textperthousand &   0.044\textperthousand &   0.044\textperthousand &   0.044\textperthousand \\
& Highest DAG $\Delta$ & \textbf{  -0.058\textperthousand} & \textbf{  0.122\textperthousand} & \textbf{  0.087\textperthousand} & \textbf{  0.061\textperthousand} & \textbf{  0.055\textperthousand} & \textbf{  0.044\textperthousand} & \textbf{  0.044\textperthousand} & \textbf{  0.044\textperthousand} \\
\bottomrule
\end{tabular}
\caption{The discrepancy $\Delta$ in the BDeu scores of PS-MINOBS and MINOBS, where a positive discrepancy indicates a better performance for PS-MINOBS. The results are based on case study EachMovie-train, and are shown across different hyperparameter settings $p$ and $m$ for PS-MINOBS. The learning runtime is restricted to four hours, and the results are depicted at each 30-minute interval.}
\end{table}


\begin{table}[H]
\scriptsize
\centering
\begin{tabular}{cccccccccc}
\toprule
 \tabincell{c}{PS-MINOBS\\Hyperparameters} & Threads & 0.5h & 1h & 1.5h & 2h & 2.5h & 3h & 3.5h & 4h\\
\hline
\multirow{11}{*}{\tabincell{c}{$p=10\%$\\$m=10$}}& S$_{1}$ &   0.128\textperthousand &  -0.065\textperthousand &  -0.088\textperthousand &  -0.101\textperthousand &  -0.137\textperthousand &  -0.139\textperthousand &  -0.159\textperthousand &  -0.163\textperthousand \\
&S$_{2}$ &   0.160\textperthousand &  -0.041\textperthousand &  -0.066\textperthousand &  -0.083\textperthousand &  -0.106\textperthousand &  -0.107\textperthousand &  -0.127\textperthousand &  -0.131\textperthousand \\
&S$_{3}$ &   0.245\textperthousand &   0.054\textperthousand &   0.033\textperthousand &   0.030\textperthousand &   0.000\textperthousand &  -0.004\textperthousand &  -0.022\textperthousand &  -0.023\textperthousand \\
&S$_{4}$ &   0.088\textperthousand &  -0.123\textperthousand &  -0.121\textperthousand &  -0.117\textperthousand &  -0.147\textperthousand &  -0.134\textperthousand &  -0.151\textperthousand &  -0.155\textperthousand \\
&S$_{5}$ &   0.178\textperthousand &   0.041\textperthousand &   0.069\textperthousand &   0.068\textperthousand &   0.044\textperthousand &   0.040\textperthousand &   0.020\textperthousand &   0.016\textperthousand \\
&S$_{6}$ &   0.019\textperthousand &  -0.161\textperthousand &  -0.174\textperthousand &  -0.178\textperthousand &  -0.212\textperthousand &  -0.214\textperthousand &  -0.234\textperthousand &  -0.204\textperthousand \\
&S$_{7}$ &   0.146\textperthousand &  -0.015\textperthousand &  -0.027\textperthousand &  -0.022\textperthousand &  -0.039\textperthousand &  -0.043\textperthousand &  -0.063\textperthousand &  -0.064\textperthousand \\
&S$_{8}$ &   0.174\textperthousand &  -0.035\textperthousand &  -0.053\textperthousand &  -0.070\textperthousand &  -0.106\textperthousand &  -0.110\textperthousand &  -0.130\textperthousand &  -0.135\textperthousand \\
&S$_{9}$ &   0.115\textperthousand &  -0.086\textperthousand &  -0.064\textperthousand &  -0.084\textperthousand &  -0.101\textperthousand &  -0.106\textperthousand &  -0.126\textperthousand &  -0.128\textperthousand \\
&S$_{10}$ &   0.083\textperthousand &  -0.121\textperthousand &  -0.124\textperthousand &  -0.137\textperthousand &  -0.173\textperthousand &  -0.157\textperthousand &  -0.177\textperthousand &  -0.181\textperthousand \\
& Highest DAG $\Delta$ & \textbf{  0.245\textperthousand} & \textbf{  0.054\textperthousand} & \textbf{  0.069\textperthousand} & \textbf{  0.068\textperthousand} & \textbf{  0.044\textperthousand} & \textbf{  0.040\textperthousand} & \textbf{  0.020\textperthousand} & \textbf{  0.016\textperthousand} \\
\hline
\multirow{6}{*}{\tabincell{c}{$p=20\%$\\$m=5$}} & S$_{1}$ &   0.151\textperthousand &  -0.041\textperthousand &  -0.072\textperthousand &  -0.041\textperthousand &  -0.076\textperthousand &  -0.063\textperthousand &  -0.083\textperthousand &  -0.087\textperthousand \\
& S$_{2}$ &   0.273\textperthousand &   0.081\textperthousand &   0.050\textperthousand &   0.041\textperthousand &   0.005\textperthousand &   0.003\textperthousand &  -0.014\textperthousand &  -0.017\textperthousand \\
& S$_{3}$ &   0.275\textperthousand &   0.060\textperthousand &   0.033\textperthousand &   0.015\textperthousand &  -0.007\textperthousand &  -0.010\textperthousand &  -0.029\textperthousand &  -0.033\textperthousand \\
& S$_{4}$ &   0.268\textperthousand &   0.074\textperthousand &   0.047\textperthousand &   0.046\textperthousand &   0.011\textperthousand &   0.010\textperthousand &  -0.009\textperthousand &  -0.013\textperthousand \\
& S$_{5}$ &   0.268\textperthousand &   0.066\textperthousand &   0.049\textperthousand &   0.027\textperthousand &  -0.004\textperthousand &  -0.008\textperthousand &  -0.027\textperthousand &  -0.023\textperthousand \\
& Highest DAG $\Delta$ & \textbf{  0.275\textperthousand} & \textbf{  0.081\textperthousand} & \textbf{  0.050\textperthousand} & \textbf{  0.046\textperthousand} & \textbf{  0.011\textperthousand} & \textbf{  0.010\textperthousand} & \textbf{ -0.009\textperthousand} & \textbf{ -0.013\textperthousand} \\
\hline
\multirow{5}{*}{\tabincell{c}{$p=25\%$\\$m=4$}} & S$_{1}$ &   0.314\textperthousand &   0.113\textperthousand &   0.085\textperthousand &   0.063\textperthousand &   0.085\textperthousand &   0.082\textperthousand &   0.063\textperthousand &   0.059\textperthousand \\
& S$_{2}$ &   0.400\textperthousand &   0.202\textperthousand &   0.176\textperthousand &   0.159\textperthousand &   0.123\textperthousand &   0.121\textperthousand &   0.101\textperthousand &   0.097\textperthousand \\
& S$_{3}$ &   0.433\textperthousand &   0.247\textperthousand &   0.243\textperthousand &   0.222\textperthousand &   0.185\textperthousand &   0.181\textperthousand &   0.163\textperthousand &   0.173\textperthousand \\
& S$_{4}$ &   0.280\textperthousand &   0.070\textperthousand &   0.059\textperthousand &   0.036\textperthousand &   0.009\textperthousand &   0.039\textperthousand &   0.038\textperthousand &   0.038\textperthousand \\
& Highest DAG $\Delta$ & \textbf{  0.433\textperthousand} & \textbf{  0.247\textperthousand} & \textbf{  0.243\textperthousand} & \textbf{  0.222\textperthousand} & \textbf{  0.185\textperthousand} & \textbf{  0.181\textperthousand} & \textbf{  0.163\textperthousand} & \textbf{  0.173\textperthousand} \\
\hline
\multirow{4}{*}{\tabincell{c}{$p=33\%$\\$m=3$}} & S$_{1}$ &   0.266\textperthousand &   0.055\textperthousand &   0.029\textperthousand &   0.024\textperthousand &  -0.011\textperthousand &  -0.016\textperthousand &  -0.035\textperthousand &  -0.037\textperthousand \\
& S$_{2}$ &   0.357\textperthousand &   0.162\textperthousand &   0.143\textperthousand &   0.134\textperthousand &   0.146\textperthousand &   0.141\textperthousand &   0.124\textperthousand &   0.120\textperthousand \\
& S$_{3}$ &   0.270\textperthousand &   0.086\textperthousand &   0.093\textperthousand &   0.073\textperthousand &   0.039\textperthousand &   0.035\textperthousand &   0.015\textperthousand &   0.019\textperthousand \\
& Highest DAG $\Delta$ & \textbf{  0.357\textperthousand} & \textbf{  0.162\textperthousand} & \textbf{  0.143\textperthousand} & \textbf{  0.134\textperthousand} & \textbf{  0.146\textperthousand} & \textbf{  0.141\textperthousand} & \textbf{  0.124\textperthousand} & \textbf{  0.120\textperthousand} \\
\hline
\multirow{3}{*}{\tabincell{c}{$p=50\%$\\$m=2$}} & S$_{1}$ &   0.296\textperthousand &   0.090\textperthousand &   0.078\textperthousand &   0.067\textperthousand &   0.035\textperthousand &   0.031\textperthousand &   0.012\textperthousand &   0.008\textperthousand \\
& S$_{2}$ &   0.166\textperthousand &   0.012\textperthousand &   0.033\textperthousand &   0.014\textperthousand &   0.024\textperthousand &   0.020\textperthousand &   0.008\textperthousand &   0.005\textperthousand \\
& Highest DAG $\Delta$ & \textbf{  0.296\textperthousand} & \textbf{  0.090\textperthousand} & \textbf{  0.078\textperthousand} & \textbf{  0.067\textperthousand} & \textbf{  0.035\textperthousand} & \textbf{  0.031\textperthousand} & \textbf{  0.012\textperthousand} & \textbf{  0.008\textperthousand} \\
\hline
\multirow{3}{*}{\tabincell{c}{$p=80\%$\\$m=2$}} & S$_{1}$ &   0.175\textperthousand &   0.051\textperthousand &   0.025\textperthousand &   0.021\textperthousand &  -0.005\textperthousand &  -0.003\textperthousand &  -0.009\textperthousand &   0.015\textperthousand \\
& S$_{2}$ &   0.174\textperthousand &   0.028\textperthousand &   0.011\textperthousand &   0.004\textperthousand &  -0.032\textperthousand &  -0.031\textperthousand &  -0.028\textperthousand &  -0.016\textperthousand \\
& Highest DAG $\Delta$ & \textbf{  0.175\textperthousand} & \textbf{  0.051\textperthousand} & \textbf{  0.025\textperthousand} & \textbf{  0.021\textperthousand} & \textbf{ -0.005\textperthousand} & \textbf{ -0.003\textperthousand} & \textbf{ -0.009\textperthousand} & \textbf{  0.015\textperthousand} \\
\bottomrule
\end{tabular}
\caption{The discrepancy $\Delta$ in the BDeu scores of PS-MINOBS and MINOBS, where a positive discrepancy indicates a better performance for PS-MINOBS. The results are based on case study Reuters-test, and are shown across different hyperparameter settings $p$ and $m$ for PS-MINOBS. The learning runtime is restricted to four hours, and the results are depicted at each 30-minute interval.}
\end{table}

\begin{table}[H]
\scriptsize
\centering
\begin{tabular}{cccccccccc}
\toprule
  \tabincell{c}{PS-MINOBS\\Hyperparameters} & Threads & 0.5h & 1h & 1.5h & 2h & 2.5h & 3h & 3.5h & 4h\\
\hline
\multirow{11}{*}{\tabincell{c}{$p=10\%$\\$m=10$}} & S$_{1}$ &   0.082\textperthousand &   0.094\textperthousand &   0.025\textperthousand &   0.004\textperthousand &   0.018\textperthousand &   0.015\textperthousand &   0.015\textperthousand &   0.015\textperthousand \\
&S$_{2}$ &   0.149\textperthousand &   0.129\textperthousand &   0.064\textperthousand &   0.033\textperthousand &   0.021\textperthousand &   0.018\textperthousand &   0.018\textperthousand &   0.019\textperthousand \\
&S$_{3}$ &   0.092\textperthousand &   0.073\textperthousand &   0.007\textperthousand &  -0.021\textperthousand &  -0.030\textperthousand &  -0.033\textperthousand &  -0.033\textperthousand &  -0.033\textperthousand \\
&S$_{4}$ &   0.088\textperthousand &   0.085\textperthousand &   0.017\textperthousand &  -0.009\textperthousand &  -0.013\textperthousand &  -0.016\textperthousand &  -0.016\textperthousand &  -0.016\textperthousand \\
&S$_{5}$ &   0.065\textperthousand &   0.081\textperthousand &   0.015\textperthousand &  -0.013\textperthousand &  -0.001\textperthousand &  -0.006\textperthousand &  -0.006\textperthousand &  -0.006\textperthousand \\
&S$_{6}$ &   0.120\textperthousand &   0.098\textperthousand &   0.034\textperthousand &   0.005\textperthousand &  -0.008\textperthousand &  -0.013\textperthousand &  -0.013\textperthousand &  -0.012\textperthousand \\
&S$_{7}$ &   0.115\textperthousand &   0.119\textperthousand &   0.052\textperthousand &   0.022\textperthousand &   0.016\textperthousand &   0.017\textperthousand &   0.017\textperthousand &   0.020\textperthousand \\
&S$_{8}$ &   0.128\textperthousand &   0.094\textperthousand &   0.025\textperthousand &  -0.001\textperthousand &  -0.013\textperthousand &  -0.018\textperthousand &  -0.006\textperthousand &   0.011\textperthousand \\
&S$_{9}$ &   0.112\textperthousand &   0.121\textperthousand &   0.052\textperthousand &   0.021\textperthousand &   0.012\textperthousand &   0.007\textperthousand &   0.007\textperthousand &   0.008\textperthousand \\
&S$_{10}$ &   0.084\textperthousand &   0.057\textperthousand &  -0.012\textperthousand &  -0.042\textperthousand &  -0.054\textperthousand &  -0.059\textperthousand &  -0.059\textperthousand &  -0.059\textperthousand \\
& Highest DAG $\Delta$ & \textbf{  0.149\textperthousand} & \textbf{  0.129\textperthousand} & \textbf{  0.064\textperthousand} & \textbf{  0.033\textperthousand} & \textbf{  0.021\textperthousand} & \textbf{  0.018\textperthousand} & \textbf{  0.018\textperthousand} & \textbf{  0.020\textperthousand} \\
\hline
\multirow{6}{*}{\tabincell{c}{$p=20\%$\\$m=5$}} & S$_{1}$ &   0.112\textperthousand &   0.088\textperthousand &   0.019\textperthousand &  -0.008\textperthousand &  -0.017\textperthousand &  -0.022\textperthousand &  -0.016\textperthousand &   0.009\textperthousand \\
& S$_{2}$ &   0.092\textperthousand &   0.077\textperthousand &   0.008\textperthousand &   0.011\textperthousand &   0.018\textperthousand &   0.015\textperthousand &   0.016\textperthousand &   0.018\textperthousand \\
& S$_{3}$ &   0.105\textperthousand &   0.095\textperthousand &   0.030\textperthousand &   0.000\textperthousand &   0.042\textperthousand &   0.039\textperthousand &   0.039\textperthousand &   0.040\textperthousand \\
& S$_{4}$ &   0.132\textperthousand &   0.117\textperthousand &   0.050\textperthousand &   0.026\textperthousand &   0.012\textperthousand &   0.010\textperthousand &   0.010\textperthousand &   0.013\textperthousand \\
& S$_{5}$ &   0.093\textperthousand &   0.080\textperthousand &   0.016\textperthousand &  -0.010\textperthousand &  -0.026\textperthousand &  -0.031\textperthousand &  -0.031\textperthousand &  -0.026\textperthousand \\
& Highest DAG $\Delta$ & \textbf{  0.132\textperthousand} & \textbf{  0.117\textperthousand} & \textbf{  0.050\textperthousand} & \textbf{  0.026\textperthousand} & \textbf{  0.042\textperthousand} & \textbf{  0.039\textperthousand} & \textbf{  0.039\textperthousand} & \textbf{  0.040\textperthousand} \\
\hline
\multirow{5}{*}{\tabincell{c}{$p=25\%$\\$m=4$}} & S$_{1}$ &   0.126\textperthousand &   0.120\textperthousand &   0.053\textperthousand &   0.022\textperthousand &   0.013\textperthousand &   0.008\textperthousand &   0.008\textperthousand &   0.009\textperthousand \\
& S$_{2}$ &   0.141\textperthousand &   0.130\textperthousand &   0.062\textperthousand &   0.031\textperthousand &   0.016\textperthousand &   0.013\textperthousand &   0.013\textperthousand &   0.013\textperthousand \\
& S$_{3}$ &   0.104\textperthousand &   0.110\textperthousand &   0.056\textperthousand &   0.025\textperthousand &   0.018\textperthousand &   0.019\textperthousand &   0.019\textperthousand &   0.023\textperthousand \\
& S$_{4}$ &   0.111\textperthousand &   0.104\textperthousand &   0.039\textperthousand &   0.009\textperthousand &   0.017\textperthousand &   0.012\textperthousand &   0.012\textperthousand &   0.013\textperthousand \\
& Highest DAG $\Delta$ & \textbf{  0.141\textperthousand} & \textbf{  0.130\textperthousand} & \textbf{  0.062\textperthousand} & \textbf{  0.031\textperthousand} & \textbf{  0.018\textperthousand} & \textbf{  0.019\textperthousand} & \textbf{  0.019\textperthousand} & \textbf{  0.023\textperthousand} \\
\hline
\multirow{4}{*}{\tabincell{c}{$p=33\%$\\$m=3$}} & S$_{1}$ &   0.063\textperthousand &   0.065\textperthousand &   0.004\textperthousand &  -0.027\textperthousand &  -0.042\textperthousand &   0.009\textperthousand &   0.011\textperthousand &   0.011\textperthousand \\
& S$_{2}$ &   0.116\textperthousand &   0.117\textperthousand &   0.052\textperthousand &   0.021\textperthousand &   0.008\textperthousand &   0.008\textperthousand &   0.010\textperthousand &   0.010\textperthousand \\
& S$_{3}$ &   0.135\textperthousand &   0.113\textperthousand &   0.051\textperthousand &   0.020\textperthousand &   0.026\textperthousand &   0.030\textperthousand &   0.030\textperthousand &   0.030\textperthousand \\
& Highest DAG $\Delta$ & \textbf{  0.135\textperthousand} & \textbf{  0.117\textperthousand} & \textbf{  0.052\textperthousand} & \textbf{  0.021\textperthousand} & \textbf{  0.026\textperthousand} & \textbf{  0.030\textperthousand} & \textbf{  0.030\textperthousand} & \textbf{  0.030\textperthousand} \\
\hline
\multirow{3}{*}{\tabincell{c}{$p=50\%$\\$m=2$}} & S$_{1}$ &   0.083\textperthousand &   0.102\textperthousand &   0.047\textperthousand &   0.021\textperthousand &   0.006\textperthousand &   0.001\textperthousand &   0.003\textperthousand &   0.003\textperthousand \\
& S$_{2}$ &   0.040\textperthousand &   0.095\textperthousand &   0.051\textperthousand &   0.026\textperthousand &   0.011\textperthousand &   0.007\textperthousand &   0.009\textperthousand &   0.014\textperthousand \\
& Highest DAG $\Delta$ & \textbf{  0.083\textperthousand} & \textbf{  0.102\textperthousand} & \textbf{  0.051\textperthousand} & \textbf{  0.026\textperthousand} & \textbf{  0.011\textperthousand} & \textbf{  0.007\textperthousand} & \textbf{  0.009\textperthousand} & \textbf{  0.014\textperthousand} \\
\hline
\multirow{3}{*}{\tabincell{c}{$p=80\%$\\$m=2$}} & S$_{1}$ &  -0.026\textperthousand &   0.047\textperthousand &   0.008\textperthousand &  -0.009\textperthousand &  -0.022\textperthousand &  -0.027\textperthousand &  -0.027\textperthousand &  -0.026\textperthousand \\
& S$_{2}$ &  -0.064\textperthousand &  -0.036\textperthousand &  -0.048\textperthousand &  -0.069\textperthousand &  -0.079\textperthousand &  -0.084\textperthousand &  -0.084\textperthousand &  -0.081\textperthousand \\
& Highest DAG $\Delta$ & \textbf{  -0.026\textperthousand} & \textbf{  0.047\textperthousand} & \textbf{  0.008\textperthousand} & \textbf{ -0.009\textperthousand} & \textbf{ -0.022\textperthousand} & \textbf{ -0.072\textperthousand} & \textbf{ -0.027\textperthousand} & \textbf{ -0.026\textperthousand} \\
\bottomrule
\end{tabular}
\caption{The discrepancy $\Delta$ in the BDeu scores of PS-MINOBS and MINOBS, where a positive discrepancy indicates a better performance for PS-MINOBS. The results are based on case study Reuters-train, and are shown across different hyperparameter settings $p$ and $m$ for PS-MINOBS. The learning runtime is restricted to four hours, and the results are depicted at each 30-minute interval.}
\end{table}

\begin{table}[H]
\scriptsize
\centering
\begin{tabular}{cccccccccc}
\toprule
 \tabincell{c}{PS-MINOBS\\Hyperparameters} & Threads & 0.5h & 1h & 1.5h & 2h & 2.5h & 3h & 3.5h & 4h\\
\hline
\multirow{11}{*}{\tabincell{c}{$p=10\%$\\$m=5$}} & S$_{1}$ &   0.180\textperthousand &   0.087\textperthousand &   0.042\textperthousand &   0.030\textperthousand &   0.008\textperthousand &   0.009\textperthousand &   0.004\textperthousand &   0.006\textperthousand \\
&S$_{2}$ &   0.179\textperthousand &   0.077\textperthousand &   0.049\textperthousand &   0.037\textperthousand &   0.013\textperthousand &   0.010\textperthousand &   0.004\textperthousand &   0.002\textperthousand \\
&S$_{3}$ &   0.181\textperthousand &   0.063\textperthousand &   0.021\textperthousand &   0.010\textperthousand &   0.003\textperthousand &   0.000\textperthousand &  -0.005\textperthousand &  -0.007\textperthousand \\
&S$_{4}$ &   0.179\textperthousand &   0.087\textperthousand &   0.046\textperthousand &   0.032\textperthousand &   0.018\textperthousand &   0.010\textperthousand &   0.006\textperthousand &   0.001\textperthousand \\
&S$_{5}$ &   0.169\textperthousand &   0.092\textperthousand &   0.049\textperthousand &   0.038\textperthousand &   0.018\textperthousand &   0.014\textperthousand &   0.009\textperthousand &   0.005\textperthousand \\
&S$_{6}$ &   0.187\textperthousand &   0.094\textperthousand &   0.050\textperthousand &   0.038\textperthousand &   0.019\textperthousand &   0.014\textperthousand &   0.009\textperthousand &   0.006\textperthousand \\
&S$_{7}$ &   0.155\textperthousand &   0.073\textperthousand &   0.033\textperthousand &   0.019\textperthousand &  -0.007\textperthousand &  -0.014\textperthousand &  -0.019\textperthousand &  -0.010\textperthousand \\
&S$_{8}$ &   0.178\textperthousand &   0.077\textperthousand &   0.034\textperthousand &   0.019\textperthousand &  -0.002\textperthousand &  -0.006\textperthousand &  -0.001\textperthousand &   0.001\textperthousand \\
&S$_{9}$ &   0.209\textperthousand &   0.108\textperthousand &   0.067\textperthousand &   0.055\textperthousand &   0.030\textperthousand &   0.022\textperthousand &   0.016\textperthousand &   0.025\textperthousand \\
&S$_{10}$ &   0.191\textperthousand &   0.088\textperthousand &   0.048\textperthousand &   0.032\textperthousand &   0.010\textperthousand &   0.003\textperthousand &  -0.002\textperthousand &  -0.009\textperthousand\\
& Highest DAG $\Delta$ & \textbf{  0.209\textperthousand} & \textbf{  0.108\textperthousand} & \textbf{  0.067\textperthousand} & \textbf{  0.055\textperthousand} & \textbf{  0.030\textperthousand} & \textbf{  0.022\textperthousand} & \textbf{  0.016\textperthousand} & \textbf{  0.025\textperthousand} \\
\hline
\multirow{6}{*}{\tabincell{c}{$p=20\%$\\$m=5$}} & S$_{1}$ &   0.155\textperthousand &   0.063\textperthousand &   0.025\textperthousand &   0.014\textperthousand &  -0.003\textperthousand &  -0.005\textperthousand &  -0.010\textperthousand &  -0.014\textperthousand \\
& S$_{2}$ &   0.166\textperthousand &   0.077\textperthousand &   0.034\textperthousand &   0.019\textperthousand &   0.003\textperthousand &   0.001\textperthousand &  -0.003\textperthousand &  -0.011\textperthousand \\
& S$_{3}$ &   0.187\textperthousand &   0.102\textperthousand &   0.059\textperthousand &   0.043\textperthousand &   0.017\textperthousand &   0.015\textperthousand &   0.010\textperthousand &   0.004\textperthousand \\
& S$_{4}$ &   0.180\textperthousand &   0.088\textperthousand &   0.050\textperthousand &   0.034\textperthousand &   0.010\textperthousand &   0.004\textperthousand &  -0.001\textperthousand &  -0.008\textperthousand \\
& S$_{5}$ &   0.151\textperthousand &   0.097\textperthousand &   0.053\textperthousand &   0.040\textperthousand &   0.021\textperthousand &   0.020\textperthousand &   0.015\textperthousand &   0.009\textperthousand \\
& Highest DAG $\Delta$ & \textbf{  0.187\textperthousand} & \textbf{  0.102\textperthousand} & \textbf{  0.059\textperthousand} & \textbf{  0.043\textperthousand} & \textbf{  0.021\textperthousand} & \textbf{  0.020\textperthousand} & \textbf{  0.015\textperthousand} & \textbf{  0.009\textperthousand} \\
\hline
\multirow{5}{*}{\tabincell{c}{$p=25\%$\\$m=4$}} & S$_{1}$ &   0.112\textperthousand &   0.097\textperthousand &   0.048\textperthousand &   0.032\textperthousand &   0.011\textperthousand &  -0.003\textperthousand &   0.001\textperthousand &  -0.003\textperthousand \\
& S$_{2}$ &   0.073\textperthousand &   0.090\textperthousand &   0.056\textperthousand &   0.032\textperthousand &   0.005\textperthousand &  -0.004\textperthousand &  -0.010\textperthousand &  -0.013\textperthousand \\
& S$_{3}$ &   0.132\textperthousand &   0.090\textperthousand &   0.061\textperthousand &   0.044\textperthousand &   0.020\textperthousand &   0.003\textperthousand &  -0.000\textperthousand &  -0.004\textperthousand \\
& S$_{4}$ &   0.101\textperthousand &   0.080\textperthousand &   0.039\textperthousand &   0.013\textperthousand &  -0.011\textperthousand &  -0.029\textperthousand &  -0.030\textperthousand &  -0.033\textperthousand \\
& Highest DAG $\Delta$ & \textbf{  0.132\textperthousand} & \textbf{  0.097\textperthousand} & \textbf{  0.061\textperthousand} & \textbf{  0.044\textperthousand} & \textbf{  0.020\textperthousand} & \textbf{  0.003\textperthousand} & \textbf{  0.001\textperthousand} & \textbf{ -0.003\textperthousand} \\
\hline
\multirow{4}{*}{\tabincell{c}{$p=33\%$\\$m=3$}} & S$_{1}$ &   0.087\textperthousand &   0.097\textperthousand &   0.068\textperthousand &   0.051\textperthousand &   0.023\textperthousand &   0.001\textperthousand &   0.005\textperthousand &   0.002\textperthousand \\
& S$_{2}$ &   0.058\textperthousand &   0.104\textperthousand &   0.067\textperthousand &   0.044\textperthousand &   0.016\textperthousand &  -0.005\textperthousand &  -0.005\textperthousand &  -0.008\textperthousand \\
& S$_{3}$ &   0.066\textperthousand &   0.095\textperthousand &   0.044\textperthousand &   0.032\textperthousand &   0.004\textperthousand &  -0.015\textperthousand &  -0.010\textperthousand &  -0.008\textperthousand \\
& Highest DAG $\Delta$ & \textbf{  0.087\textperthousand} & \textbf{  0.104\textperthousand} & \textbf{  0.068\textperthousand} & \textbf{  0.051\textperthousand} & \textbf{  0.023\textperthousand} & \textbf{  0.001\textperthousand} & \textbf{  0.005\textperthousand} & \textbf{  0.002\textperthousand} \\
\hline
\multirow{3}{*}{\tabincell{c}{$p=50\%$\\$m=2$}} & S$_{1}$ &   0.030\textperthousand &   0.067\textperthousand &   0.047\textperthousand &   0.035\textperthousand &   0.017\textperthousand &  -0.007\textperthousand &  -0.006\textperthousand &   0.003\textperthousand \\
& S$_{2}$ &   0.065\textperthousand &   0.059\textperthousand &   0.026\textperthousand &   0.035\textperthousand &   0.031\textperthousand &   0.007\textperthousand &   0.004\textperthousand &   0.008\textperthousand \\
& Highest DAG $\Delta$ & \textbf{  0.065\textperthousand} & \textbf{  0.067\textperthousand} & \textbf{  0.047\textperthousand} & \textbf{  0.035\textperthousand} & \textbf{  0.031\textperthousand} & \textbf{  0.007\textperthousand} & \textbf{  0.004\textperthousand} & \textbf{  0.008\textperthousand} \\
\hline
\multirow{3}{*}{\tabincell{c}{$p=80\%$\\$m=2$}} & S$_{1}$ &  -0.020\textperthousand &   0.036\textperthousand &   0.020\textperthousand &   0.022\textperthousand &   0.012\textperthousand &   0.002\textperthousand &   0.013\textperthousand &   0.010\textperthousand \\
& S$_{2}$ &  -0.036\textperthousand &   0.015\textperthousand &   0.003\textperthousand &  -0.009\textperthousand &  -0.018\textperthousand &  -0.032\textperthousand &  -0.021\textperthousand &  -0.024\textperthousand \\
& Highest DAG $\Delta$ & \textbf{  -0.020\textperthousand} & \textbf{  0.036\textperthousand} & \textbf{  0.020\textperthousand} & \textbf{  0.022\textperthousand} & \textbf{  0.012\textperthousand} & \textbf{  0.002\textperthousand} & \textbf{  0.013\textperthousand} & \textbf{  0.010\textperthousand} \\
\bottomrule
\end{tabular}
\caption{The discrepancy $\Delta$ in the BDeu scores of PS-MINOBS and MINOBS, where a positive discrepancy indicates a better performance for PS-MINOBS. The results are based on case study NewsGroup-test, and are shown across different hyperparameter settings $p$ and $m$ for PS-MINOBS. The learning runtime is restricted to four hours, and the results are depicted at each 30-minute interval.}
\end{table}


\begin{table}[H]
\scriptsize
\centering
\begin{tabular}{cccccccccc}
\toprule
 \tabincell{c}{PS-MINOBS\\Hyperparameters} & Threads & 0.5h & 1h & 1.5h & 2h & 2.5h & 3h & 3.5h & 4h\\
\hline
\multirow{11}{*}{\tabincell{c}{$p=10\%$\\$m=10$}} &S$_{1}$ &   0.369\textperthousand &   0.234\textperthousand &   0.154\textperthousand &   0.122\textperthousand &   0.089\textperthousand &   0.070\textperthousand &   0.064\textperthousand &   0.044\textperthousand \\
&S$_{2}$ &   0.280\textperthousand &   0.229\textperthousand &   0.145\textperthousand &   0.111\textperthousand &   0.081\textperthousand &   0.059\textperthousand &   0.053\textperthousand &   0.034\textperthousand \\
&S$_{3}$ &   0.312\textperthousand &   0.223\textperthousand &   0.136\textperthousand &   0.103\textperthousand &   0.074\textperthousand &   0.064\textperthousand &   0.062\textperthousand &   0.043\textperthousand \\
&S$_{4}$ &   0.273\textperthousand &   0.244\textperthousand &   0.157\textperthousand &   0.123\textperthousand &   0.090\textperthousand &   0.071\textperthousand &   0.064\textperthousand &   0.045\textperthousand \\
&S$_{5}$ &   0.292\textperthousand &   0.223\textperthousand &   0.138\textperthousand &   0.103\textperthousand &   0.068\textperthousand &   0.047\textperthousand &   0.043\textperthousand &   0.024\textperthousand \\
&S$_{6}$ &   0.267\textperthousand &   0.203\textperthousand &   0.145\textperthousand &   0.116\textperthousand &   0.081\textperthousand &   0.060\textperthousand &   0.071\textperthousand &   0.052\textperthousand \\
&S$_{7}$ &   0.269\textperthousand &   0.225\textperthousand &   0.135\textperthousand &   0.100\textperthousand &   0.071\textperthousand &   0.069\textperthousand &   0.065\textperthousand &   0.046\textperthousand \\
&S$_{8}$ &   0.326\textperthousand &   0.249\textperthousand &   0.159\textperthousand &   0.124\textperthousand &   0.098\textperthousand &   0.082\textperthousand &   0.076\textperthousand &   0.057\textperthousand \\
&S$_{9}$ &   0.344\textperthousand &   0.244\textperthousand &   0.148\textperthousand &   0.113\textperthousand &   0.079\textperthousand &   0.062\textperthousand &   0.056\textperthousand &   0.037\textperthousand \\
&S$_{10}$ &   0.267\textperthousand &   0.176\textperthousand &   0.104\textperthousand &   0.069\textperthousand &   0.036\textperthousand &   0.021\textperthousand &   0.022\textperthousand &   0.004\textperthousand \\
& Highest DAG $\Delta$ & \textbf{  0.369\textperthousand} & \textbf{  0.249\textperthousand} & \textbf{  0.159\textperthousand} & \textbf{  0.124\textperthousand} & \textbf{  0.098\textperthousand} & \textbf{  0.082\textperthousand} & \textbf{  0.076\textperthousand} & \textbf{  0.057\textperthousand} \\
\hline
\multirow{6}{*}{\tabincell{c}{$p=20\%$\\$m=5$}} & S$_{1}$ &   0.302\textperthousand &   0.223\textperthousand &   0.144\textperthousand &   0.117\textperthousand &   0.088\textperthousand &   0.069\textperthousand &   0.067\textperthousand &   0.048\textperthousand \\
& S$_{2}$ &   0.250\textperthousand &   0.202\textperthousand &   0.142\textperthousand &   0.114\textperthousand &   0.079\textperthousand &   0.069\textperthousand &   0.066\textperthousand &   0.046\textperthousand \\
& S$_{3}$ &   0.240\textperthousand &   0.182\textperthousand &   0.115\textperthousand &   0.082\textperthousand &   0.049\textperthousand &   0.048\textperthousand &   0.054\textperthousand &   0.036\textperthousand \\
& S$_{4}$ &   0.330\textperthousand &   0.218\textperthousand &   0.164\textperthousand &   0.129\textperthousand &   0.091\textperthousand &   0.069\textperthousand &   0.068\textperthousand &   0.051\textperthousand \\
& S$_{5}$ &   0.234\textperthousand &   0.140\textperthousand &   0.074\textperthousand &   0.049\textperthousand &   0.011\textperthousand &   0.020\textperthousand &   0.024\textperthousand &   0.005\textperthousand \\
& Highest DAG $\Delta$ & \textbf{  0.330\textperthousand} & \textbf{  0.223\textperthousand} & \textbf{  0.164\textperthousand} & \textbf{  0.129\textperthousand} & \textbf{  0.091\textperthousand} & \textbf{  0.069\textperthousand} & \textbf{  0.068\textperthousand} & \textbf{  0.051\textperthousand} \\
\hline
\multirow{5}{*}{\tabincell{c}{$p=25\%$\\$m=4$}} & S$_{1}$ &   0.278\textperthousand &   0.201\textperthousand &   0.131\textperthousand &   0.112\textperthousand &   0.074\textperthousand &   0.057\textperthousand &   0.066\textperthousand &   0.052\textperthousand \\
& S$_{2}$ &   0.276\textperthousand &   0.227\textperthousand &   0.137\textperthousand &   0.112\textperthousand &   0.074\textperthousand &   0.052\textperthousand &   0.047\textperthousand &   0.030\textperthousand \\
& S$_{3}$ &   0.275\textperthousand &   0.199\textperthousand &   0.115\textperthousand &   0.090\textperthousand &   0.057\textperthousand &   0.046\textperthousand &   0.043\textperthousand &   0.026\textperthousand \\
& S$_{4}$ &   0.258\textperthousand &   0.202\textperthousand &   0.121\textperthousand &   0.098\textperthousand &   0.060\textperthousand &   0.043\textperthousand &   0.039\textperthousand &   0.026\textperthousand \\
& Highest DAG $\Delta$ & \textbf{  0.278\textperthousand} & \textbf{  0.227\textperthousand} & \textbf{  0.137\textperthousand} & \textbf{  0.112\textperthousand} & \textbf{  0.074\textperthousand} & \textbf{  0.057\textperthousand} & \textbf{  0.066\textperthousand} & \textbf{  0.052\textperthousand} \\
\hline
\multirow{4}{*}{\tabincell{c}{$p=33\%$\\$m=3$}} & S$_{1}$ &   0.244\textperthousand &   0.190\textperthousand &   0.130\textperthousand &   0.122\textperthousand &   0.089\textperthousand &   0.065\textperthousand &   0.061\textperthousand &   0.043\textperthousand \\
& S$_{2}$ &   0.221\textperthousand &   0.181\textperthousand &   0.106\textperthousand &   0.114\textperthousand &   0.080\textperthousand &   0.058\textperthousand &   0.051\textperthousand &   0.040\textperthousand \\
& S$_{3}$ &   0.249\textperthousand &   0.206\textperthousand &   0.138\textperthousand &   0.120\textperthousand &   0.085\textperthousand &   0.063\textperthousand &   0.061\textperthousand &   0.046\textperthousand \\
& Highest DAG $\Delta$ & \textbf{  0.249\textperthousand} & \textbf{  0.206\textperthousand} & \textbf{  0.138\textperthousand} & \textbf{  0.122\textperthousand} & \textbf{  0.089\textperthousand} & \textbf{  0.065\textperthousand} & \textbf{  0.061\textperthousand} & \textbf{  0.046\textperthousand} \\
\hline
\multirow{3}{*}{\tabincell{c}{$p=50\%$\\$m=2$}} & S$_{1}$ &   0.169\textperthousand &   0.150\textperthousand &   0.103\textperthousand &   0.082\textperthousand &   0.066\textperthousand &   0.058\textperthousand &   0.055\textperthousand &   0.036\textperthousand \\
& S$_{2}$ &   0.122\textperthousand &   0.096\textperthousand &   0.076\textperthousand &   0.079\textperthousand &   0.061\textperthousand &   0.052\textperthousand &   0.051\textperthousand &   0.031\textperthousand \\
& Highest DAG $\Delta$ & \textbf{  0.169\textperthousand} & \textbf{  0.150\textperthousand} & \textbf{  0.103\textperthousand} & \textbf{  0.082\textperthousand} & \textbf{  0.066\textperthousand} & \textbf{  0.058\textperthousand} & \textbf{  0.055\textperthousand} & \textbf{  0.036\textperthousand} \\
\hline
\multirow{3}{*}{\tabincell{c}{$p=80\%$\\$m=2$}} & S$_{1}$ &   0.105\textperthousand &   0.080\textperthousand &   0.040\textperthousand &   0.059\textperthousand &   0.051\textperthousand &   0.054\textperthousand &   0.053\textperthousand &   0.045\textperthousand \\
& S$_{2}$ &   0.053\textperthousand &   0.023\textperthousand &   0.011\textperthousand &   0.022\textperthousand &   0.014\textperthousand &   0.009\textperthousand &   0.009\textperthousand &  -0.002\textperthousand \\
& Highest DAG $\Delta$ & \textbf{  0.105\textperthousand} & \textbf{  0.080\textperthousand} & \textbf{  0.040\textperthousand} & \textbf{  0.059\textperthousand} & \textbf{  0.051\textperthousand} & \textbf{  0.054\textperthousand} & \textbf{  0.053\textperthousand} & \textbf{  0.045\textperthousand} \\
\bottomrule
\end{tabular}
\caption{The discrepancy $\Delta$ in the BDeu scores of PS-MINOBS and MINOBS, where a positive discrepancy indicates a better performance for PS-MINOBS. The results are based on case study NewsGroup-valid, and are shown across different hyperparameter settings $p$ and $m$ for PS-MINOBS. The learning runtime is restricted to four hours, and the results are depicted at each 30-minute interval.}
\end{table}

Overall, the results show that PS-MINOBS finds a higher scoring graph than MINOBS in eight out of the nine case studies, and across most of the hyperparameter settings investigated. However, MINOBS performs considerably better than PS-MINOBS on case study Pumsb-star-test (refer to Table 9), especially when the number of parallel processes is high and sampling percentage is low. This outcome appears to be an outlier relative to the results obtained based on the other eight case studies, and on this basis we decided to investigate this outlier further. Our hypothesis is that the true underlying network of the Pumsb-star-test data set (which is unknown) must be considerably denser than the underlying true networks of the other eight data sets. This would explain why many of the highest scoring CPSs sampled by PS-MINOBS are not part of the optimal DAGs. This hypothesis is supported by the results shown in Figure B.1 in Appendix B, which shows that both the discrepancy $\Delta$ in the BDeu scores and average in-degree (which is a measure of graph density) increase with higher sampling rates.

We summarise the results in Figure 5. Specifically, Figure 5a illustrates how the median percentage gain in BDeu score changes over runtime, and Figure 5b how the gain in BDeu score is influenced by the hyperparameters $p$ and $m$. Figure 5a suggests that the most significant gain in BDeu is found early in the learning process, and this gain deteriorates with the learning time. This is because, given unlimited runtime, MINOBS will eventually outperform PS-MINOBS, and the same applies to exact learning algorithms. Figure 5b suggests that there is no clear difference between scores when $m>2$, although one could say that there is a minor tendency for $\Delta$ to increase with the number of threads executed and the reduction of sampling rate. This is intuitive because PS-MINOBS approximates MINOBS as the sampling rate approaches $100\%$ and $m$ decreases. Figure 5c repeats the analysis of Figure 5b but depicts the number\footnote{The counter considers all of the hyperparameter combinations tested.} of times PS-MINOBS produced the highest scoring graph, rather than the average gain in $\Delta$. In contrast to Figure 5b which suggest that there is no meaningful difference between most of hyperparameter input combinations, Figure 5c shows rather convincingly that a higher number of threads in conjunction with lower sampling rates tends to produce better results relative to MINOBS. This inconsistency between the two figures occurs because the highest scoring DAG produced by PS-MINOBS will not always have the same gain in $\Delta$ relative to MINOBS. While these results show that increasing the number of threads, while at the same time decreasing the sampling rate, improve performance, it is important to clarify that this does not imply that the gain in accuracy will continue to increase by further increasing thread-count and reducing sampling rate.

\begin{figure}[H]
\centering
\includegraphics[height=9cm,width=18cm]{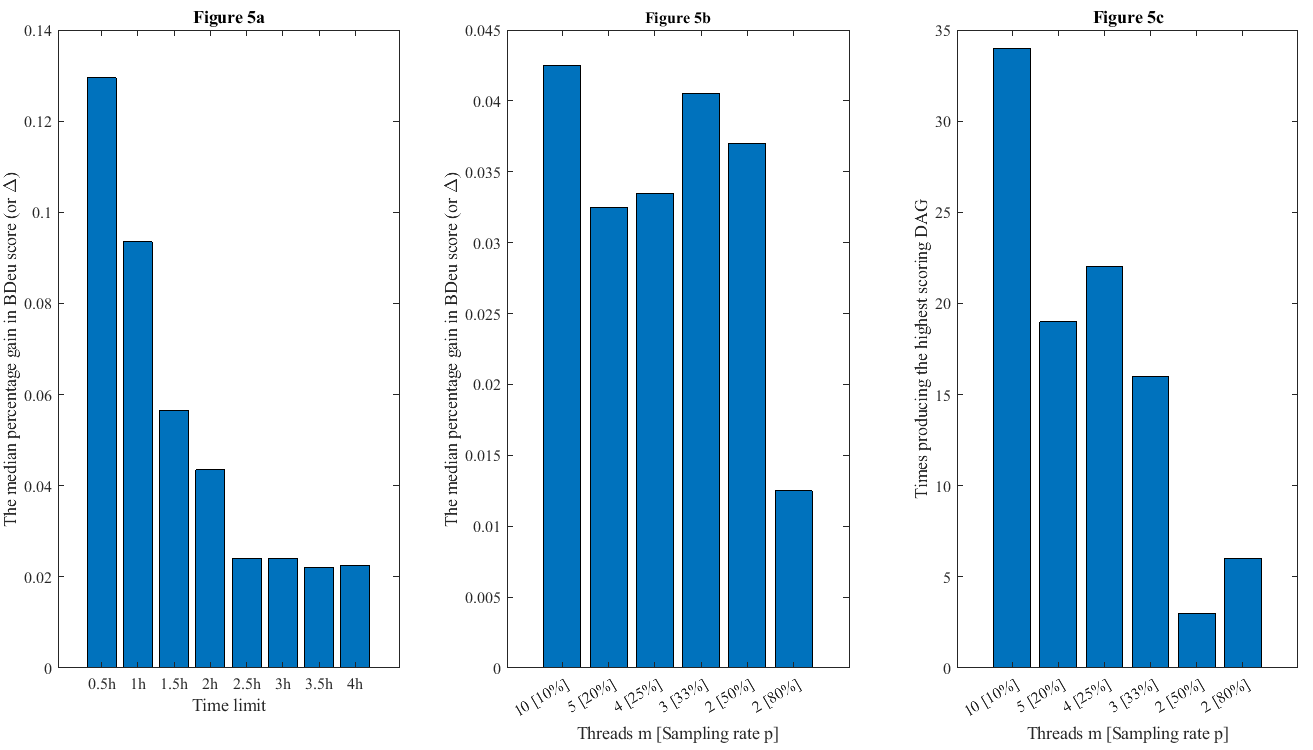}
\caption{The median percentage gain in BDeu score (or $\Delta$) of PS-MINOBS relative to MINOBS, distributed over the different runtime limits (Figure 5a) and the different hyperparameter combinations of threads and sampling rates (Figure 5b), and the number of times PS-MINOBS produced a higher scoring DAG than MINOBS over the same hyperparameters (Figure 5c).}
\end{figure}

\section{Conclusions}

This paper presented PS-MINOBS, which can be viewed as an extension of MINOBS that is suitable for structure learning from high dimensional data, and specifically when we are interested in minimising runtime. The proposed algorithm extends MINOBS in two ways. Firstly, it considers only part of the CPSs, by sampling the CPSs that are more likely to be present in the optimal graph as determined by the Bayesian score. Secondly, it utilises parallel processing and splits sampling into independent suboptimal optimisation problems, and then selects the highest scoring DAG across all parallel optimisations as the output structure.

The limitations of PS-MINOBS are two-fold. First, given infinite runtime, an approach such as PS-MINOBS that samples CPSs is guaranteed to be inferior than an approach that considers all of the CPSs, including MINOBS and exact learning approaches. However, infinite runtime represents an unrealistic scenario in practice. This means that PS-MINOBS can be more suitable in cases where runtime is important factor in determining which structure learning algorithm to consider, and this importance increases with the dimensionality of the input data. Secondly, PS-MINOBS is likely not to be suitable for learning potentially highly dense networks. This is because it samples the most ‘important’ (highest scoring) CPSs, and the highest scoring CPSs might not be sufficient to recover dense structures that typically contain lower scoring CPSs needed to avoid cycles.

Lastly, PS-MINOBS relies on model-selection to output the highest scoring graph across all parallel optimisations. An interesting future research direction would be to investigate how model-averaging approaches could be utilised and applied over CPSs, to further improve accuracy and efficiency. Another direction worth exploring involves introducing some dependency between parallel optimisations, but in such a way that the dependency between threads minimises the negative repercussions on efficiency and maximises accuracy.

\vspace{6pt}

\section*{Acknowledgements}
This research was supported by the ERSRC Fellowship project EP/S001646/1 on Bayesian Artificial Intelligence for Decision Making under Uncertainty \citep{Constantinou-2018}, and by The Alan Turing Institute in the UK under the EPSRC grant EP/N510129/1.

\bibliography{references}
\appendix
\renewcommand{\appendixname}{Appendix~}
\section{Difference in raw BDeu scores for each of the case studies}
\setcounter{table}{0} 


\begin{table}[H]
\scriptsize
\centering
\begin{tabular}{cccccccccc}
\toprule
 \tabincell{c}{PS-MINOBS\\Hyperparameters} & Threads & 0.5h & 1h & 1.5h & 2h & 2.5h & 3h & 3.5h & 4h\\
\hline
 & MINOBS & -620010.1 & -620008.8 & -620005.1 & -620005.1 & -620005.1 & -620005.1 & -620005.1 & -620005.1 \\
\hline
\multirow{11}{*}{\tabincell{c}{$p=10\%$\\$m=10$}} & S$_{1}$ & -620373.9 & -620373.9 & -620372.7 & -620005.1 & -620005.1 & -619992.9 & -619990.9 & -619990.9 \\
 & S$_{2}$ & -619990.9 & -619989.0 & -619989.0 & -619989.0 & -619989.0 & -619989.0 & -619989.0 & -619989.0 \\
 & S$_{3}$ & -620025.1 & -620023.7 & -620013.9 & -620013.9 & -620013.9 & -620013.9 & -620013.9 & -620013.9 \\
 & S$_{4}$ & -619993.1 & -619993.1 & -619993.1 & -619993.1 & -619993.1 & -619993.1 & -619993.1 & -619993.1 \\
 & S$_{5}$ & -620025.7 & -620025.7 & -620025.7 & -620025.7 & -620025.7 & -620025.7 & -620025.7 & -620025.7 \\
 & S$_{6}$ & -620007.6 & -620007.6 & -620003.0 & -619990.9 & -619990.9 & -619990.9 & -619990.9 & -619990.9 \\
 & S$_{7}$ & -619996.6 & -619996.6 & -619992.9 & -619990.9 & -619990.9 & -619990.9 & -619990.9 & -619990.9 \\
 & S$_{8}$ & -619995.4 & -619994.1 & -619994.1 & -619994.1 & -619994.1 & -619994.1 & -619994.1 & -619994.1 \\
 & S$_{9}$ & -620012.2 & -620012.2 & -620012.2 & -620001.4 & -620001.4 & -620001.4 & -620001.4 & -620001.4 \\
 & S$_{10}$ & -620014.2 & -620014.2 & -620014.2 & -620014.2 & -620014.2 & -620014.2 & -620014.2 & -620014.2 \\
 & Highest DAG & \textbf{-619990.9} & \textbf{-619989.0} & \textbf{-619989.0} & \textbf{-619989.0} & \textbf{-619989.0} & \textbf{-619989.0} & \textbf{-619989.0} & \textbf{-619989.0} \\
\hline
\multirow{6}{*}{\tabincell{c}{$p=20\%$\\$m=5$}} & S$_{1}$ & -620003.8 & -620003.8 & -620003.8 & -620003.8 & -620003.8 & -620003.8 & -620003.8 & -620003.8 \\
 & S$_{2}$ & -619998.6 & -619991.6 & -619991.6 & -619991.6 & -619991.6 & -619991.6 & -619991.6 & -619991.6 \\
 & S$_{3}$ & -620003.8 & -620003.8 & -620003.8 & -620003.8 & -620003.8 & -620003.8 & -620003.8 & -620003.8 \\
 & S$_{4}$ & -620005.1 & -620003.8 & -620003.8 & -620003.8 & -619991.6 & -619991.6 & -619991.6 & -619991.6 \\
 & S$_{5}$ & -619991.6 & -619991.6 & -619991.6 & -619991.6 & -619991.6 & -619991.6 & -619991.6 & -619991.6 \\
 & Highest DAG & \textbf{-619991.6} & \textbf{-619991.6} & \textbf{-619991.6} & \textbf{-619991.6} & \textbf{-619991.6} & \textbf{-619991.6} & \textbf{-619991.6} & \textbf{-619991.6} \\
\hline
\multirow{5}{*}{\tabincell{c}{$p=25\%$\\$m=4$}} & S$_{1}$ & -619991.6 & -619991.6 & -619991.6 & -619991.6 & -619991.6 & -619991.6 & -619991.6 & -619991.6 \\
 & S$_{2}$ & -620003.8 & -619991.6 & -619991.6 & -619991.6 & -619991.6 & -619991.6 & -619991.6 & -619991.6 \\
 & S$_{3}$ & -619991.6 & -619991.6 & -619991.6 & -619991.6 & -619991.6 & -619991.6 & -619991.6 & -619991.6 \\
 & S$_{4}$ & -620027.1 & -620025.7 & -620025.7 & -620025.7 & -620025.7 & -620025.7 & -620025.7 & -620025.7 \\
 & Highest DAG & \textbf{-619991.6} & \textbf{-619991.6} & \textbf{-619991.6} & \textbf{-619991.6} & \textbf{-619991.6} & \textbf{-619991.6} & \textbf{-619991.6} & \textbf{-619991.6} \\
\hline
\multirow{4}{*}{\tabincell{c}{$p=33\%$\\$m=3$}} & S$_{1}$ & -620003.8 & -620003.8 & -620003.8 & -620003.8 & -619991.6 & -619991.6 & -619991.6 & -619991.6 \\
 & S$_{2}$ & -619991.6 & -619991.6 & -619991.6 & -619991.6 & -619991.6 & -619991.6 & -619991.6 & -619991.6 \\
 & S$_{3}$ & -620035.7 & -620035.7 & -620035.7 & -620035.7 & -620035.7 & -620035.7 & -620035.7 & -620035.7 \\
 & Highest DAG & \textbf{-619991.6} & \textbf{-619991.6} & \textbf{-619991.6} & \textbf{-619991.6} & \textbf{-619991.6} & \textbf{-619991.6} & \textbf{-619991.6} & \textbf{-619991.6} \\
\hline
\multirow{3}{*}{\tabincell{c}{$p=50\%$\\$m=2$}} & S$_{1}$ & -619992.9 & -619992.9 & -619992.9 & -619991.6 & -619991.6 & -619991.6 & -619991.6 & -619991.6 \\
 & S$_{2}$ & -620012.2 & -620010.9 & -620010.9 & -620010.9 & -620010.9 & -620010.9 & -620010.9 & -620010.9 \\
 & Highest DAG & \textbf{-619992.9} & \textbf{-619992.9} & \textbf{-619992.9} & \textbf{-619991.6} & \textbf{-619991.6} & \textbf{-619991.6} & \textbf{-619991.6} & \textbf{-619991.6} \\
\hline
\multirow{3}{*}{\tabincell{c}{$p=80\%$\\$m=2$}} & S$_{1}$ & -620005.1 & -620003.8 & -620003.8 & -620003.8 & -620003.8 & -620003.8 & -620003.8 & -620003.8 \\
 & S$_{2}$ & -620037.5 & -620033.8 & -620032.5 & -620032.5 & -620032.5 & -620032.5 & -620032.5 & -620032.5 \\
 & Highest DAG & \textbf{-620005.1} & \textbf{-620003.8} & \textbf{-620003.8} & \textbf{-620003.8} & \textbf{-620003.8} & \textbf{-620003.8} & \textbf{-620003.8} & \textbf{-620003.8} \\
\bottomrule
\end{tabular}
\caption{BDeu scores of MINOBS and PS-MINOBS algorithm for case study, Audio-train, under different settings $p=10\%$, $m=10$, $p=20\%$, $m=5$, $p=25\%$, $m=4$, $p=33\%$, $m=3$, $p=50\%$, $m=2$, and $p=80\%$, $m=2$, within different time limits.}
\end{table}
\begin{table}[H]
\scriptsize
\centering
\begin{tabular}{cccccccccc}
\toprule
 \tabincell{c}{PS-MINOBS\\Hyperparameters} & Threads & 0.5h & 1h & 1.5h & 2h & 2.5h & 3h & 3.5h & 4h\\
\hline
 & MINOBS & -493911.5 & -493899.5 & -493898.5 & -493896.9 & -493895.8 & -493895.8 & -493895.8 & -493895.8 \\
\hline
\multirow{11}{*}{\tabincell{c}{$p=10\%$\\$m=10$}} & S$_{1}$ & -494186.4 & -493953.1 & -493937.7 & -493901.4 & -493896.9 & -493892.0 & -493892.0 & -493892.0 \\
 & S$_{2}$ & -493879.2 & -493879.2 & -493879.2 & -493879.2 & -493879.2 & -493879.2 & -493879.2 & -493879.2 \\
 & S$_{3}$ & -493866.3 & -493866.3 & -493866.3 & -493866.3 & -493866.3 & -493866.3 & -493866.3 & -493866.3 \\
 & S$_{4}$ & -493882.4 & -493882.4 & -493882.4 & -493882.4 & -493882.4 & -493882.4 & -493882.4 & -493882.4 \\
 & S$_{5}$ & -493858.1 & -493858.1 & -493850.9 & -493850.9 & -493850.9 & -493850.9 & -493850.9 & -493850.9 \\
 & S$_{6}$ & -494185.1 & -494185.1 & -494182.9 & -494177.8 & -494177.8 & -494177.8 & -494177.8 & -494177.8 \\
 & S$_{7}$ & -493866.3 & -493866.3 & -493866.3 & -493866.3 & -493866.3 & -493866.3 & -493866.3 & -493866.3 \\
 & S$_{8}$ & -493899.2 & -493899.2 & -493899.2 & -493899.2 & -493899.2 & -493899.2 & -493899.2 & -493899.2 \\
 & S$_{9}$ & -493896.9 & -493886.4 & -493869.4 & -493853.9 & -493853.9 & -493853.9 & -493853.9 & -493853.9 \\
 & S$_{10}$ & -493884.8 & -493879.3 & -493866.3 & -493866.3 & -493866.3 & -493866.3 & -493866.3 & -493866.3 \\
 & Highest DAG & \textbf{-493858.1} & \textbf{-493858.1} & \textbf{-493850.9} & \textbf{-493850.9} & \textbf{-493850.9} & \textbf{-493850.9} & \textbf{-493850.9} & \textbf{-493850.9} \\
\hline
\multirow{6}{*}{\tabincell{c}{$p=20\%$\\$m=5$}} & S$_{1}$ & -493887.7 & -493887.7 & -493887.7 & -493887.7 & -493887.7 & -493887.7 & -493877.5 & -493877.5 \\
 & S$_{2}$ & -493946.4 & -493946.4 & -493946.4 & -493946.4 & -493946.4 & -493946.4 & -493946.4 & -493946.4 \\
 & S$_{3}$ & -493929.8 & -493929.8 & -493929.8 & -493929.8 & -493929.8 & -493929.8 & -493929.8 & -493929.8 \\
 & S$_{4}$ & -493859.3 & -493859.3 & -493848.9 & -493848.9 & -493848.9 & -493848.9 & -493848.9 & -493848.9 \\
 & S$_{5}$ & -493853.2 & -493851.6 & -493851.6 & -493851.6 & -493851.6 & -493851.6 & -493851.6 & -493851.6 \\
 & Highest DAG & \textbf{-493853.2} & \textbf{-493851.6} & \textbf{-493848.9} & \textbf{-493848.9} & \textbf{-493848.9} & \textbf{-493848.9} & \textbf{-493848.9} & \textbf{-493848.9} \\
\hline
\multirow{5}{*}{\tabincell{c}{$p=25\%$\\$m=4$}} & S$_{1}$ & -493880.5 & -493880.5 & -493880.5 & -493880.5 & -493880.5 & -493880.5 & -493880.5 & -493880.5 \\
 & S$_{2}$ & -493848.9 & -493848.9 & -493848.9 & -493848.9 & -493848.9 & -493848.9 & -493848.9 & -493848.9 \\
 & S$_{3}$ & -493870.4 & -493850.4 & -493848.9 & -493848.9 & -493848.9 & -493848.9 & -493848.9 & -493848.9 \\
 & S$_{4}$ & -493899.0 & -493868.9 & -493868.9 & -493868.9 & -493868.9 & -493868.9 & -493868.9 & -493868.9 \\
 & Highest DAG & \textbf{-493848.9} & \textbf{-493848.9} & \textbf{-493848.9} & \textbf{-493848.9} & \textbf{-493848.9} & \textbf{-493848.9} & \textbf{-493848.9} & \textbf{-493848.9} \\
\hline
\multirow{4}{*}{\tabincell{c}{$p=33\%$\\$m=3$}} & S$_{1}$ & -493848.9 & -493848.9 & -493848.9 & -493848.9 & -493848.9 & -493848.9 & -493848.9 & -493848.9 \\
 & S$_{2}$ & -493890.4 & -493863.9 & -493863.9 & -493863.9 & -493863.9 & -493863.9 & -493863.9 & -493863.9 \\
 & S$_{3}$ & -493900.1 & -493882.6 & -493881.9 & -493881.9 & -493881.9 & -493881.9 & -493881.9 & -493881.9 \\
 & Highest DAG & \textbf{-493848.9} & \textbf{-493848.9} & \textbf{-493848.9} & \textbf{-493848.9} & \textbf{-493848.9} & \textbf{-493848.9} & \textbf{-493848.9} & \textbf{-493848.9} \\
\hline
\multirow{3}{*}{\tabincell{c}{$p=50\%$\\$m=2$}} & S$_{1}$ & -493878.6 & -493877.5 & -493877.5 & -493877.5 & -493877.5 & -493877.5 & -493877.5 & -493877.5 \\
 & S$_{2}$ & -493899.5 & -493889.6 & -493889.6 & -493889.6 & -493889.6 & -493889.6 & -493889.6 & -493889.6 \\
 & Highest DAG & \textbf{-493878.6} & \textbf{-493877.5} & \textbf{-493877.5} & \textbf{-493877.5} & \textbf{-493877.5} & \textbf{-493877.5} & \textbf{-493877.5} & \textbf{-493877.5} \\
\hline
\multirow{3}{*}{\tabincell{c}{$p=80\%$\\$m=2$}} & S$_{1}$ & -493905.8 & -493901.6 & -493880.5 & -493880.5 & -493877.5 & -493877.5 & -493877.5 & -493877.5 \\
 & S$_{2}$ & -493885.9 & -493877.5 & -493877.5 & -493877.5 & -493877.5 & -493877.5 & -493877.5 & -493877.5 \\
 & Highest DAG & \textbf{-493885.9} & \textbf{-493877.5} & \textbf{-493877.5} & \textbf{-493877.5} & \textbf{-493877.5} & \textbf{-493877.5} & \textbf{-493877.5} & \textbf{-493877.5} \\

\bottomrule
\end{tabular}
\caption{BDeu scores of MINOBS and PS-MINOBS algorithm for case study, Jester-train, under different settings $p=10\%$, $m=10$, $p=20\%$, $m=5$, $p=25\%$, $m=4$, $p=33\%$, $m=3$, $p=50\%$, $m=2$, and $p=80\%$, $m=2$, within different time limits.}
\end{table}

\begin{table}[H]
\scriptsize
\centering
\begin{tabular}{cccccccccc}
\toprule
 \tabincell{c}{PS-MINOBS\\Hyperparameters} & Threads & 0.5h & 1h & 1.5h & 2h & 2.5h & 3h & 3.5h & 4h\\
\hline
 & MINOBS & -5712.6 & -5711.2 & -5710.8 & -5710.6 & -5710.2 & -5707.3 & -5707.3 & -5706.9 \\
\hline
\multirow{11}{*}{\tabincell{c}{$p=10\%$\\$m=10$}} & S$_{1}$ & -5759.1 & -5758.4 & -5758.3 & -5758.3 & -5754.6 & -5753.8 & -5753.6 & -5753.6 \\
 & S$_{2}$ & -5745.7 & -5743.6 & -5743.4 & -5743.4 & -5743.4 & -5743.4 & -5743.4 & -5743.4 \\
 & S$_{3}$ & -5746.7 & -5744.2 & -5743.8 & -5743.8 & -5743.8 & -5743.8 & -5743.8 & -5743.7 \\
 & S$_{4}$ & -5749.7 & -5746.9 & -5745.5 & -5745.5 & -5745.2 & -5745.2 & -5745.1 & -5744.9 \\
 & S$_{5}$ & -5747.5 & -5743.0 & -5742.4 & -5742.3 & -5742.1 & -5742.1 & -5742.1 & -5741.9 \\
 & S$_{6}$ & -5752.0 & -5751.2 & -5750.6 & -5750.2 & -5750.2 & -5750.2 & -5750.2 & -5750.2 \\
 & S$_{7}$ & -5746.9 & -5744.9 & -5744.3 & -5744.3 & -5744.1 & -5744.1 & -5744.1 & -5743.8 \\
 & S$_{8}$ & -5746.9 & -5743.7 & -5743.3 & -5743.1 & -5742.9 & -5742.2 & -5741.6 & -5740.8 \\
 & S$_{9}$ & -5745.4 & -5743.9 & -5743.8 & -5743.8 & -5743.8 & -5742.9 & -5742.9 & -5742.9 \\
 & S$_{10}$ & -5748.6 & -5744.7 & -5742.8 & -5742.8 & -5742.7 & -5742.4 & -5742.4 & -5742.4 \\
 & Highest DAG & \textbf{-5745.4} & \textbf{-5743.0} & \textbf{-5742.4} & \textbf{-5742.3} & \textbf{-5742.1} & \textbf{-5742.1} & \textbf{-5741.6} & \textbf{-5740.8} \\
\hline
\multirow{6}{*}{\tabincell{c}{$p=20\%$\\$m=5$}} & S$_{1}$ & -5735.9 & -5734.7 & -5734.1 & -5733.9 & -5733.9 & -5733.9 & -5733.9 & -5733.9 \\
 & S$_{2}$ & -5723.3 & -5721.9 & -5721.5 & -5721.5 & -5721.5 & -5721.5 & -5721.5 & -5721.5 \\
 & S$_{3}$ & -5725.9 & -5723.4 & -5722.1 & -5721.9 & -5721.8 & -5721.8 & -5721.8 & -5721.8 \\
 & S$_{4}$ & -5722.2 & -5721.2 & -5721.0 & -5721.0 & -5720.8 & -5720.8 & -5720.8 & -5720.8 \\
 & S$_{5}$ & -5727.8 & -5725.5 & -5725.2 & -5724.3 & -5724.3 & -5724.0 & -5724.0 & -5724.0 \\
 & Highest DAG & \textbf{-5722.2} & \textbf{-5721.2} & \textbf{-5721.0} & \textbf{-5721.0} & \textbf{-5720.8} & \textbf{-5720.8} & \textbf{-5720.8} & \textbf{-5720.8} \\
\hline
\multirow{5}{*}{\tabincell{c}{$p=25\%$\\$m=4$}} & S$_{1}$ & -5728.0 & -5721.6 & -5721.1 & -5720.6 & -5720.6 & -5720.6 & -5720.6 & -5720.6 \\
 & S$_{2}$ & -5718.4 & -5717.6 & -5717.2 & -5717.2 & -5717.2 & -5717.2 & -5717.2 & -5717.2 \\
 & S$_{3}$ & -5720.2 & -5718.5 & -5717.7 & -5717.5 & -5717.3 & -5717.2 & -5717.2 & -5717.2 \\
 & S$_{4}$ & -5720.3 & -5718.9 & -5718.6 & -5718.5 & -5718.5 & -5718.2 & -5718.2 & -5718.2 \\
 & Highest DAG & \textbf{-5718.4} & \textbf{-5717.6} & \textbf{-5717.2} & \textbf{-5717.2} & \textbf{-5717.2} & \textbf{-5717.2} & \textbf{-5717.2} & \textbf{-5717.2} \\
\hline
\multirow{4}{*}{\tabincell{c}{$p=33\%$\\$m=3$}} & S$_{1}$ & -5722.2 & -5716.8 & -5714.6 & -5714.2 & -5714.1 & -5714.1 & -5714.1 & -5714.1 \\
 & S$_{2}$ & -5717.6 & -5715.2 & -5715.2 & -5715.2 & -5714.9 & -5714.9 & -5714.8 & -5714.8 \\
 & S$_{3}$ & -5715.4 & -5713.7 & -5713.6 & -5713.6 & -5713.4 & -5713.3 & -5713.3 & -5713.3 \\
 & Highest DAG & \textbf{-5715.4} & \textbf{-5713.7} & \textbf{-5713.6} & \textbf{-5713.6} & \textbf{-5713.4} & \textbf{-5713.3} & \textbf{-5713.3} & \textbf{-5713.3} \\
\hline
\multirow{3}{*}{\tabincell{c}{$p=50\%$\\$m=2$}} & S$_{1}$ & -5715.2 & -5712.3 & -5711.8 & -5710.8 & -5710.8 & -5710.8 & -5710.8 & -5710.8 \\
 & S$_{2}$ & -5712.6 & -5707.9 & -5707.2 & -5707.2 & -5707.2 & -5707.1 & -5707.1 & -5707.1 \\
 & Highest DAG & \textbf{-5712.6} & \textbf{-5707.9} & \textbf{-5707.2} & \textbf{-5707.2} & \textbf{-5707.2} & \textbf{-5707.1} & \textbf{-5707.1} & \textbf{-5707.1} \\
\hline
\multirow{3}{*}{\tabincell{c}{$p=80\%$\\$m=2$}} & S$_{1}$ & -5711.4 & -5707.8 & -5707.4 & -5707.3 & -5707.3 & -5707.3 & -5707.3 & -5707.3 \\
 & S$_{2}$ & -5711.3 & -5707.9 & -5707.6 & -5707.2 & -5707.1 & -5707.1 & -5707.0 & -5707.0 \\
 & Highest DAG & \textbf{-5711.3} & \textbf{-5707.8} & \textbf{-5707.4} & \textbf{-5707.2} & \textbf{-5707.1} & \textbf{-5707.1} & \textbf{-5707.0} & \textbf{-5707.0} \\

\bottomrule
\end{tabular}
\caption{BDeu scores of MINOBS and PS-MINOBS algorithm for case study, Pumsb-star-test, under different settings $p=10\%$, $m=10$, $p=20\%$, $m=5$, $p=25\%$, $m=4$, $p=33\%$, $m=3$, $p=50\%$, $m=2$, and $p=80\%$, $m=2$, within different time limits.}
\end{table}


\begin{table}[H]
\scriptsize
\centering
\begin{tabular}{cccccccccc}
\toprule
 \tabincell{c}{PS-MINOBS\\Hyperparameters} & Threads & 0.5h & 1h & 1.5h & 2h & 2.5h & 3h & 3.5h & 4h\\
\hline
 & MINOBS & -369471.9 & -369449.8 & -369440.6 & -369440.6 & -369440.6 & -369440.6 & -369440.6 & -369440.6 \\
\hline
\multirow{11}{*}{\tabincell{c}{$p=10\%$\\$m=10$}} & S$_{1}$ & -369443.5 & -369443.5 & -369443.5 & -369442.0 & -369442.0 & -369442.0 & -369442.0 & -369442.0 \\
 & S$_{2}$ & -369454.4 & -369454.4 & -369454.4 & -369454.4 & -369454.4 & -369454.4 & -369454.4 & -369454.4 \\
 & S$_{3}$ & -369456.8 & -369456.8 & -369456.8 & -369456.8 & -369456.8 & -369456.8 & -369456.8 & -369456.8 \\
 & S$_{4}$ & -369438.5 & -369438.5 & -369438.5 & -369438.5 & -369438.5 & -369438.5 & -369438.5 & -369438.5 \\
 & S$_{5}$ & -369442.0 & -369442.0 & -369442.0 & -369442.0 & -369442.0 & -369442.0 & -369442.0 & -369442.0 \\
 & S$_{6}$ & -369457.6 & -369454.4 & -369442.0 & -369442.0 & -369442.0 & -369442.0 & -369442.0 & -369442.0 \\
 & S$_{7}$ & -369458.9 & -369455.8 & -369447.2 & -369447.2 & -369447.2 & -369447.2 & -369447.2 & -369447.2 \\
 & S$_{8}$ & -369450.9 & -369447.9 & -369447.9 & -369447.9 & -369447.9 & -369445.9 & -369444.7 & -369444.7 \\
 & S$_{9}$ & -369448.2 & -369448.2 & -369445.2 & -369445.2 & -369445.2 & -369445.2 & -369445.2 & -369442.0 \\
 & S$_{10}$ & -369448.2 & -369448.2 & -369448.2 & -369448.2 & -369445.2 & -369445.2 & -369445.2 & -369445.2 \\
 & Highest DAG & \textbf{-369438.5} & \textbf{-369438.5} & \textbf{-369438.5} & \textbf{-369438.5} & \textbf{-369438.5} & \textbf{-369438.5} & \textbf{-369438.5} & \textbf{-369438.5} \\
\hline
\multirow{6}{*}{\tabincell{c}{$p=20\%$\\$m=5$}} & S$_{1}$ & -369441.7 & -369438.5 & -369438.5 & -369438.5 & -369438.5 & -369438.5 & -369438.5 & -369438.5 \\
 & S$_{2}$ & -369441.6 & -369441.6 & -369441.6 & -369441.6 & -369438.5 & -369438.5 & -369438.5 & -369438.5 \\
 & S$_{3}$ & -369443.2 & -369440.6 & -369439.1 & -369439.1 & -369439.1 & -369439.1 & -369439.1 & -369439.1 \\
 & S$_{4}$ & -369460.8 & -369444.9 & -369441.9 & -369441.9 & -369441.9 & -369441.9 & -369441.9 & -369441.9 \\
 & S$_{5}$ & -369441.6 & -369441.6 & -369441.6 & -369441.6 & -369441.6 & -369438.5 & -369438.5 & -369438.5 \\
 & Highest DAG & \textbf{-369441.6} & \textbf{-369438.5} & \textbf{-369438.5} & \textbf{-369438.5} & \textbf{-369438.5} & \textbf{-369438.5} & \textbf{-369438.5} & \textbf{-369438.5} \\
\hline
\multirow{5}{*}{\tabincell{c}{$p=25\%$\\$m=4$}} & S$_{1}$ & -369445.1 & -369441.6 & -369441.6 & -369441.6 & -369441.6 & -369441.6 & -369441.6 & -369441.6 \\
 & S$_{2}$ & -369444.8 & -369444.8 & -369441.7 & -369441.7 & -369441.7 & -369441.7 & -369441.7 & -369441.7 \\
 & S$_{3}$ & -369447.2 & -369444.0 & -369440.9 & -369440.9 & -369440.9 & -369440.9 & -369440.9 & -369440.9 \\
 & S$_{4}$ & -369455.2 & -369455.2 & -369455.2 & -369455.2 & -369455.2 & -369455.2 & -369455.2 & -369455.2 \\
 & Highest DAG & \textbf{-369444.8} & \textbf{-369441.6} & \textbf{-369440.9} & \textbf{-369440.9} & \textbf{-369440.9} & \textbf{-369440.9} & \textbf{-369440.9} & \textbf{-369440.9} \\
\hline
\multirow{4}{*}{\tabincell{c}{$p=33\%$\\$m=3$}} & S$_{1}$ & -369440.8 & -369438.5 & -369438.5 & -369438.5 & -369438.5 & -369438.5 & -369438.5 & -369438.5 \\
 & S$_{2}$ & -369457.4 & -369457.4 & -369441.6 & -369441.6 & -369441.6 & -369441.6 & -369441.6 & -369441.6 \\
 & S$_{3}$ & -369440.0 & -369438.5 & -369438.5 & -369438.5 & -369438.5 & -369438.5 & -369438.5 & -369438.5 \\
 & Highest DAG & \textbf{-369440.0} & \textbf{-369438.5} & \textbf{-369438.5} & \textbf{-369438.5} & \textbf{-369438.5} & \textbf{-369438.5} & \textbf{-369438.5} & \textbf{-369438.5} \\
\hline
\multirow{3}{*}{\tabincell{c}{$p=50\%$\\$m=2$}} & S$_{1}$ & -369461.4 & -369444.3 & -369441.7 & -369441.7 & -369441.7 & -369441.7 & -369441.7 & -369441.7 \\
 & S$_{2}$ & -369460.7 & -369445.9 & -369445.9 & -369444.4 & -369444.4 & -369444.4 & -369444.4 & -369444.4 \\
 & Highest DAG & \textbf{-369460.7} & \textbf{-369444.3} & \textbf{-369441.7} & \textbf{-369441.7} & \textbf{-369441.7} & \textbf{-369441.7} & \textbf{-369441.7} & \textbf{-369441.7} \\
\hline
\multirow{3}{*}{\tabincell{c}{$p=80\%$\\$m=2$}} & S$_{1}$ & -369468.6 & -369449.3 & -369441.7 & -369441.7 & -369441.7 & -369441.7 & -369441.7 & -369441.7 \\
 & S$_{2}$ & -369469.9 & -369457.4 & -369441.6 & -369441.6 & -369441.6 & -369441.6 & -369441.6 & -369441.6 \\
 & Highest DAG & \textbf{-369468.6} & \textbf{-369449.3} & \textbf{-369441.6} & \textbf{-369441.6} & \textbf{-369441.6} & \textbf{-369441.6} & \textbf{-369441.6} & \textbf{-369441.6} \\
\bottomrule
\end{tabular}
\caption{BDeu scores of MINOBS and PS-MINOBS algorithm for case study, Kosarek-train, under different settings $p=10\%$, $m=10$, $p=20\%$, $m=5$, $p=25\%$, $m=4$, $p=33\%$, $m=3$, $p=50\%$, $m=2$, and $p=80\%$, $m=2$, within different time limits.}
\end{table}


\begin{table}[H]
\scriptsize
\centering
\begin{tabular}{cccccccccc}
\toprule
 \tabincell{c}{PS-MINOBS\\Hyperparameters} & Threads & 0.5h & 1h & 1.5h & 2h & 2.5h & 3h & 3.5h & 4h\\
\hline
 & MINOBS & -267889.8 & -267865.1 & -267847. & -267838.9 & -267835.7 & -267832.7 & -267832.7 & -267832.7 \\
\hline
\multirow{11}{*}{\tabincell{c}{$p=10\%$\\$m=10$}} & S$_{1}$ & -267822.5 & -267817.9 & -267814.5 & -267814.5 & -267814.5 & -267814.5 & -267814.4 & -267814.3 \\
 & S$_{2}$ & -267822.7 & -267820.0 & -267819.2 & -267819.2 & -267819.2 & -267819.2 & -267819.2 & -267816.0 \\
 & S$_{3}$ & -267832.3 & -267830.2 & -267830.0 & -267829.8 & -267828.1 & -267821.9 & -267821.8 & -267821.7 \\
 & S$_{4}$ & -267826.8 & -267824.5 & -267822.7 & -267822.6 & -267822.6 & -267822.6 & -267822.1 & -267821.9 \\
 & S$_{5}$ & -267866.1 & -267865.6 & -267865.5 & -267865.5 & -267865.3 & -267865.3 & -267865.3 & -267865.3 \\
 & S$_{6}$ & -267817.4 & -267816.6 & -267816.3 & -267816.3 & -267816.3 & -267814.5 & -267814.3 & -267814.3 \\
 & S$_{7}$ & -267856.6 & -267838.9 & -267833.4 & -267833.1 & -267830.9 & -267830.9 & -267830.9 & -267830.9 \\
 & S$_{8}$ & -267825.4 & -267818.6 & -267816.2 & -267816.2 & -267816.2 & -267816.2 & -267816.0 & -267816.0 \\
 & S$_{9}$ & -267819.6 & -267818.0 & -267818.0 & -267816.4 & -267816.1 & -267816.1 & -267816.1 & -267816.1 \\
 & S$_{10}$ & -267817.7 & -267816.9 & -267816.5 & -267816.5 & -267815.9 & -267815.9 & -267815.9 & -267815.9 \\
 & Highest DAG & \textbf{-267817.4} & \textbf{-267816.6} & \textbf{-267814.5} & \textbf{-267814.5} & \textbf{-267814.5} & \textbf{-267814.5} & \textbf{-267814.3} & \textbf{-267814.3} \\
\hline
\multirow{6}{*}{\tabincell{c}{$p=20\%$\\$m=5$}} & S$_{1}$ & -267836.5 & -267820.0 & -267817.2 & -267816.2 & -267815.8 & -267815.8 & -267814.3 & -267814.3 \\
 & S$_{2}$ & -267866.9 & -267865.0 & -267847.8 & -267847.8 & -267843.0 & -267843.0 & -267841.4 & -267839.8 \\
 & S$_{3}$ & -267833.9 & -267831.9 & -267828.1 & -267828.1 & -267828.1 & -267828.1 & -267827.2 & -267827.1 \\
 & S$_{4}$ & -267830.0 & -267821.9 & -267821.9 & -267820.2 & -267819.9 & -267819.9 & -267816.7 & -267816.5 \\
 & S$_{5}$ & -267835.9 & -267822.9 & -267822.6 & -267822.4 & -267822.4 & -267818.1 & -267816.8 & -267815.9 \\
 & Highest DAG & \textbf{-267830.0} & \textbf{-267820.0} & \textbf{-267817.2} & \textbf{-267816.2} & \textbf{-267815.8} & \textbf{-267815.8} & \textbf{-267814.3} & \textbf{-267814.3} \\
\hline
\multirow{5}{*}{\tabincell{c}{$p=25\%$\\$m=4$}} & S$_{1}$ & -267836.7 & -267828.1 & -267819.1 & -267815.1 & -267815.1 & -267814.6 & -267814.6 & -267814.6 \\
 & S$_{2}$ & -267832.3 & -267818.5 & -267817.8 & -267815.8 & -267815.6 & -267815.5 & -267815.5 & -267815.5 \\
 & S$_{3}$ & -267841.7 & -267836.7 & -267821.8 & -267818.1 & -267818.1 & -267817.5 & -267817.4 & -267817.4 \\
 & S$_{4}$ & -267831.7 & -267823.8 & -267816.7 & -267812.8 & -267812.6 & -267812.6 & -267812.6 & -267812.6 \\
 & Highest DAG & \textbf{-267831.7} & \textbf{-267818.5} & \textbf{-267816.7} & \textbf{-267812.8} & \textbf{-267812.6} & \textbf{-267812.6} & \textbf{-267812.6} & \textbf{-267812.6} \\
\hline
\multirow{4}{*}{\tabincell{c}{$p=33\%$\\$m=3$}} & S$_{1}$ & -267835.5 & -267821.3 & -267821.3 & -267820.9 & -267820.9 & -267820.7 & -267816.9 & -267815.6 \\
 & S$_{2}$ & -267902.2 & -267845.5 & -267845.3 & -267845.3 & -267844.9 & -267844.9 & -267844.9 & -267844.9 \\
 & S$_{3}$ & -267824.3 & -267822.7 & -267822.7 & -267822.6 & -267818.4 & -267818.4 & -267818.2 & -267818.2 \\
 & Highest DAG & \textbf{-267824.3} & \textbf{-267821.3} & \textbf{-267821.3} & \textbf{-267820.9} & \textbf{-267818.4} & \textbf{-267818.4} & \textbf{-267816.9} & \textbf{-267815.6} \\
\hline
\multirow{3}{*}{\tabincell{c}{$p=50\%$\\$m=2$}} & S$_{1}$ & -267869.3 & -267826.5 & -267824.5 & -267824.5 & -267824.5 & -267824.3 & -267824.3 & -267824.3 \\
 & S$_{2}$ & -267893.6 & -267820.7 & -267816.9 & -267816.9 & -267816.9 & -267815.2 & -267814.5 & -267814.5 \\
 & Highest DAG & \textbf{-267869.3} & \textbf{-267820.7} & \textbf{-267816.9} & \textbf{-267816.9} & \textbf{-267816.9} & \textbf{-267815.2} & \textbf{-267814.5} & \textbf{-267814.5} \\
\hline
\multirow{3}{*}{\tabincell{c}{$p=80\%$\\$m=2$}} & S$_{1}$ & -267962.1 & -267848.4 & -267826.6 & -267824.9 & -267821.7 & -267821.7 & -267821.7 & -267821.6 \\
 & S$_{2}$ & -267905.2 & -267832.3 & -267824.5 & -267822.7 & -267821.1 & -267821.1 & -267821.0 & -267820.8 \\
 & Highest DAG & \textbf{-267905.2} & \textbf{-267832.3} & \textbf{-267824.5} & \textbf{-267822.7} & \textbf{-267821.1} & \textbf{-267821.1} & \textbf{-267821.0} & \textbf{-267820.8} \\
\bottomrule
\end{tabular}
\caption{BDeu scores of MINOBS and PS-MINOBS algorithm for case study, EachMovie-Train, under different settings $p=10\%$, $m=10$, $p=20\%$, $m=5$, $p=25\%$, $m=4$, $p=33\%$, $m=3$, $p=50\%$, $m=2$, and $p=80\%$, $m=2$, within different time limits.}
\end{table}


\begin{table}[H]
\scriptsize
\centering
\begin{tabular}{cccccccccc}
\toprule
 \tabincell{c}{PS-MINOBS\\Hyperparameters} & Threads & 0.5h & 1h & 1.5h & 2h & 2.5h & 3h & 3.5h & 4h\\
\hline
 & MINOBS & -131735.3 & -131705.9 & -131701.9 & -131698.8 & -131694.1 & -131693.5 & -131690.8 & -131690.3 \\
\hline
\multirow{11}{*}{\tabincell{c}{$p=10\%$\\$m=10$}} & S$_{1}$ & -131718.5 & -131714.5 & -131713.5 & -131712.1 & -131712.1 & -131711.8 & -131711.8 & -131711.8 \\
 & S$_{2}$ & -131714.3 & -131711.3 & -131710.6 & -131709.8 & -131708.1 & -131707.5 & -131707.5 & -131707.5 \\
 & S$_{3}$ & -131702.9 & -131698.8 & -131697.6 & -131694.9 & -131694.1 & -131693.9 & -131693.8 & -131693.3 \\
 & S$_{4}$ & -131723.7 & -131722.2 & -131717.9 & -131714.2 & -131713.5 & -131711.0 & -131710.8 & -131710.7 \\
 & S$_{5}$ & -131711.9 & -131700.6 & -131692.8 & -131689.8 & -131688.3 & -131688.2 & -131688.2 & -131688.2 \\
 & S$_{6}$ & -131732.8 & -131727.2 & -131724.8 & -131722.3 & -131721.9 & -131721.7 & -131721.7 & -131717.1 \\
 & S$_{7}$ & -131716.0 & -131708.0 & -131705.4 & -131701.7 & -131699.3 & -131699.2 & -131699.2 & -131698.7 \\
 & S$_{8}$ & -131712.4 & -131710.5 & -131708.9 & -131708.0 & -131708.0 & -131707.9 & -131707.9 & -131707.9 \\
 & S$_{9}$ & -131720.2 & -131717.3 & -131710.3 & -131709.89 & -131707.4 & -131707.4 & -131707.4 & -131707.2 \\
 & S$_{10}$ & -131724.4 & -131721.9 & -131718.2 & -131716.8 & -131716.8 & -131714.2 & -131714.2 & -131714.1 \\
 & Highest DAG & \textbf{-131702.9} & \textbf{-131698.8} & \textbf{-131692.8} & \textbf{-131689.9} & \textbf{-131688.3} & \textbf{-131688.2} & \textbf{-131688.2} & \textbf{-131688.2} \\
\hline
\multirow{6}{*}{\tabincell{c}{$p=20\%$\\$m=5$}} & S$_{1}$ & -131715.4 & -131711.4 & -131711.3 & -131704.2 & -131704.0 & -131701.7 & -131701.7 & -131701.7 \\
 & S$_{2}$ & -131699.4 & -131695.3 & -131695.3 & -131693.5 & -131693.5 & -131693.0 & -131692.7 & -131692.5 \\
 & S$_{3}$ & -131699.1 & -131698.0 & -131697.5 & -131696.9 & -131694.9 & -131694.8 & -131694.6 & -131694.6 \\
 & S$_{4}$ & -131700.0 & -131696.2 & -131695.7 & -131692.8 & -131692.6 & -131692.1 & -131691.9 & -131691.9 \\
 & S$_{5}$ & -131699.9 & -131697.3 & -131695.5 & -131695.3 & -131694.6 & -131694.5 & -131694.3 & -131693.3 \\
 & Highest DAG & \textbf{-131699.1} & \textbf{-131695.3} & \textbf{-131695.3} & \textbf{-131692.8} & \textbf{-131692.6} & \textbf{-131692.1} & \textbf{-131691.9} & \textbf{-131691.9} \\
\hline
\multirow{5}{*}{\tabincell{c}{$p=25\%$\\$m=4$}} & S$_{1}$ & -131693.9 & -131691.2 & -131690.6 & -131690.6 & -131682.9 & -131682.7 & -131682.5 & -131682.5 \\
 & S$_{2}$ & -131682.7 & -131679.4 & -131678.6 & -131677.9 & -131677.9 & -131677.6 & -131677.6 & -131677.6 \\
 & S$_{3}$ & -131678.3 & -131673.4 & -131669.9 & -131669.6 & -131669.6 & -131669.6 & -131669.4 & -131667.5 \\
 & S$_{4}$ & -131698.4 & -131696.7 & -131694.1 & -131694.1 & -131692.8 & -131688.4 & -131685.9 & -131685.2 \\
 & Highest DAG & \textbf{-131678.3} & \textbf{-131673.4} & \textbf{-131669.9} & \textbf{-131669.6} & \textbf{-131669.6} & \textbf{-131669.6} & \textbf{-131669.4} & \textbf{-131667.5} \\
\hline
\multirow{4}{*}{\tabincell{c}{$p=33\%$\\$m=3$}} & S$_{1}$ & -131700.3 & -131698.8 & -131698.0 & -131695.6 & -131695.5 & -131695.5 & -131695.4 & -131695.1 \\
 & S$_{2}$ & -131688.3 & -131684.7 & -131683.1 & -131681.2 & -131674.8 & -131674.8 & -131674.5 & -131674.5 \\
 & S$_{3}$ & -131699.7 & -131694.6 & -131689.6 & -131689.2 & -131688.9 & -131688.8 & -131688.8 & -131687.8 \\
 & Highest DAG & \textbf{-131688.3} & \textbf{-131684.7} & \textbf{-131683.1} & \textbf{-131681.2} & \textbf{-131674.8} & \textbf{-131674.8} & \textbf{-131674.5} & \textbf{-131674.5} \\
\hline
\multirow{3}{*}{\tabincell{c}{$p=50\%$\\$m=2$}} & S$_{1}$ & -131696.3 & -131694.1 & -131691.7 & -131690.0 & -131689.5 & -131689.4 & -131689.3 & -131689.3 \\
 & S$_{2}$ & -131713.4 & -131704.3 & -131697.5 & -131697.0 & -131690.9 & -131690.8 & -131689.7 & -131689.6 \\
 & Highest DAG & \textbf{-131696.3} & \textbf{-131694.1} & \textbf{-131691.7} & \textbf{-131690.0} & \textbf{-131689.5} & \textbf{-131689.4} & \textbf{-131689.3} & \textbf{-131689.3} \\
\hline
\multirow{3}{*}{\tabincell{c}{$p=80\%$\\$m=2$}} & S$_{1}$ & -131712.3 & -131699.2 & -131698.6 & -131696.1 & -131694.8 & -131693.9 & -131692.0 & -131688.3 \\
 & S$_{2}$ & -131712.4 & -131702.3 & -131700.4 & -131698.4 & -131698.3 & -131697.5 & -131694.5 & -131692.3 \\
 & Highest DAG & \textbf{-131712.3} & \textbf{-131699.2} & \textbf{-131698.6} & \textbf{-131696.1} & \textbf{-131694.8} & \textbf{-131693.9} & \textbf{-131692.0} & \textbf{-131688.3} \\
\bottomrule
\end{tabular}
\caption{BDeu scores of MINOBS and PS-MINOBS algorithm for case study, Reuters-test, under different settings $p=10\%$, $m=10$, $p=20\%$, $m=5$, $p=25\%$, $m=4$, $p=33\%$, $m=3$, $p=50\%$, $m=2$, and $p=80\%$, $m=2$, within different time limits.}
\end{table}


\begin{table}[H]
\scriptsize
\centering
\begin{tabular}{cccccccccc}
\toprule
 \tabincell{c}{PS-MINOBS\\Hyperparameters} & Threads & 0.5h & 1h & 1.5h & 2h & 2.5h & 3h & 3.5h & 4h\\
\hline
 & MINOBS & -645193.1 & -645162.3 & -645117.8 & -645097.9 & -645088.2 & -645084.9 & -645084.9 & -645084.9 \\
\hline
\multirow{11}{*}{\tabincell{c}{$p=10\%$\\$m=10$}} & S$_{1}$ & -645140.2 & -645101.8 & -645101.8 & -645095.2 & -645076.4 & -645075.2 & -645075.2 & -645075.2 \\
 & S$_{2}$ & -645096.8 & -645079.0 & -645076.6 & -645076.6 & -645074.9 & -645073.4 & -645073.4 & -645072.4 \\
 & S$_{3}$ & -645133.8 & -645115.2 & -645113.1 & -645111.5 & -645107.4 & -645106.1 & -645106.1 & -645106.1 \\
 & S$_{4}$ & -645136.6 & -645107.3 & -645107.2 & -645104.1 & -645096.4 & -645094.9 & -645094.9 & -645094.9 \\
 & S$_{5}$ & -645151.4 & -645109.8 & -645108.3 & -645106.2 & -645088.8 & -645088.8 & -645088.8 & -645088.8 \\
 & S$_{6}$ & -645115.8 & -645099.4 & -645096.0 & -645094.6 & -645093.1 & -645093.1 & -645093.1 & -645092.6 \\
 & S$_{7}$ & -645118.9 & -645085.3 & -645083.9 & -645083.9 & -645077.7 & -645073.9 & -645073.9 & -645071.7 \\
 & S$_{8}$ & -645110.7 & -645101.5 & -645101.5 & -645098.5 & -645096.8 & -645096.8 & -645088.9 & -645077.5 \\
 & S$_{9}$ & -645120.8 & -645084.4 & -645084.4 & -645084.4 & -645080.3 & -645080.3 & -645080.3 & -645079.7 \\
 & S$_{10}$ & -645138.9 & -645125.6 & -645125.6 & -645125.1 & -645122.9 & -645122.9 & -645122.9 & -645122.9 \\
 & Highest DAG & \textbf{-645096.8} & \textbf{-645079.0} & \textbf{-645076.6} & \textbf{-645076.6} & \textbf{-645074.9} & \textbf{-645073.4} & \textbf{-645073.4} & \textbf{-645071.7} \\
\hline
\multirow{6}{*}{\tabincell{c}{$p=20\%$\\$m=5$}} & S$_{1}$ & -645121.0 & -645105.4 & -645105.4 & -645103.4 & -645098.9 & -645098.9 & -645095.0 & -645079.1 \\
 & S$_{2}$ & -645134.1 & -645112.8 & -645112.8 & -645091.0 & -645076.2 & -645075.2 & -645074.8 & -645073.4 \\
 & S$_{3}$ & -645125.3 & -645100.7 & -645098.8 & -645097.7 & -645061.1 & -645059.6 & -645059.6 & -645059.2 \\
 & S$_{4}$ & -645107.9 & -645087.1 & -645085.5 & -645081.0 & -645080.3 & -645078.3 & -645078.3 & -645076.2 \\
 & S$_{5}$ & -645132.8 & -645110.4 & -645107.6 & -645104.7 & -645104.7 & -645104.7 & -645104.7 & -645101.9 \\
 & Highest DAG & \textbf{-645107.9} & \textbf{-645087.1} & \textbf{-645085.5} & \textbf{-645081.0} & \textbf{-645061.1} & \textbf{-645059.6} & \textbf{-645059.6} & \textbf{-645059.2} \\
\hline
\multirow{5}{*}{\tabincell{c}{$p=25\%$\\$m=4$}} & S$_{1}$ & -645111.5 & -645084.7 & -645083.6 & -645083.6 & -645079.5 & -645079.5 & -645079.5 & -645079.0 \\
 & S$_{2}$ & -645102.11 & -645078.34 & -645077.71 & -645077.71 & -645077.71 & -645076.47 & -645076.47 & -645076.47 \\
 & S$_{3}$ & -645125.9 & -645091.6 & -645082.0 & -645081.9 & -645076.9 & -645072.6 & -645072.6 & -645070.1 \\
 & S$_{4}$ & -645121.3 & -645095.5 & -645093.0 & -645092.4 & -645077.0 & -645077.0 & -645077.0 & -645076.7 \\
 & Highest DAG & \textbf{-645102.1} & \textbf{-645078.3} & \textbf{-645077.7} & \textbf{-645077.7} & \textbf{-645076.9} & \textbf{-645072.6} & \textbf{-645072.6} & \textbf{-645070.1} \\
\hline
\multirow{4}{*}{\tabincell{c}{$p=33\%$\\$m=3$}} & S$_{1}$ & -645152.6 & -645120.6 & -645115.5 & -645115.5 & -645115.5 & -645079.2 & -645077.5 & -645077.5 \\
 & S$_{2}$ & -645118.6 & -645087.0 & -645084.2 & -645084.2 & -645082.9 & -645079.9 & -645078.3 & -645078.3 \\
 & S$_{3}$ & -645105.9 & -645089.5 & -645084.8 & -645084.8 & -645071.3 & -645065.6 & -645065.6 & -645065.6 \\
 & Highest DAG & \textbf{-645105.9} & \textbf{-645087.0} & \textbf{-645084.2} & \textbf{-645084.2} & \textbf{-645071.3} & \textbf{-645065.6} & \textbf{-645065.6} & \textbf{-645065.6} \\
\hline
\multirow{3}{*}{\tabincell{c}{$p=50\%$\\$m=2$}} & S$_{1}$ & -645139.9 & -645096.3 & -645087.8 & -645084.1 & -645084.1 & -645084.1 & -645083.1 & -645082.8 \\
 & S$_{2}$ & -645167.1 & -645100.8 & -645084.7 & -645081.1 & -645081.1 & -645080.6 & -645079.0 & -645076.1 \\
 & Highest DAG & \textbf{-645139.9} & \textbf{-645096.3} & \textbf{-645084.7} & \textbf{-645081.1} & \textbf{-645081.1} & \textbf{-645080.6} & \textbf{-645079.0} & \textbf{-645076.1} \\
\hline
\multirow{3}{*}{\tabincell{c}{$p=80\%$\\$m=2$}} & S$_{1}$ & -645209.9 & -645132.2 & -645112.4 & -645103.5 & -645102.5 & -645102.5 & -645102.5 & -645101.6 \\
 & S$_{2}$ & -645234.6 & -645185.3 & -645148.7 & -645142.4 & -645139.2 & -645138.9 & -645138.9 & -645137.1 \\
 & Highest DAG & \textbf{-645209.9} & \textbf{-645132.2} & \textbf{-645112.4} & \textbf{-645103.5} & \textbf{-645102.5} & \textbf{-645102.5} & \textbf{-645102.5} & \textbf{-645101.6} \\
\bottomrule
\end{tabular}
\caption{BDeu scores of MINOBS and PS-MINOBS algorithm for case study, Reuters-train, under different settings $p=10\%$, $m=10$, $p=20\%$, $m=5$, $p=25\%$, $m=4$, $p=33\%$, $m=3$, $p=50\%$, $m=2$, and $p=80\%$, $m=2$, within different time limits.}
\end{table}


\begin{table}[H]
\scriptsize
\centering
\begin{tabular}{cccccccccc}
\toprule
 \tabincell{c}{PS-MINOBS\\Hyperparameters} & Threads & 0.5h & 1h & 1.5h & 2h & 2.5h & 3h & 3.5h & 4h\\
\hline
 & MINOBS & -586321.5 & -586239.1 & -586211.8 & -586202.2 & -586185.7 & -586181.0 & -586177.8 & -586173.6 \\
 \hline
\multirow{11}{*}{\tabincell{c}{$p=10\%$\\$m=10$}} & S$_{1}$ & -586215.9 & -586188.0 & -586187.2 & -586184.6 & -586180.9 & -586175.6 & -586175.3 & -586169.8 \\
 & S$_{2}$ & -586216.3 & -586194.1 & -586183.2 & -586180.5 & -586178.1 & -586175.2 & -586175.2 & -586172.2 \\
 & S$_{3}$ & -586215.2 & -586201.9 & -586199.7 & -586196.2 & -586183.9 & -586180.7 & -586180.7 & -586177.4 \\
 & S$_{4}$ & -586216.4 & -586188.3 & -586184.7 & -586183.4 & -586175.1 & -586174.9 & -586174.2 & -586172.8 \\
 & S$_{5}$ & -586222.3 & -586184.9 & -586183.2 & -586179.8 & -586174.9 & -586172.5 & -586172.4 & -586170.8 \\
 & S$_{6}$ & -586211.9 & -586183.7 & -586182.5 & -586180.1 & -586174.6 & -586172.8 & -586172.5 & -586170.3 \\
 & S$_{7}$ & -586230.8 & -586196.0 & -586192.7 & -586191.0 & -586189.8 & -586189.1 & -586189.1 & -586179.3 \\
 & S$_{8}$ & -586216.9 & -586193.7 & -586191.6 & -586190.9 & -586186.8 & -586184.8 & -586178.4 & -586172.9 \\
 & S$_{9}$ & -586199.1 & -586175.9 & -586172.6 & -586169.9 & -586168.1 & -586168.1 & -586168.1 & -586158.8 \\
 & S$_{10}$ & -586209.3 & -586187.3 & -586183.5 & -586183.5 & -586179.6 & -586179.1 & -586179.1 & -586179.1 \\
 & Highest DAG & \textbf{-586199.1} & \textbf{-586175.9} & \textbf{-586172.6} & \textbf{-586169.9} & \textbf{-586168.1} & \textbf{-586168.1} & \textbf{-586168.1} & \textbf{-586158.8} \\
\hline
\multirow{6}{*}{\tabincell{c}{$p=20\%$\\$m=5$}} & S$_{1}$ & -586230.8 & -586202.2 & -586196.9 & -586194.2 & -586187.5 & -586183.7 & -586183.7 & -586181.7 \\
 & S$_{2}$ & -586223.9 & -586194.1 & -586191.9 & -586191.2 & -586184.1 & -586180.2 & -586179.8 & -586179.8 \\
 & S$_{3}$ & -586212.1 & -586179.1 & -586177.1 & -586176.8 & -586175.8 & -586172.0 & -586172.0 & -586171.3 \\
 & S$_{4}$ & -586215.7 & -586187.4 & -586182.8 & -586181.9 & -586179.6 & -586178.4 & -586178.1 & -586177.9 \\
 & S$_{5}$ & -586232.9 & -586181.9 & -586180.5 & -586178.8 & -586173.1 & -586169.2 & -586169.2 & -586168.0 \\
 & Highest DAG & \textbf{-586212.1} & \textbf{-586179.1} & \textbf{-586177.1} & \textbf{-586176.8} & \textbf{-586173.1} & \textbf{-586169.2} & \textbf{-586169.2} & \textbf{-586168.0} \\
\hline
\multirow{5}{*}{\tabincell{c}{$p=25\%$\\$m=4$}} & S$_{1}$ & -586223.4 & -586204.8 & -586192.6 & -586192.6 & -586186.9 & -586178.6 & -586177.5 & -586174.4 \\
 & S$_{2}$ & -586220.9 & -586198.2 & -586192.3 & -586192.1 & -586187.6 & -586183.5 & -586183.5 & -586182.7 \\
 & S$_{3}$ & -586225.9 & -586195.8 & -586185.6 & -586185.6 & -586182.4 & -586177.9 & -586177.9 & -586177.9 \\
 & S$_{4}$ & -586234.2 & -586208.1 & -586203.5 & -586202.4 & -586199.1 & -586195.39 & -586195.4 & -586194.4 \\
 & Highest DAG & \textbf{-586220.9} & \textbf{-586195.8} & \textbf{-586185.6} & \textbf{-586185.6} & \textbf{-586182.4} & \textbf{-586177.9} & \textbf{-586177.5} & \textbf{-586174.4} \\
\hline
\multirow{4}{*}{\tabincell{c}{$p=33\%$\\$m=3$}} & S$_{1}$ & -586232.3 & -586194.4 & -586181.6 & -586181.1 & -586180.1 & -586174.8 & -586174.8 & -586174.8 \\
 & S$_{2}$ & -586233.7 & -586192.4 & -586186.1 & -586185.4 & -586185.4 & -586182.2 & -586180.6 & -586180.5 \\
 & S$_{3}$ & -586232.6 & -586206.3 & -586192.7 & -586192.7 & -586192.7 & -586186.5 & -586180.3 & -586180.3 \\
 & Highest DAG & \textbf{-586232.3} & \textbf{-586192.4} & \textbf{-586181.6} & \textbf{-586181.1} & \textbf{-586180.1} & \textbf{-586174.8} & \textbf{-586174.8} & \textbf{-586174.8} \\
\hline
\multirow{3}{*}{\tabincell{c}{$p=50\%$\\$m=2$}} & S$_{1}$ & -586260.6 & -586208.3 & -586197.3 & -586186.9 & -586184.9 & -586184.9 & -586176.8 & -586171.4 \\
 & S$_{2}$ & -586258.1 & -586218.3 & -586198.9 & -586185.4 & -586176.6 & -586175.7 & -586173.7 & -586168.9 \\
 & Highest DAG & \textbf{-586258.1} & \textbf{-586208.3} & \textbf{-586197.3} & \textbf{-586185.4} & \textbf{-586176.6} & \textbf{-586175.7} & \textbf{-586173.7} & \textbf{-586168.9} \\
\hline
\multirow{3}{*}{\tabincell{c}{$p=80\%$\\$m=2$}} & S$_{1}$ & -586276.5 & -586229.9 & -586201.8 & -586192.3 & -586179.6 & -586176.9 & -586169.9 & -586169.9 \\
 & S$_{2}$ & -586281.9 & -586239.4 & -586219.8 & -586208.9 & -586203.9 & -586193.7 & -586189.9 & -586189.9 \\
 & Highest DAG & \textbf{-586276.5} & \textbf{-586229.9} & \textbf{-586201.8} & \textbf{-586192.3} & \textbf{-586179.6} & \textbf{-586176.9} & \textbf{-586169.9} & \textbf{-586169.9} \\
\bottomrule
\end{tabular}
\caption{BDeu scores of MINOBS and PS-MINOBS algorithm for case study, NewsGroup-test, under different settings $p=10\%$, $m=10$, $p=20\%$, $m=5$, $p=25\%$, $m=4$, $p=33\%$, $m=3$, $p=50\%$, $m=2$, and $p=80\%$, $m=2$, within different time limits.}
\end{table}


\begin{table}[H]
\scriptsize
\centering
\begin{tabular}{cccccccccc}
\toprule
 \tabincell{c}{PS-MINOBS\\Hyperparameters} & Threads & 0.5h & 1h & 1.5h & 2h & 2.5h & 3h & 3.5h & 4h\\
\hline
 & MINOBS & -458992.9 & -458919.2 & -458868.6 & -458852.6 & -458835.2 & -458824.6 & -458821.4 & -458812.5 \\
 \hline
\multirow{11}{*}{\tabincell{c}{$p=10\%$\\$m=10$}} & S$_{1}$ & -458823.5 & -458811.9 & -458797.9 & -458796.7 & -458794.2 & -458792.2 & -458792.2 & -458792.2 \\
 & S$_{2}$ & -458864.2 & -458813.9 & -458801.9 & -458801.9 & -458798.1 & -458797.7 & -458797.1 & -458796.9 \\
 & S$_{3}$ & -458849.9 & -458816.7 & -458806.4 & -458805.1 & -458801.5 & -458795.2 & -458792.9 & -458792.8 \\
 & S$_{4}$ & -458867.7 & -458807.1 & -458796.4 & -458796.4 & -458793.9 & -458792.2 & -458791.9 & -458791.9 \\
 & S$_{5}$ & -458858.9 & -458816.7 & -458805.2 & -458805.2 & -458803.9 & -458802.9 & -458801.7 & -458801.7 \\
 & S$_{6}$ & -458870.4 & -458825.9 & -458802.3 & -458799.6 & -458798.1 & -458796.9 & -458788.8 & -458788.8 \\
 & S$_{7}$ & -458869.5 & -458815.9 & -458806.5 & -458806.5 & -458802.7 & -458792.9 & -458791.6 & -458791.4 \\
 & S$_{8}$ & -458843.1 & -458804.9 & -458795.6 & -458795.6 & -458790.1 & -458787.1 & -458786.4 & -458786.4 \\
 & S$_{9}$ & -458835.2 & -458807.0 & -458800.8 & -458800.8 & -458798.9 & -458796.3 & -458795.7 & -458795.3 \\
 & S$_{10}$ & -458870.4 & -458838.4 & -458820.9 & -458820.9 & -458818.8 & -458814.9 & -458811.5 & -458810.6 \\
 & Highest DAG & \textbf{-458823.5} & \textbf{-458804.9} & \textbf{-458795.6} & \textbf{-458795.6} & \textbf{-458790.1} & \textbf{-458787.1} & \textbf{-458786.4} & \textbf{-458786.4} \\
\hline
\multirow{6}{*}{\tabincell{c}{$p=20\%$\\$m=5$}} & S$_{1}$ & -458854.1 & -458816.9 & -458802.5 & -458798.9 & -458794.8 & -458792.9 & -458790.7 & -458790.3 \\
 & S$_{2}$ & -458878.0 & -458826.6 & -458803.3 & -458800.2 & -458798.9 & -458792.9 & -458791.3 & -458791.3 \\
 & S$_{3}$ & -458882.6 & -458835.9 & -458815.9 & -458815.0 & -458812.8 & -458802.6 & -458796.8 & -458795.9 \\
 & S$_{4}$ & -458841.6 & -458818.9 & -458793.5 & -458793.5 & -458793.5 & -458792.8 & -458790.2 & -458788.9 \\
 & S$_{5}$ & -458885.6 & -458855.0 & -458834.5 & -458830.1 & -458830.1 & -458815.5 & -458810.2 & -458810.2 \\
 & Highest DAG & \textbf{-458841.6} & \textbf{-458816.9} & \textbf{-458793.5} & \textbf{-458793.5} & \textbf{-458793.5} & \textbf{-458792.8} & \textbf{-458790.2} & \textbf{-458788.9} \\
\hline
\multirow{5}{*}{\tabincell{c}{$p=25\%$\\$m=4$}} & S$_{1}$ & -458865.1 & -458826.8 & -458808.6 & -458801.1 & -458801.1 & -458798.4 & -458791.0 & -458788.5 \\
 & S$_{2}$ & -458866.0 & -458815.0 & -458805.9 & -458801.3 & -458801.3 & -458800.6 & -458799.7 & -458798.7 \\
 & S$_{3}$ & -458866.6 & -458827.9 & -458815.8 & -458811.2 & -458808.9 & -458803.3 & -458801.6 & -458800.7 \\
 & S$_{4}$ & -458874.4 & -458826.6 & -458812.9 & -458807.8 & -458807.8 & -458804.9 & -458803.4 & -458800.7 \\
 & Highest DAG & \textbf{-458865.1} & \textbf{-458815.0} & \textbf{-458805.9} & \textbf{-458801.1} & \textbf{-458801.1} & \textbf{-458798.4} & \textbf{-458791.0} & \textbf{-458788.5} \\
\hline
\multirow{4}{*}{\tabincell{c}{$p=33\%$\\$m=3$}} & S$_{1}$ & -458880.8 & -458831.9 & -458808.8 & -458796.8 & -458794.5 & -458794.5 & -458793.5 & -458792.6 \\
 & S$_{2}$ & -458891.5 & -458836.2 & -458819.9 & -458800.3 & -458798.7 & -458798.1 & -458798.1 & -458794.1 \\
 & S$_{3}$ & -458878.6 & -458824.5 & -458805.2 & -458797.8 & -458796.3 & -458795.5 & -458793.6 & -458791.4 \\
 & Highest DAG & \textbf{-458878.6} & \textbf{-458824.5} & \textbf{-458805.2} & \textbf{-458796.8} & \textbf{-458794.5} & \textbf{-458794.5} & \textbf{-458793.5} & \textbf{-458791.4} \\
\hline
\multirow{3}{*}{\tabincell{c}{$p=50\%$\\$m=2$}} & S$_{1}$ & -458915.4 & -458850.5 & -458821.3 & -458815.2 & -458805.0 & -458798.2 & -458796.0 & -458796.0 \\
 & S$_{2}$ & -458936.9 & -458875.4 & -458833.7 & -458816.6 & -458807.2 & -458800.7 & -458798.0 & -458798.0 \\
 & Highest DAG & \textbf{-458915.4} & \textbf{-458850.5} & \textbf{-458821.3} & \textbf{-458815.2} & \textbf{-458805.0} & \textbf{-458798.2} & \textbf{-458796.0} & \textbf{-458796.0} \\
\hline
\multirow{3}{*}{\tabincell{c}{$p=80\%$\\$m=2$}} & S$_{1}$ & -458944.8 & -458882.3 & -458850.2 & -458825.7 & -458811.8 & -458799.6 & -458797.1 & -458791.9 \\
 & S$_{2}$ & -458968.5 & -458908.8 & -458863.6 & -458842.6 & -458828.9 & -458820.4 & -458817.2 & -458813.3 \\
 & Highest DAG & \textbf{-458944.8} & \textbf{-458882.3} & \textbf{-458850.2} & \textbf{-458825.7} & \textbf{-458811.8} & \textbf{-458799.6} & \textbf{-458797.1} & \textbf{-458791.9} \\
\bottomrule
\end{tabular}
\caption{BDeu scores of MINOBS and PS-MINOBS algorithm for case study, NewsGroup-valid, under different settings $p=10\%$, $m=10$, $p=20\%$, $m=5$, $p=25\%$, $m=4$, $p=33\%$, $m=3$, $p=50\%$, $m=2$, and $p=80\%$, $m=2$, within different time limits.}
\end{table}

\renewcommand{\appendixname}{Appendix~}
\section{Supplementary results for case study Pumsb-star-test}
\setcounter{figure}{0} 
\setcounter{table}{0} 

\begin{figure}[H]
\centering
\includegraphics[height=9cm,width=18cm]{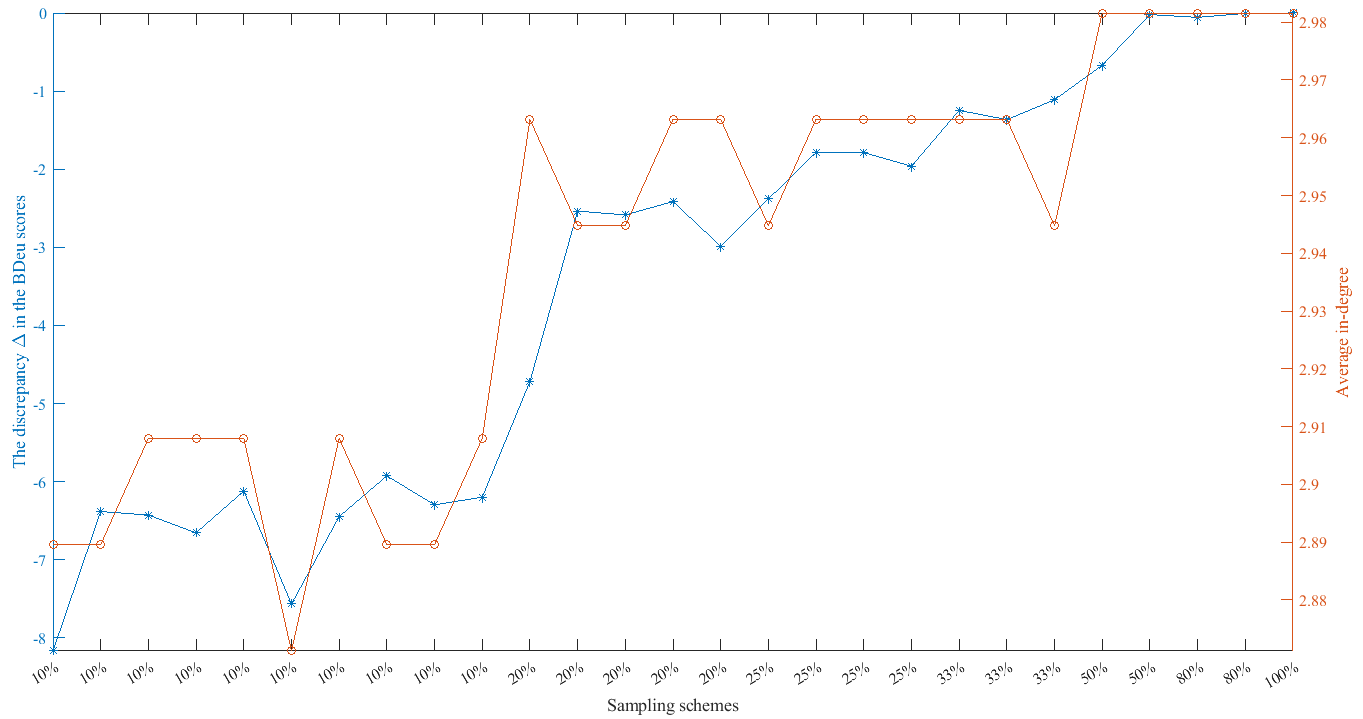}
\caption{The discrepancy $\Delta$ in the BDeu scores and the average in-degree of the Pumsb-star-test structures learnt under different sampling rates. Higher sampling rates consider a higher number of potential CPSs and hence, include many lower scoring CPSs.}
\end{figure}

\end{document}